\documentclass[
 reprint,
nofootinbib,
 amsmath,amssymb,
 aps,
]{revtex4-2}

\usepackage{graphicx}
\usepackage{dcolumn}
\usepackage{bm}
\usepackage{lipsum}
\usepackage[dvipsnames]{xcolor}

\usepackage{multirow}
\usepackage{array}
\usepackage{booktabs}
\usepackage{pifont}

\usepackage{hyperref}

\graphicspath{{./}{./Figures/}}

\newcommand{\x}{\mathbf{x}}

\newcommand{\eg}{\textit{e.g.} }

\newcommand{\vs}{\textit{vs} }

\newcommand{\HD}{D_\mathcal{H} }
\newcommand{\bestmodel}{\textit{BEsT} }

\begin{document}

\preprint{APS/123-QED}

\title{Producing Plankton Classifiers that are Robust to Dataset Shift}

\author{C. Chen}
 \email{cheng.chen@eawag.ch}
\affiliation{Eawag, \"Uberlandstrasse 133, CH-8600, Switzerland}
\author{S. Kyathanahally}
\affiliation{Eawag, \"Uberlandstrasse 133, CH-8600, Switzerland}
\author{M. Reyes}
\affiliation{Eawag, \"Uberlandstrasse 133, CH-8600, Switzerland}
\author{S. Merkli}
\affiliation{Eawag, \"Uberlandstrasse 133, CH-8600, Switzerland}
\author{E. Merz}
\affiliation{Eawag, \"Uberlandstrasse 133, CH-8600, Switzerland}
\author{E. Francazi}
\affiliation{Eawag, \"Uberlandstrasse 133, CH-8600, Switzerland}
\author{M. Hoege}
\affiliation{Eawag, \"Uberlandstrasse 133, CH-8600, Switzerland}
\author{F. Pomati}
\affiliation{Eawag, \"Uberlandstrasse 133, CH-8600, Switzerland}
\author{M. Baity-Jesi}
 \email{marco.baityjesi@eawag.ch}
\affiliation{Eawag, \"Uberlandstrasse 133, CH-8600, Switzerland}

\date{\today}

\begin{abstract}

Modern plankton high-throughput monitoring relies on deep learning classifiers for species recognition in water ecosystems. Despite satisfactory nominal performances, a significant challenge arises from Dataset Shift, which causes performances to drop during deployment. In our study, we integrate the ZooLake dataset, which consists of dark-field images of lake plankton~\cite{kyathanahally:21}, with manually-annotated images from 10 independent days of deployment, serving as \textit{test cells} to benchmark Out-Of-Dataset (OOD) performances.
Our analysis reveals instances where classifiers, initially performing well in In-Dataset conditions, encounter notable failures in practical scenarios. For example, a MobileNet with a 92\% nominal test accuracy shows a 77\% OOD accuracy. We systematically investigate conditions leading to OOD performance drops and propose a preemptive assessment method to identify potential pitfalls when classifying new data, and pinpoint features in OOD images that adversely impact classification.
We present a three-step pipeline: (i) identifying OOD degradation compared to nominal test performance, (ii) conducting a diagnostic analysis of degradation causes, and (iii) providing solutions. We find that ensembles of BEiT vision transformers, with targeted augmentations addressing OOD robustness, geometric ensembling, and rotation-based test-time augmentation, constitute the most robust model, which we call \bestmodel. It achieves an 83\% OOD accuracy, with errors concentrated on container classes. Moreover, it exhibits lower sensitivity to dataset shift, and reproduces well the plankton abundances.
Our proposed pipeline is applicable to generic plankton classifiers, contingent on the availability of suitable test cells.
By identifying critical shortcomings and offering practical procedures to fortify models against dataset shift, our study contributes to the development of more reliable plankton classification technologies.
\end{abstract}

\keywords{plankton, machine learning, classification, out of domain}

\maketitle

\section{Introduction}
Plankton are a fundamental element of water ecosystems. They are responsible for carbon and nutrient upcycling and are used in water treatment plants and fisheries~\cite{behrenfeld:01,falkowski:12}; they are a robust indicator of the health of ecosystems~\cite{xu:01};
and they also are a paradigm of complex ecosystem which is studied as a model system for understanding community composition and dynamics~\cite{franks:02,sergiacomi:18,hu:22}. Additionally, algal blooms, which are sudden increases in the abundance of specific phytoplankton species, can have disruptive effects on water ecosystems. Consequently, there is an ongoing effort to develop forecasting methods for predicting the abundances of plankton taxa at future times~\cite{zohdi:19}.

For these and various other reasons, high-throughput \textit{in-situ} monitoring systems have been increasingly deployed in recent years~\cite{benfield:07,cowen:08,campbell:13,orenstein:20,merz:21}, generating vast amounts of data~\cite{orenstein:15}. These systems often employ underwater cameras, producing an overwhelming volume of images that makes manual annotation unapproachable.
This challenge is addressed through the usage of deep learning classifiers, which use state-of-the-art methods, ranging from convolutional networks~\cite{py:16,ellen:19,orenstein:17,lumini:19,eerola:20,kyathanahally:21} to vision transformers~\cite{kyathanahally:22,yue:23,maracani:23}. The results seem promising, exhibiting test accuracies and F1 scores well above 90\%~\cite{eerola:23}. As a result, ecological studies have commenced leveraging data from these classifiers~\cite{kenitz:20,merz:21,merkli:24}.

However, already in the early attempts at plankton classification, a mismatch between the nominal performance and that at deployment time was noticed~\cite{solow:01,bell:08}.
In fact, it was pointed out that plankton systems are very likely to be subject to \textit{dataset shift} (DS)~\cite{beijbom:15,gonzalez:17,orenstein:20,walker:21}. In simple terms, DS means that the train/test dataset can be so different from the data encountered once the model is deployed, that the declared test performances are not meaningful~\cite{quinonero:09,morenotorres:12}. 
When the difference between testing and deployment are in terms of taxon abundances, we talk of \textit{distributional} DS. 
In plankton monitoring, not only the relatives abundances can change, but also the way images from a same taxon appear, in which case we talk of \textit{compositional} DS.
Factors which can induce compositional DS include changes in the species traits, in environmental factors such as water turbidity, and in instrument conditions like lens cleanliness. 
In the presence of DS, it becomes crucial that machine learning models not only perform well on the test set (which we call \textit{in-dataset}, ID), but also that they generalize well on \textit{out-of-dataset} (OOD) images that may come from a different data distribution than the original dataset.

One way to address this is to adopt an out-of-domain detection approach, in which one tries to infer which images come from a different data distribution, with the intention of abstaining from classifying those images~\cite{saadati:23}. One approach with such an intent, applied to plankton, is based on discarding the examples for which the classifier has a low confidence~\cite{ellen:15,yang:22}.
Otherwise, out-of-domain methods are often used with the aim of improving the generalization to a target domain. There is a large number of existing strategies for this~\cite{wang:22}, but they mostly rely on the availability of more than one training domain,\footnote{Out-of-domain generalization usually refers to training on $M$ distinct domains, and generalizing to a further one~\cite{wang:22}. The specific case of $M=1$ is usually called single-source out-of-domain distribution~\cite{peng:22,ouyang:22}.} and on a well-defined target domain, which in our problem we usually do not have access to.\footnote{While it is outside the scope of this work, it is however possible to define multiple domains for plankton classification. For example, one could take images from different imaging systems, and each would constitute a different domain.}
For this reason, one often talks about DS, where the changes in the data distribution are not clearly defined \textit{a priori}. 

For this reason, recent literature on plankton classification focuses rather on addressing DS through open-set recognition. For example, Ref.~\cite{badreldeen:22} addresses the problem of varying class compositions between different geographic regions, as well as the appearing of novel kinds of external objects, within an open-set recognition framework.

Since the final objective of these classifiers is to infer abundances, with the individual images having little importance, several works focus on quantification~\cite{beijbom:15,gonzalez:17b,gonzalez:19,orenstein:20b}, that is, methods that optimize the estimation of the abundances even in the presence of imperfect classifiers. Focusing on quantification has the additional advantage that, instead of retraining a classifier, one can simply recalculate the correction to the counts that would give the best estimates of the real population.

Independently of the used classification methods, one key issue with DS is to be able to assess performances on a dataset that is constantly varying. This can be addressed through the creation of several additional annotated datasets, called test cells, which represent deployment conditions~\cite{beijbom:15,gonzalez:17}.

\bigskip
Most solutions for assessing performance under possible DS are based on the availability and the production of large amounts of data (see App.~\ref{app:related-work} for a summary of recent work on DS in plankton monitoring), 
which must be annotated by expert taxonomists. While this data is available for datasets such as WHOI~\cite{orenstein:15}, this is often not the case with other datasets, which are much smaller, and there are therefore no studies of DS.
Furthermore, while these methods assess the presence of DS, they do not address its origin: \textit{What are the characteristics of plankton images, that cause DS?}
While it is clear that distributional DS appears in the abundances that are present in the ecosystem, it is not clear what else is changing in the images, giving rise to compositional DS. The evidence of further changes beyond the abundance distribution is in fact indirect, by assessing that the test performance drop at deployment stage. 
One would therefore like to be able to quantitatively link model performances with DS, in order to answer questions such as:
\textit{Are the model performances directly related to DS? Which are the image features which most influence the DS robustness? Which classes are harder/easier more robust under DS? How well will a classifier perform on a new unlabeled dataset?} 
Moreover, this kind of assessment is model-dependent, and there is to our knowledge no systematic study of the model dependence of DS.
Finally, once one observes that DS is present, solutions are needed to address it: \textit{Which models and methods are the most efficient to address DS?}

\bigskip
In this paper, we address DS on the freshwater plankton camera images from the Dual Scripps Plankton Camera~\cite{orenstein:20,merz:21}, in a situation that is much more data-scarce than the one addressed in~\cite{beijbom:15}.\footnote{In Ref.~\cite{beijbom:15}, the ID dataset contained over 1000 per class, and is the result of a massive annotating campaign, lasted years, which can hardly be replicated by future studies on novel monitoring systems.} 
We train models on the ZooLake2.0 dataset, which is a minor extension of the ZooLake dataset~\cite{kyathanahally:21b}. We perform a typical train/validation/test splitting, and call the corresponding performances \textit{in-dataset} (ID).
We then simulate the deployment of the models in the field, by further manually annotating 10 \textit{out-of-dataset} (OOD) test cells, each corresponding to a separate different day of sampling, in a different time of the year. An alike OOD scheme was already used in the past~\cite{beijbom:15,gonzalez:17,walker:21}. With this setup we conduct our work in three steps as shown in Fig.~\ref{fig:diagram}:

\begin{figure*}[t]
    \centering    \includegraphics[width=\textwidth]{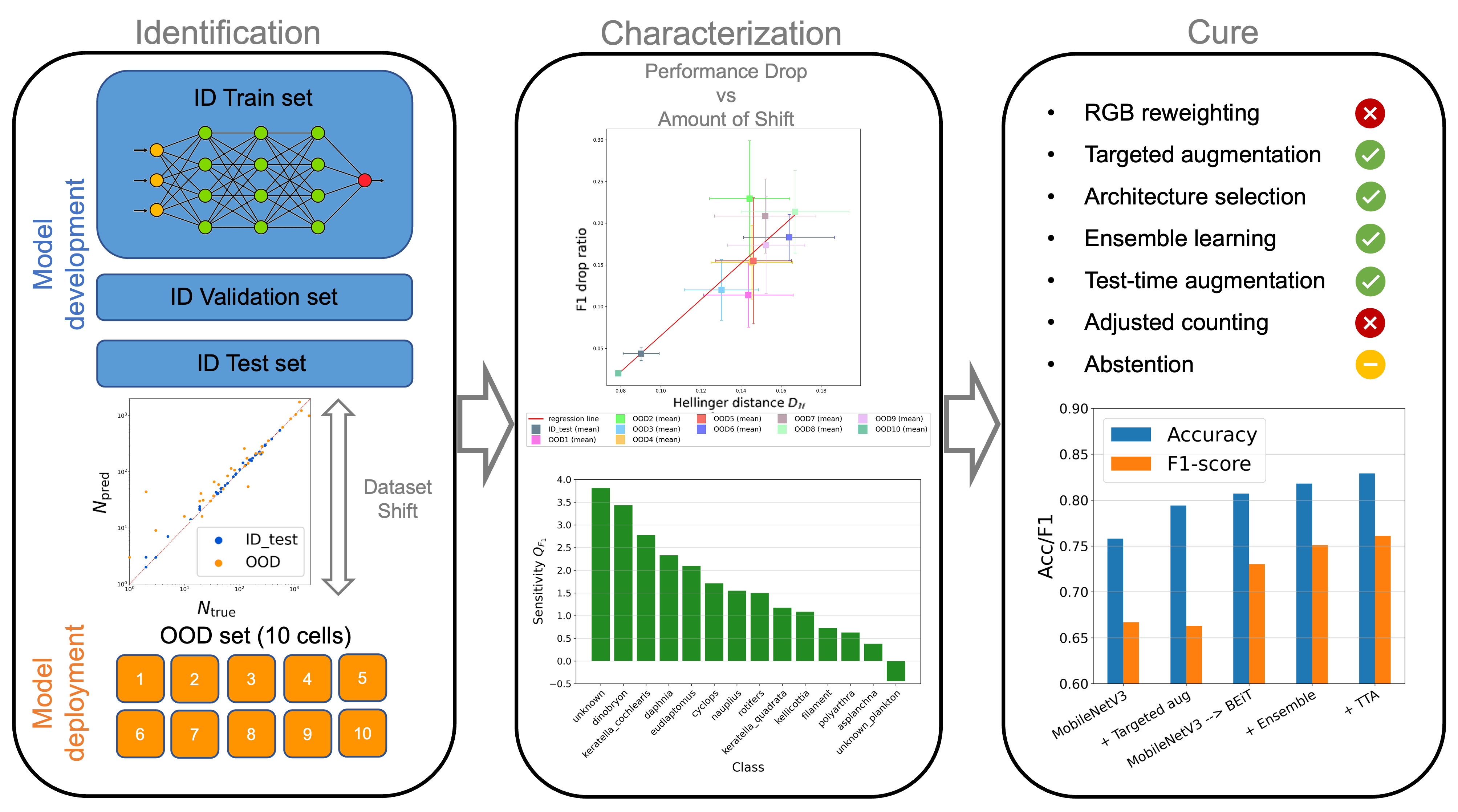}
    \caption{Schematic flowchart of our pipeline to address dataset shift. 
    \textbf{Identification:}
    Models are usually trained, validated and tested on in-dataset (ID) data, which, due to dataset shift (DS), differs from the data found at deployment. By using 10 out-of-dataset (OOD) test cells, we simulate the performance at deployment, which is lower than the nominal ID test performance.
    \textbf{Characterization:}
    We quantify dataset shift and the sensitivity of our models to dataset shift.
    \textbf{Cure:}
    We study the effect of several modeling choices on combating DS.
    }
    \label{fig:diagram}
\end{figure*}

\begin{description}
    \item [Identify dataset shift]  
    In medium-size lakes, variations in the observed taxa and debris can be argued to be lower, and events such as the appearing of new taxa are rare. 
    We check for dataset shift in the most stringent setting (no clear change in conditions of the instruments, lake conditions, new species, and so on), and still find that it has a very strong influence on plankton classification. For example, during regular deployment, the classifier performance drops by between 10\% and 20\% with respect to the ID test accuracy.
    We find that while distributional DS is present, the performance drop must be attributed mainly to compositional DS. We quantify the two kinds of shifts and propose a way to re-assess the performance of a classifier on new data, based on the size of these shifts and how they affect different classes (since some classes suffer strongly, while others barely do).
    
    \item [Characterize dataset shift] 
        Compositional shift is due to variations between ID and OOD, but it is not
    clear what these variations are. We identify these
    variations by studying which features of the images
    mostly affect classification. Inferring whether there
    are specific features of the images which are making
    the OOD classification harder, allows us to specifically act on those.

    \item [Cure dataset shift] We are interested in finding the models that best deal with dataset shift. We analyze different architectures to see which is more robust with respect to DS, and use several methods for dealing with DS.
    We find that RGB histogram reweighting and adjusted-count quantification do not help improve the OOD performance. Instead, ensembling, targeted data augmentation, test-time augmentation, and architecture selection are helpful to address DS.
\end{description}

Our work follows a pipeline that can be fully reproduced for model selection with any instrument and in any plankton ecosystem (or any generic monitoring task). We also provide our codes at \url{https://github.com/cchen07/Plankiformer_OOD}.

\section{Materials and Procedures}

\subsection{Dataset shift}\label{sec:ds}

The aim of any machine learning classification application is to have a model $f(\x)$ which, given an input $\x$ (in our case $\x$ is an image), can assign it the correct label $y$ with a low error rate. The error rate is measured through a loss function $\ell(f(\x),y)$, which compares the output of the model, $f(\x)$, with the true label. Ideally, the model needs to perform well over all possible examples it may see at deployment stage. That is, the ultimate aim is to minimize the \emph{population loss},
\begin{equation}\label{eq:poploss}
    \mathcal{L}_\mathrm{pop} = \int P(\x,y)\, \ell(f(\x),y)\, d\x\, dy\,,
\end{equation}
where $P(\x,y)$ is the probability that an input $\x$, with label $y$ will be found at deployment stage.

Since, while training, validating and testing the model, we do not have access to $P(\x,y)$, we usually resort to the \emph{empirical loss},
\begin{align}\label{eq:emploss}
    \mathcal{L}_\mathrm{emp}^\mathrm{(train)} &= \frac{1}{M_\mathrm{train}} \sum_{i=1}^{M_\mathrm{train}} \ell(f(\x_i),y_i)\,,\\
    \mathcal{L}_\mathrm{emp}^\mathrm{(val)} &= \frac{1}{M_\mathrm{val}} \sum_{i=1}^{M_\mathrm{val}} \ell(f(\x_i),y_i)\,,\\
    \mathcal{L}_\mathrm{emp}^\mathrm{(test)} &= \frac{1}{M_\mathrm{test}} \sum_{i=1}^{M_\mathrm{test}} \ell(f(\x_i),y_i)\,,
\end{align}
where we emphasized that one usually splits the data into a training set of size $M_\mathrm{train}$ (for finding optimal weights), validation set of size $M_\mathrm{val}$ (for model and hyperparameter selection), and test set of size $M_\mathrm{test}$ (for performance reporting).

The examples $\{(\x_i,y_i)\}_{i=1}^M$ in the empirical loss are respectively distributed according to $P_\mathrm{train}(\x,y),P_\mathrm{val}(\x,y)$ and $P_\mathrm{test}(\x,y)$. If these are representative of the population data distribution, that is if $P_\mathrm{train}(\x,y)=P_\mathrm{val}(\x,y) = P_\mathrm{test}(\x,y) = P(\x,y)$, then the reported test performance is representative of the population loss.

While, by enforcing a random splitting of the dataset, stratified by abundance,\footnote{This means we enforce that the relative abundance of each class stay stable among the splittings. When the nature of the distribution shift is known, stratification can be used to provide robust performance assessments, by ensuring that the same shift occurs from train to validation to test set (see \textit{e.g.} Ref.~\cite{schuer:23}).
} 
one can obtain $P_\mathrm{train}(\x,y)\approx P_\mathrm{val}(\x,y) \approx P_\mathrm{test}(\x,y)$, there are several reasons why the dataset might not be representative of $P(\x,y)$:
\begin{description}
    \item[High dimensionality] The inputs $\x$ are very high-dimensional, therefore, and to appropriately reproduce $P(\x,y)$, it needs to be densely sampled within the test set.
    \item[Biased sampling] Often, dataset collection (hence, test set creation) does not faithfully reproduce the conditions found at deployment stage. 
    For example, there can be an augmented effort to find examples belonging to minority classes, junk images can be discarded, or also, images where the human annotator is unsure may be discarded.
    \item[Non-stationarity] The function $P(\x,y)$ represents the images that are found at deployment stage, but these may vary in time for reasons that go from environmental changes and instrument degradation, to the community dynamics itself. Instead, the test set is collected once and for all, so it will eventually not be fully representative of what is found at deployment stage.
\end{description}

Since we are interested in how well the test performances represent those that one finds at deployment time, we will mostly focus on the similarity between $P_\mathrm{test}(\x,y)$ and $P(\x,y)$, the first being ID, and the second OOD.

Since, in plankton monitoring, $P(\x,y)$ varies over time, it was suggested to build 
$P_\mathrm{test}(\x,y)$ as an unbiased union of separate sampling sessions~\cite{gonzalez:17}. While this still cannot reproduce the full variability of plankton populations, it at least reduces distribution-shift effects due to sampling bias.
Likewise, one could suggest that, if one wants to add a specific taxon to a dataset, they should not only add images related to the target taxon, but a whole sampling period containing that taxon, in order not to influence the distribution of labels found in the field. This however (i) does not address compositional shift and (ii) only partially addresses distributional shift, since it 
influences the frequency of appearance of certain communities with respect to others, and in any case it is unlikely that any two far-enough sampling days have a similar $P(y)$. Furthermore, as we will see, the $P(y)$ is only a smaller part of the domain shift problem.

\paragraph*{Distributional and compositional dataset shift}
We can use the definition of conditional probability (this is related to Bayes' formula), 
\begin{equation}\label{eq:bayes}
P(\x,y)=P(\x|y)P(y)\,,    
\end{equation}
to elicit the sources of DS in the population loss from Eq.~\eqref{eq:poploss},
\begin{equation}\label{eq:poploss-bayes}
    \mathcal{L}_\mathrm{pop} = \int P(\x|y)P(y)\, \ell(f(\x),y)\, d\x\, dy\,.
\end{equation}
In Eq.~\eqref{eq:poploss-bayes}, we see that the value of the population loss changes if either $P(y)$ or $P(\x|y)$ are different from their equivalents in the test set, $P^\mathrm{(test)}(y)$ and $P^\mathrm{(test)}(\x|y)$. 
So, if either of these two quantities in the test set does not reflect what is found in the field, the model performance will be different than expected. 
If $P(y)\neq P^\mathrm{(test)}(y)$, we talk of distributional DS.
If $P(\x|y)\neq P^\mathrm{(test)}(\x|y)$, we have compositional DS.

In plankton classification, $P(y)$ represents the abundance of each taxonomic unit. Instead, $P(\x|y)$ is the probability that, given a class $y$, it has an appearance $\x$. In practice, it represents the fact that a specific taxon (or auxiliary class) can appear in different ways. For example, there can be morphological differences within the same taxonomic groups (due \eg to nutrient scarcity), more turbidity in the water, or varying light conditions. In all those cases, the same class will have a different appearance.

\begin{figure*}[t]
    \centering
    \includegraphics[width=\textwidth]{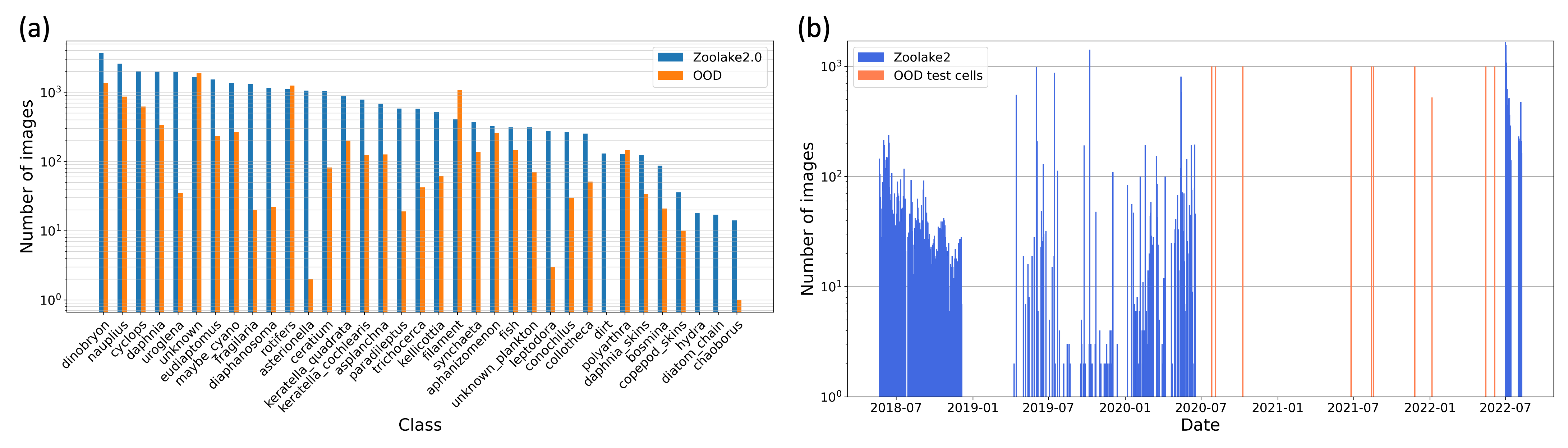}
    \caption{\textbf{(a):} Distribution of class abundances in the ZooLake2.0 and OOD dataset.
    \textbf{(b):} Dates of collection of the images in the ZooLake2.0 dataset and in the test cells.}
    \label{fig:dataset}
\end{figure*}

\subsection{Data}\label{sec:data}
\subsubsection{ZooLake2.0 dataset}
We use images from the Dual-magnification Scripps Plankton Camera (DSPC) deployed in lake Greifensee, in Switzerland~\cite{merz:21}. These are color images of varying size with a black background, taken at three meters depth, of objects which range from 0.1mm to 10mm. These are images from the 0.5x magnification of the DSPC, which mainly targets zooplankton and large phytoplankton colonies. The DSPC camera in the lake Greifensee takes one picture every second for 10 minutes every hour. To avoid repeated imaging of the same organism, we subset the sampling frequency to one picture every 6 seconds~\cite{merz:21}. Each picture includes several single plankton organisms, small images are cut out of large picture. The dataset is an extension of the ZooLake dataset~\cite{kyathanahally:21b}, which we integrated with: (i) the addition of a larger number of junk images, which better approaches the ratio of junk images (\eg \texttt{unknown} or \texttt{dirt} categories) that is found in typical monitoring days;\footnote{The reason why we felt the need to extend the need to integrate ZooLake with new images is that we did not want our results to trivially descend from a distribution which under-represents junk classes.
} 
(ii) the discard of 4 mislabeled examples.\footnote{Two images of \texttt{dinobryon}, one of \texttt{asterionella}, and one of \texttt{rotifers} are discarded because of mislabelling or simultaneous occurrence of multiple species.} The new dataset, that we call ZooLake2.0, contains 29499 images, separated in 35 classes.\footnote{Throughout this paper, we use the term \textit{class} to indicate a classifier category, and \textit{not} the taxonomic rank.} We show the abundance distribution in Fig.~\ref{fig:dataset}a (blue column). Most of the classes indicate zooplankton taxa, identified at a genus or species level (\eg \texttt{daphnia} or \texttt{keratella\_quadrata}). Others, \eg \texttt{asterionella}, identify phytoplankton colonies, non-plankton objects such (\eg \texttt{dirt} or \texttt{fish}), and unidentifiable objects (\eg \texttt{unknown} or \texttt{unknown\_plankton}). More details on the ZooLake2.0 dataset can be found in App.~\ref{app:data}.
We provide the ZooLake2.0 dataset for free access at \url{https://doi.org/10.25678/000C6M}.

\subsubsection{OOD test cells}
The ZooLake2.0 dataset is treated as a normal dataset for training machine learning models. We call it the ID dataset, and perform an 80:5:15 splitting in train, validation and test set, stratified by class abundance.
However, if we want to estimate the OOD performance, we need to go further.

We build 10 OOD test cells, that we will use to validate the OOD performance of our models. Following previous work on the topic~\cite{orenstein:15,gonzalez:17}, we make sure that each test cell represents what would actually be found in a generic deployment context.
Each test cell represents a randomly chosen day, with some mild restrictions. Cells 1-5 are random dates without blooms. For cells 6-10 we did not explicitly impose an absence of blooms, but we restricted the sampling to days with Chl-a levels under 40 to not have a dataset with a phytoplankton bloom. The reason for avoiding blooms is that we wanted our test cells not to be only representing few taxa.
Since the plankton populations vary throughout the day, for each day we evenly sampled from uniformly distributed hours of the day: 1:00, 6:00, 11:00, 16:00, 21:00.
To spare human annotation effort, when a test cell contains more than 1000 images (9 out of 10 cases), we restrict to 1000 random images. 
This ensures that human-induced bias in the image distributions is minimized, and that the test cells represent what would actually be found in a generic deployment context.

\begin{figure}
    \centering
    \includegraphics[width=\columnwidth]{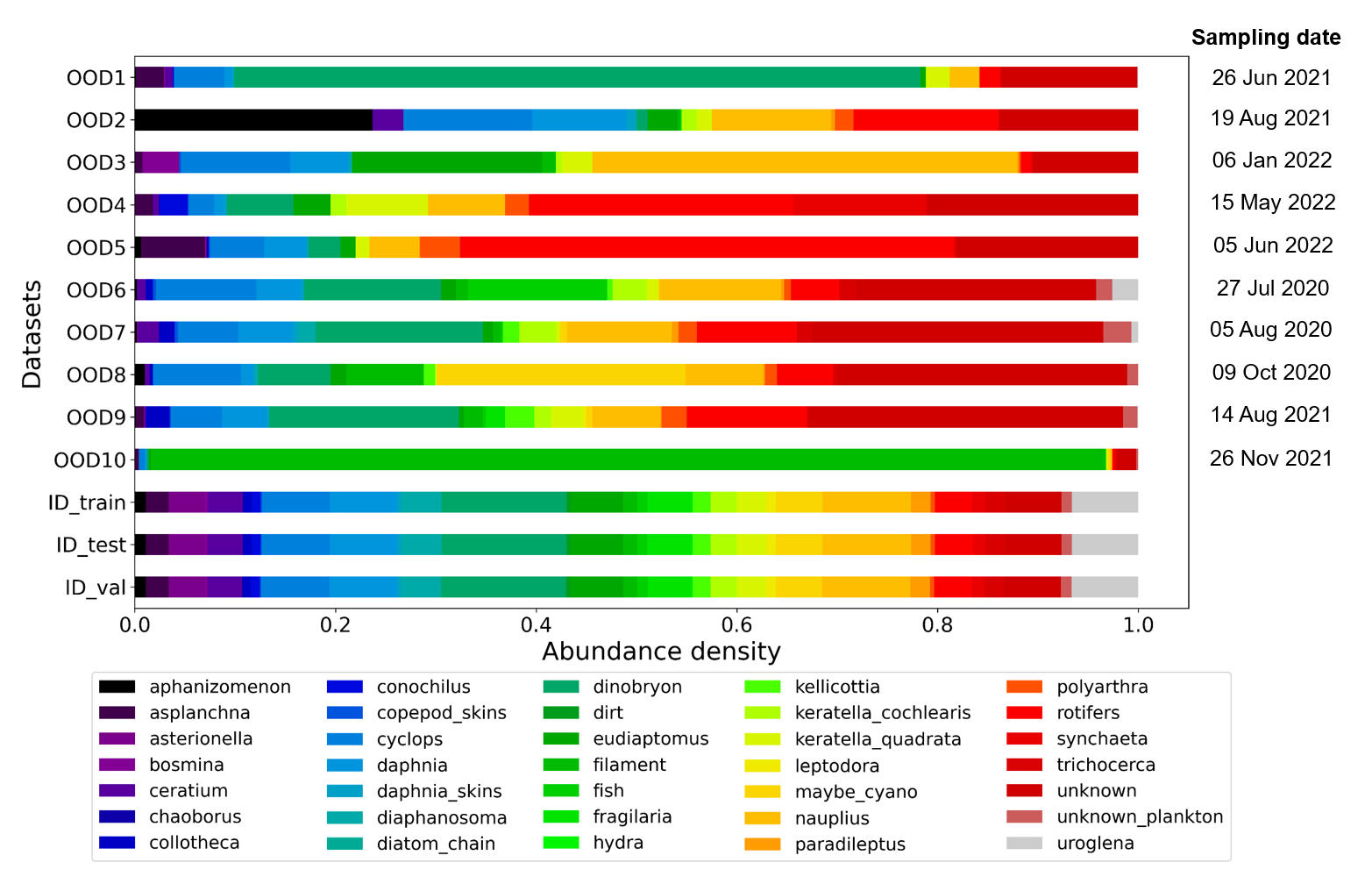}
    \caption{
    Distribution of classes in the 10 test cells. The column on the right indicates the day of sampling that the test cell comes from. The ID dataset is not assigned a date, because it comes from different days (Fig.~\ref{fig:dataset}b).
    }
    \label{fig:test-cells}
\end{figure}

In this work we want to investigate DS in its purest version, \textit{i.e.} a DS
that is not clearly attributable to any external factor (as could be changing camera, site, aging of the instrument, and so on). Any of these additional factors would add up to what we are finding, and be a source of increased DS. 
Therefore, all the images (both from ZooLake2.0 and from the OOD test cells) come from the same camera, deployed at the same depth in the same station, throughout the whole year. In order to ensure that the aging of the instrument is not a dominant factor, the ZooLake2.0 images come with a 2-year gap, and the test cells are randomly sampled in this 2-year gap (Fig.~\ref{fig:dataset}-right).
With this construction, we can only impute dataset shift to intrinsic factors of plankton imaging.
In Fig.~\ref{fig:test-cells} we show the class distribution, $P(y)$, in each of the OOD test cells, and of the ID dataset.

\subsection{Model training}\label{sec:models}
We train two different families of models, CNNs and Vision Transformers, including the following deep learning architectures: 
\begin{description}
    \item[Convolutional Networks] MobileNetV3-Large~\cite{howard2019searching}, DenseNet-161~\cite{huang2017densely}, EfficientNet-B2~\cite{tan2019efficientnet},
    EfficientNet-B7~\cite{tan2019efficientnet}.
    \item[Transformers] DeiT-Base~\cite{touvron:20}, ViT-Base~\cite{dosovitskiy2020image}, BEiT-Base~\cite{bao2021beit}, Swin-Base~\cite{liu2021swin}.
\end{description}

The specific choice of the architectures is motivated by previous literature on plankton classification: DenseNets were found to be well-performing in Ref.~\cite{lumini:19}, MobileNets and EfficientNets in Ref.~\cite{kyathanahally:21}, DeiTs in Ref.~\cite{kyathanahally:22}, Swin transformers in Ref.~\cite{yue:23}, and BEiTs in Ref.~\cite{maracani:23}. 

In particular, MobileNets are lightweight models with a performance that is only slightly lower than that of much bigger models, so they can be used to systematically test the effectiveness of several techniques, with a relatively low investment in terms of computing resources.

For all models, we make use of transfer learning~\cite{tan:18}, using pretrained weight configurations as an initialization for our models.
Each architecture is trained three times.\footnote{The resulting trained network is different because of the initialization of the final layer, of the
randomness in the learning dynamics induced by stochastic gradient descent (also called gradient noise), and of data augmentation.} This allows us to use the best model based on the validation dataset, and to create ensemble models.
Details on training and hyperparameter tuning are provided in App.~\ref{app:models}.

\subsubsection{Data augmentation}\label{sec:augment}
\begin{figure}[t]
    \centering
    \includegraphics[width=\columnwidth]{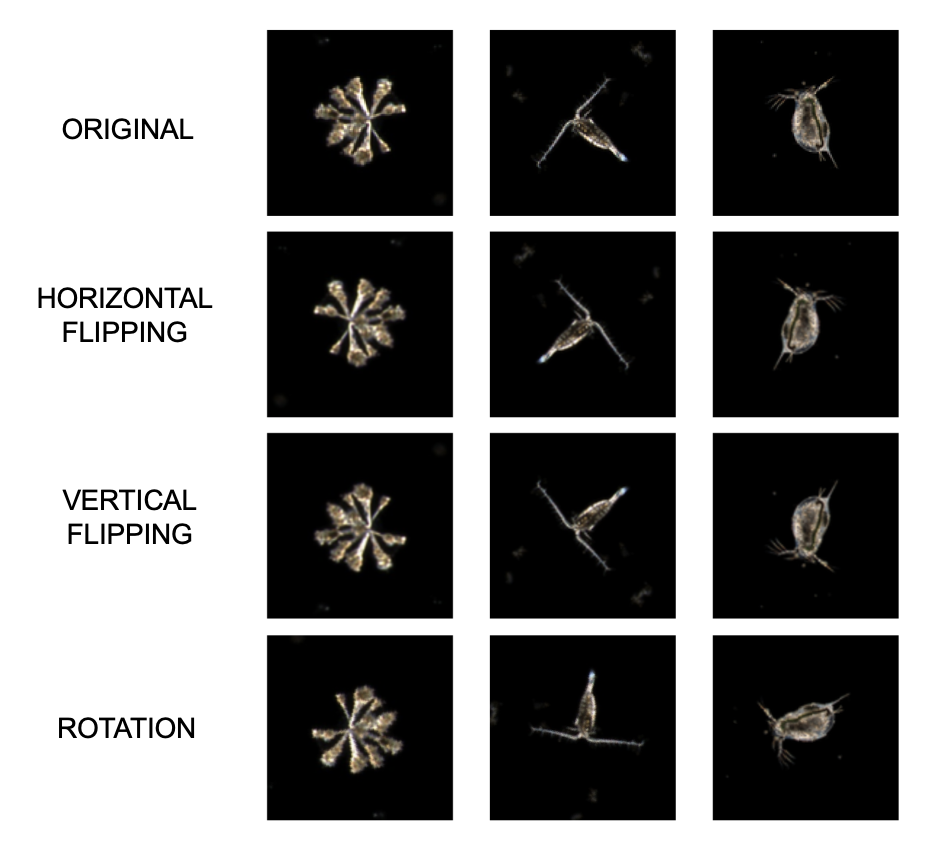}
    \caption{
    Graphical examples of basic augmentations compared to original images, including random flipping and rotation.
    }
    \label{fig:aug_basic}
\end{figure}

Data augmentation is a common strategy to improve the generalization of machine learning models. However, the number of possible augmentations is very large and hard to select. We compare four different kinds of augmentation:
\begin{description}
    \item[None] This is a baseline to show the model performances when augmentation is not used.
    \item[Basic] These models are trained with the basic augmentations presented in Ref.~\cite{kyathanahally:21}, which are random rotation and random flipping. Some examples of the augmented image are shown in Fig.~\ref{fig:aug_basic}. The probability of applying horizontal and vertical flipping are both 0.5, the degree of rotation is between $0^\circ$ and $180^\circ$. 
    \item[Targeted] This is a set of augmentations which target OOD sensitivity. We calculate the OOD sensitivities of a model, identify the features to which the model is most sensitive, and apply augmentations that target those specific features. We describe it in App.~\ref{app:augment}.
    \item[Extra] In addition to applying the basic and targeted augmentations, we add four more augmentations, which are random posterizing, random solarizing, random sharpness and random contrast. These are commonly used augmentations that transform the color characteristics of image. 
    
\end{description}

\subsubsection{Ensemble models}\label{sec:ensemble}
We also train ensemble models, that is, we use $n_m=3$ different models to produce predictions on each image, and determine the class of the image based on the average confidence vector. 

Given $N_c$ classes and $n_m$ models to be ensembled, we write the $i^\mathrm{th}$ component related to the $j^\mathrm{th}$ as $c_{i}^{(j)}$.
The average confidence vector is obtained in two ways:
\begin{description}
    \item[Arithmetic] For each component $i$, the ensemble prediction confidence is $c_{i}^\mathrm{(arith)} = \frac{1}{n_m}\sum_j^{n_m}c_{i}^{(j)}$.
    \item[Geometric] For each component $i$, the ensemble prediction confidence is $c_{i}^\mathrm{(geo)} = \sqrt[{n_m}]{\prod_j^{n_m} c_{i}^{(j)}}$.
\end{description}

\subsection{Metrics}\label{sec:metrics}
We here list the metrics that we use throughout the paper.

\subsubsection{Per-class descriptors}
Let us call the $TP(y), FP(y)$ and $FN(y)$ the true positives, false positives and false negatives related to class $y$. We can then define the per-class recall\footnote{The recall is sometimes called true-positive rate, or sensitivity} related to class $y$ as
\begin{equation}
    r(y) = \frac{TP(y)}{TP(y)+FN(y)}\,,
\end{equation}
the per-class precision as
\begin{equation}
    p(y) = \frac{TP(y)}{TP(y)+FP(y)}\,,
\end{equation}
and the per-class $F_1$ score as
\begin{equation}
    F_1(y) = 2\frac{r(y)\,p(y)}{r(y)+p(y)}\,.
\end{equation}

Finally, we also define the fall-out (also known as false-positive rate), which will be useful for adjusted quantification
\begin{equation}
    f(y) = \frac{FP(y)}{TN(y)+FP(y)}\,.
\end{equation}

\subsubsection{Micro-averaged descriptors}
Micro-averaged descriptors give the same weight to all the examples in the testing set, so the errors will be dominated by the largest population.

The accuracy is the number of correct examples out of the total,
\begin{equation}\label{eq:acc}
    A = \frac{TP}{n} = \frac{\sum_y TP(y)}{n}
\end{equation}
where we defined the total number of true positives as $TP = \sum_y TP(y)$, and the total number of examples $n$, which is connected to the number of examples per class, $n(y)$, through $n=\sum_y n(y)$. 

By noticing that $n(y)=TP(y)+FN(y)$, we can also write the accuracy in terms of the per-class recall,
\begin{equation}\label{eq:micro-rec}
    A = \sum_y \frac{n(y)}{n}\,r(y)\,.
\end{equation}
From Eq.~\eqref{eq:micro-rec} we can see that the accuracy and the micro-averaged recall are the same quantity.

\subsubsection{Expected accuracy}\label{sec:exp-acc}
The ratio $\frac{n(y)}{n}$ is trivially an estimator of $P(y)$.
Equivalently, $r(y)$ is related to $P(\x|y)$, since it is a measure of how good the classifier recognizes elements of a class, given its appearance. If $r(y)$ is fully representative of an immutable $P(\x|y)$, then future measurements (also performed on data with the same $P(\x|y)$) will give the same $r(y)$, regardless of $P(y)$ but not of $P(\x|y)$.

Therefore, in the total absence of dataset shift going from ID to OOD ($P_\mathrm{ID}(y)=P_\mathrm{OOD}(y)$ and $P_\mathrm{ID}(\x|y)=P_\mathrm{OOD}(\x|y)$ do not change), if we measure the accuracy in a test set, it will remain stable when using another test set. If the only source of dataset shift is $P(y)$, then $r(y)$ stays stable, and one expects that the accuracy, measured through Eq.~\eqref{eq:acc} will be equal to
\begin{equation}\label{eq:exp-acc}
    A_\mathrm{exp} = \sum_y \frac{n_\mathrm{OOD}(y)}{n_\mathrm{OOD}}\,r(y)\,.
\end{equation}
The condition $A=A_\mathrm{exp}$ must be met if the dataset shift is purely distributional. 

As for purely compositional DS, it is easier to check that this never happens. In order to have purely compositional DS, the ID $P(y)$ should stay stable throughout the OOD cells. Fig.~\ref{fig:test-cells} shows that this is not the case.

\subsubsection{Macro-averaged descriptors}
Since $\frac{n(y)}{n}$ in Eq.~\eqref{eq:micro-rec} is an estimator of $P(y)$, this term makes $A$ dependent of $P(y)$, so it is more influenced by more abundant classes. To give every class the same influence on $A$, we can replace it with $\frac1{N_c}$, that is the same for all classes, and get the \textit{macro-averaged} recall, which is not directly affected by the imbalance,
\begin{equation}\label{eq:macro-recall}
    R = \frac{1}{N_c} \sum_{y=1}^{N_c}r(y)\,.
\end{equation}
Equivalently, the macro-averaged precision and $F_1$ score are
\begin{align}
\label{eq:macro-precision}
    P &= \frac{1}{N_c} \sum_{y=1}^{N_c}p(y)\,,\\
\label{eq:macro-f1}
    F_1 &= \frac{1}{N_c} \sum_{y=1}^{N_c}F_1(y)\,.
\end{align}

\subsubsection{Further macro-averaged metrics for quantification}
Metrics for quantification do not check the images one by one, but rather check how well the abundances of each class are reconstructed. Since in plankton monitoring one is usually only interested in taxon abundances, it is normal to resort to such quantities~\cite{beijbom:15,gonzalez:17}.

If we call $\hat{n}(y)$ the abundance estimated by the classifier and $n(y)$ the true abundance, we can define the per-class bias on a dataset as:
\begin{equation}
    B(y) = \hat{n}(y)-n(y)\,.
\end{equation}
This quantity only makes sense when defined per-class, since it averages to zero when taking all the classes ($\sum_y[\hat{n}(y)-n(y)]=n-n=0$), but has the advantage of indicating whether some classes are systematically over- or under-represented by the classifiers.

The two metrics that we use here are the Normalized Mean Absolute Error (NMAE)
\begin{equation}
    NMAE = \frac{1}{N_c} \sum_{y=1}^{N_c} \frac{|\hat{n}(y)-n(y)|}{n(y)}\,,
\end{equation}
and the Bray-Curtis dissimilarity
\begin{equation}
    \mathcal{D} = \frac{\sum_y|\hat{n}(y)-n(y)|}{\sum_y [\hat{n}(y)+n(y)]}\,,
\end{equation}
which is 0 for two identical  distributions, and 1 for two maximally different ones.

\subsubsection{Drop in performance}
Since we are interested in assessing whether and how much the performance drops from ID to OOD, we also define some measurements of the drop in performance.
Given a metric $\mathcal{M}$, with a $\mathcal{M}_\mathrm{train}$ value calculated on the training set and a value $\mathcal{M}_\mathrm{test}$ calculated on a generic test set, we define drop as
\begin{equation} \label{eq:drop}
    \mathcal{M}^\mathrm{(drop)} = \mathcal{M}_\mathrm{train} - \mathcal{M}_\mathrm{test}\,.
\end{equation}
By expressing $\mathcal{M}^\mathrm{(drop)}$ with respect to the training set instead of the whole ID dataset, we can compare the DS of the OOD cells with that of the ID test set, which would serve as a baseline.

\subsubsection{OOD performances}
We have 10 OOD test cells. OOD metrics can be calculated in two ways:
\begin{description}
    \item [macro-OOD] We calculate the performance separately on each of the OOD sets, and then average.
    \item [micro-OOD] We merge all the OOD sets together, and then calculate the metric on the resulting aggregated OOD set.
\end{description}

While it is usually desirable to treat the OOD sets one by one (\textit{i.e.} macro-OOD), metrics such as the F1 score give artificially low results when calculated on test sets where some of the classes have a very low number of examples. Therefore, in such cases micro-OOD performances are more informative.

\subsection{Describing the images}\label{sec:feature}
We describe the images through 67 standard descriptors, listed in App.~\ref{app:feature}. 
Additionally, we define two quantities that we use for a preliminary description of our data:

\paragraph{Crop Index}
We want to quantify when and how much plankton images are cropped, due to the organism being on the border of the camera. To this aim, we define a crop index, which counts the number of pixels of the organism that overlap with the image boundary. Since our images have a black background, this quantity is simply calculated by counting the number of non-black pixels at the image boundary (pixels with value smaller than 5 are still considered as black).

\paragraph{Blurriness}
In determining the image's degree of blurriness, we initially compute its Laplacian, a two-dimensional isotropic representation of the image's second spatial derivative.  
The Laplacian highlights regions in the image where there are rapid intensity change. The degree of blurriness of an image is then calculated by the average of the absolute values of Laplacian.

\subsection{Quantifying dataset shift}\label{sec:dist}

\subsubsection{Quantifying distributional dataset shift}\label{sec:disty}
To calculate the dissimilarity between two distributions of classes, $P(y)$ and $Q(y)$, we use the scalar product,
\begin{equation}\label{eq:distributional-DS}
    d_y = 1 - P\cdot Q \equiv 1-\frac{1}{N_c}\sum_{y=1}^{N_c} P(y) Q(y)
    \,.
\end{equation}
The quantity $d_y$ is 0 if $P(y)=Q(y)$, and 1 if the classes appearing in $P(y)$ do not appear in $Q(y)$ and vice versa.

\subsubsection{Estimating compositional dataset shift}
To estimate compositional dataset shift, we need to have a measure of the dissimilarity of $P(\x|y)$ between training and deployment.

The quantity $P(\x|y)$ is conditioned on the class, so the total distance must be a sum over the classes of a per-class distance. The input $\x$ is the image, and we can describe it in two ways:
\begin{description}
    \item[$\x$ is a vector of pixels] This is a natural choice, since it is exactly the input that the models take. 
    However, if we then compute the dissimilarity on a per-pixel basis, we are neglecting correlations between pixels which can be important in the definition of the features of a family of images. One way of taking into account some kind of correlations with pixels is to perform a PCA. However, the components of this vector are not interpretable, and as we will see this is a desirable property.
    \item[$\x$ is a vector of features] 
    An alternative solution is to extract a series of computer-vision descriptors of the image (App.~\ref{app:feature}), and then use those as $\x$. The shortcoming of this solution is that it is not clear which and how many different descriptors are necessary, that they are potentially collinear, and they act on different scales.\\
    To solve this issue, we resort to a relatively large number of descriptors, and apply a principal component analysis (PCA), in order to obtain an orthonormal space on which to calculate dissimilarities.\\
    When comparing the importance of different features, since we are not summing them with each other, we do not need to orthogonalize. Therefore, no PCA is needed, but we still need to standardize the features so that their scale does not artificially make some distances larger than others.
\end{description}

\paragraph*{Hellinger Distance}
In order to estimate shifts in $P(\x|y)$, we define and compare several dissimilarity measures. 
In the main paper, we only report the definition of the Hellinger distance~\cite{gonzalez-castro:13}, which we found to best correlate with the performance drops. In App.~\ref{app:dist} we also show results using the Wasserstein distance and the Kullback-Leibler divergence.

The Hellinger distance was already used in plankton classification~\cite{gonzalez:17,gonzalez:19}.
Given two normalized distributions $p$ and $q$ defined on a discrete one-dimensional support (with bins $i=1,\ldots,n_\mathrm{bins}$), the Hellinger distance is
\begin{equation}\label{eq:hdf}
    \HD^{f} = \frac{1}{\sqrt{2}} \sqrt{\sum_{i=1}^{n_\mathrm{bins}} \left(\sqrt{p_i}-\sqrt{q_i}\right)^2 }
\end{equation}
The superscript $f$ in Eq.~\eqref{eq:hdf} represents a single dimension of $\x$ (so a pixel, a  principal component, or a feature of the image). Since the values of dimension $f$ are min-max normalized, the distance ranges from 0 to 1.
The full distance between two distributions is obtained by averaging the Hellinger distance along all the components,
\begin{equation}\label{eq:df}
    \HD=\frac{1}{N_f}\sum_{f=1}^{N_f} \HD^{f}\,.
\end{equation}
Note that $\HD$ especially in the presence of small abundances per class, is dependent on the binning.
For this reason, unless specified otherwise, we restrict $\HD$ to those classes that have at least 10 images in  the OOD cells.

\subsection{Using the distances}\label{sec:using-dist}
By checking how much the OOD performance drops, we can use the dissimilarities defined in Sec.~\ref{sec:dist} to address several questions. This procedure expands on ideas initially presented in Ref.~\cite{gonzalez:17}, which showed that cross-validation folds with larger $\HD$ have a lower performance.
As described in the following, each question needs the distances to be calculated in a slightly different way.

\paragraph{Are the model performances directly related to dataset shift? (Sec.~\ref{sec:pxy-overall})}
Here, we want to estimate how much the $P(\x|y)$ changes, and then average over $y$. We therefore need to calculate the distances on a per-class basis, and then take the average over the classes. This corresponds to taking a macro-averaged distance.

Since $\x$ represents several features that could be mutually correlated, and we want an orthonormal basis, we perform a PCA, and select the number of principal components such that the amount of explained variance is greater than 95\%. There are large differences between the ranges of initial features, those features with larger ranges will dominate over those with smaller range and lead to biased result. Therefore, we standardize the features by subtracting the mean and dividing by the standard deviation, so that each feature contributes similarly to PCA. We then calculate the distance according to each principal component following Eqs.~\eqref{eq:hdf} and~\eqref{eq:df}.

\paragraph{Which are the image features which most influence the DS robustness? (Sec.~\ref{sec:pxy-features})}
When comparing different features, we do not need to perform the PCA, but it is important that we rescale different features in such a way that they are on the same scale (otherwise, the distances related to some features may artificially seem larger than others). Since the value of different features are in different scales, and because of the presence of outliers,
we apply some transformations on the original feature data before distance calculation. First, the feature values are standardized by subtracting the mean and dividing by the standard deviation to ensure they are all on a same scale. Then a logarithmic transformation is applied to reduce the impact of outliers,
\begin{equation}\label{eq:log}
    x' = \begin{cases}
    \log(1+x), & \text{if $x\geq0$}\\
    -\log(1-x), & \text{otherwise}.
  \end{cases}
\end{equation}
Since with this construction the components are not orthogonalized, collinear features will have similar values of $\HD$, but this is not a problem for the analyses presented in this paper (for example, it is not a problem if both height and area appear as important).

\paragraph{Which classes are harder/easier more robust under dataset shift? (Sec.~\ref{sec:pxy-classes})}
To estimate which classes suffer most from dataset shift we take the distance on a per-class basis. Also here we perform a PCA on the features.

\paragraph{How well will a classifier perform on a new unlabeled dataset? (Sec.~\ref{sec:pxy-unlabeled})}
If presented with a totally new, unlabeled dataset, we do not have knowledge on the classes which compose it, so we cannot estimate $P(\x|y)$. Therefore we directly estimate the distances on the distribution $P(\x)$, which corresponds to calculating a micro-averaged distance. Here we perform a PCA on the input features as well.

\subsection{Performance Sensitivity}\label{sec:sensitivity}
The bigger the performance drop as a function of the dissimilarity, the bigger we expect that a specific feature or class will be critical in the OOD robustness of the classification. We can therefore think in terms of a quantity
\begin{align}
\label{eq:qacc}
   Q_\mathrm{acc} &= \frac{A_\mathrm{train}-A_\mathrm{test}}{D(\text{train, test})}\,,\\
\label{eq:qf1}
   Q_\mathrm{F_1} &= \frac{{F_1}_\mathrm{train}-{F_1}_\mathrm{test}}{D(\text{train, test})}\,,
\end{align}
where $D(\text{train, test})$ is any of the distances we defined in Sec.~\ref{sec:dist}.
The larger $Q$, the larger the influence of dataset shift.
In the limit of small variations, $\frac{dA}{d\x}$ (or $\frac{dF_1}{d\x}$), this quantity is usually called a sensitivity in statistics~\cite{tortorelli:94} or a susceptibility in physics~\cite{huang:87}.

\subsection{Improving performances under dataset shift}
We test simple methods for countering dataset shift, based on model choice, model training, and interpretation of the results.

\subsubsection{RGB reweighting}\label{sec:rgb}
If we assume that color variations could have a strong effect in the OOD performances, it can make sense to make sure to impose that the intensity distributions along the overall color distributions do not vary when the input data changes. Therefore, we rescale the input data in such a way that it matches the RGB distributions of the training set (which is different from that of ImageNet, given that our images have a black background).
Operatively, the mean and standard deviation of RGB values of the whole training set are calculated. Then during training, validation and testing, the RGB channels of all images are standardized by the means and standard deviations.

\subsubsection{Targeted augmentations} 
Data augmentation can help increase the robustness of models against distribution shift~\cite{hendrycks:21b}. Therefore, we test whether we can address OOD performance drops with data augmentations that are generally not used within plankton classification.

To identify which kinds of augmentations have the best potential of improvement, we measure which features have the largest sensitivities $Q$, defined in Sec.~\ref{sec:sensitivity}, and perform augmentations based on these sensitivities.

\subsubsection{Ensembling} 
Besides choosing the models with the best robustness against dataset shift, we assess whether and how using ensemble models produces performances that are more robust under dataset shift. We compare the model performances of single model, arithmetic averaged ensemble model and geometric averaged ensemble model, as mentioned in Sec.~\ref{sec:ensemble}.

\subsubsection{Architecture} 
Different architectures may have different robustness to DS. In particular, it was argued that vision transformers are more robust than CNNs~\cite{paul:22}. We compare the performance of 8 selected architectures listed in Sec.~\ref{sec:models}.

\subsubsection{Test-time augmentation}\label{sec:tta}
Test-time augmentation (TTA) consists of performing test predictions on augmented versions of an image, in addition to the original one, and taking the average prediction. For example, if one takes an image, and the same image rotated of $180^\circ$, the model prediction should not change. Averaging the predictions over these two images will produced a more stable confidence vector where the noise is flattened out.

To our knowledge, TTA was proposed for the first time in Ref.~\cite{krizhevsky:12}, where they ensembled the prediction vector of five cropped patches (four corner patches and one center patch) while evaluating the model performance on test set. This gives the model a more robust prediction. However, most of the images in our plankton dataset have only one organism in the center surrounded by a dark background, so cropping does not introduce much diversity. Instead, we apply rotation during test time. The final prediction of an image is the average of the predictions on $k$ rotated versions of image, with equally spaced angles. 

\subsubsection{Quantification through adjusted counts}\label{sec:adjust_count}
Another solution is to keep the classifiers untouched, and reinterpret the population counts. This has the advantage of not requiring retraining models. 

Here, we use unsupervised quantification algorithms proposed in earlier plankton literature~\cite{gonzalez:17,gonzalez:19,orenstein:20b}.
The basic quantification algorithm is Classify and Count (CC), which estimate an abundance by summing up the number of counts that the classifier associates to that class:
\begin{equation}
    n_\mathrm{CC}(y) = TP(y)+FP(y)\,.
\end{equation}

Adjusted counts methods correct the CC classification by using knowledge of the ID test performance of the models.
Although these methods are generally only thought for distributional dataset shift (since the knowledge of the ID performance could become obsolete), we check whether within lake plankton classification they can also be helpful in a more generic setting.

The adjusted count method~\cite{forman:08}, adjusts the CC counts based on the expectation of how often false  positives are found:
\begin{equation}
    n_\mathrm{AC}(y) =  n_\mathrm{test} \frac{n_\mathrm{CC}(y)/n_\mathrm{test} - f(y)}{r(y)-f(y)}\,.
\end{equation}

Another method consists in using the classifier softmax\footnote{The we specify "softmax" because the confidence vectors should be normalized ($\sum_y c(y)=1$), which is a property of softmax.} confidences $c_i(y)$ as probabilities for an adjusted count~\cite{bella:10}. Following Ref.~\cite{gonzalez:19} and call it \textit{probabilistic (adjusted) count}. We have both a probabilistic classify and count (PCC) and a probabilistic adjusted count (PAC), respectively defined as
\begin{align}
    n_\mathrm{PCC}(y) =& \sum_{i=1}^{n_\mathrm{test}} c_i(y)\,,\\
    n_\mathrm{PAC}(y) =& n_\mathrm{test} \frac{n_\mathrm{PCC}(y)/n_\mathrm{test} - f^\mathrm{pa}(y)}{r^\mathrm{pa}(y)-f^\mathrm{pa}(y)}\,.
\end{align}
where $r^\mathrm{pa}(y)$ and $f^\mathrm{pa}(y)$ are the \textit{probability-average} recall and fall-out, defined through measurements on the training set,
\begin{align}
    r^\mathrm{pa}(y) =& \frac{\displaystyle\sum_{i\in y}^{n_\mathrm{train}(y)} c_i(y)}{n_\mathrm{train}(y)}\,, \\[2ex]
    f^\mathrm{pa}(y) =& \frac{\displaystyle\sum_{i\notin y}^{[n_\mathrm{train}-n_\mathrm{train}(y)]} c_i(y)}{\bigg[n_\mathrm{train}-n_\mathrm{train}(y)\bigg]}\,,
\end{align}
where $n_\mathrm{train}(y)$ is the number of training images belonging to class $y$, and $n_\mathrm{train}$ is the total number of training images.

In Ref.~\cite{gonzalez:19}, they see that quantification algorithms do not add much with respect to Classify and Count, unless the models have low performances. So we can think that they are normally not useful, except when the model starts failing.

We highlight that one can also use semi-supervised quantification methods. These involve periodic checks by a human expert manually verifying by how much the counts should be adjusted. These yielded good results within binary plankton classification~\cite{orenstein:20b}, but are out of the scope of this work.

\subsubsection{Abstention}
One can hypothesize that the classifier confidence correlates with the probability of a prediction being correct. In that case, one can decide to discard the low-confidence predictions, thus increasing the model precision in exchange for a decrease in recall.

Formally, we can decide to define an abstention threshold $\theta\in[0,1]$, and only accept predictions where the confidence on the predicted class is larger than $\theta$.
With $\theta=0$ there is no abstention and all images get classified. With $\theta=1$ all the images get discarded.

\section{Assessment}
\subsection{Identifying Performance Degradation}

\subsubsection{OOD performance degrades}
\begin{figure*}[t]
    \centering
    \includegraphics[width=\textwidth]{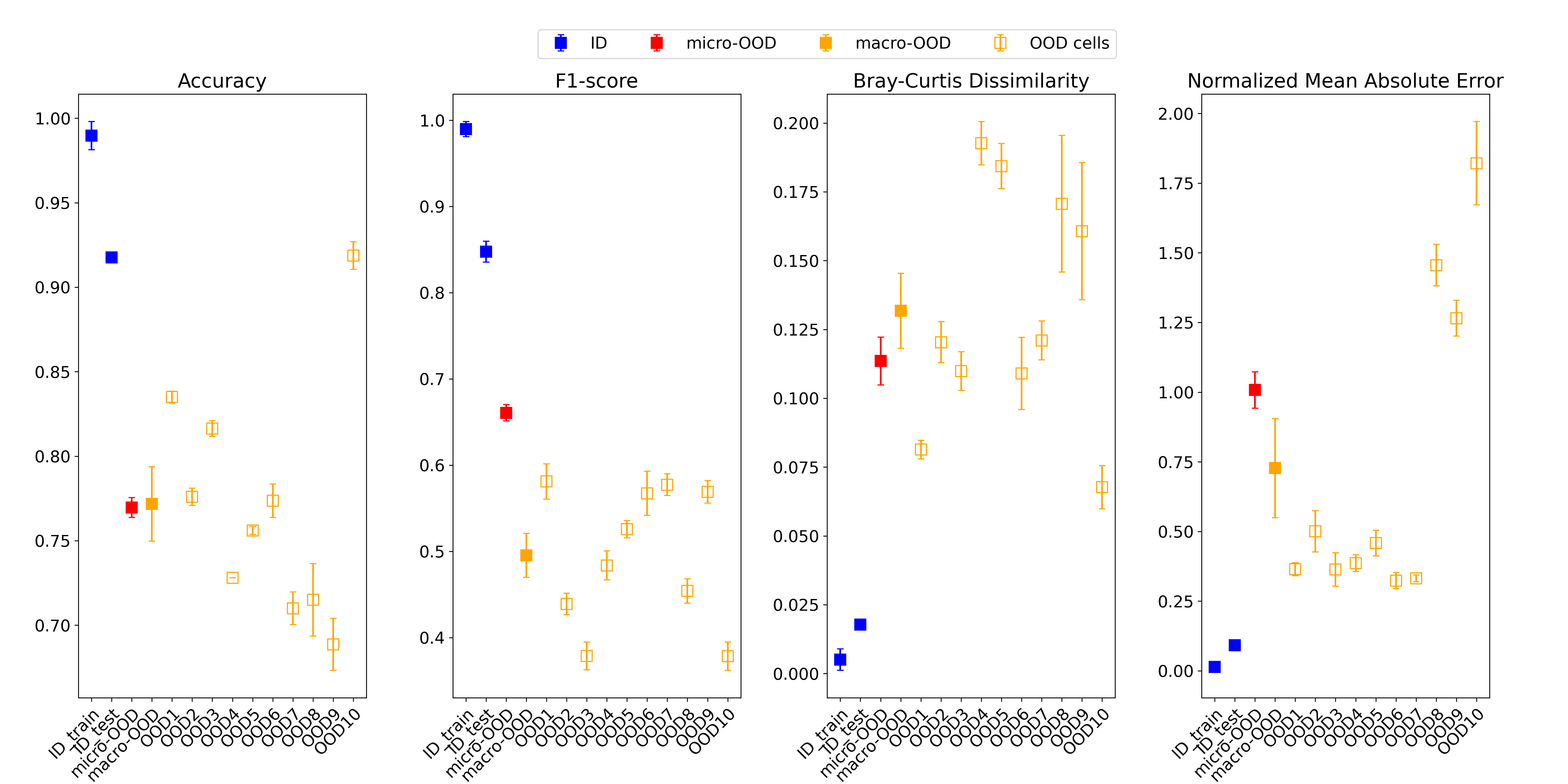}
    \caption{
    Four different performance metrics (accuracy, F1 score, Bray-Curtis dissimilarity, Normalized Mean Absolute Error) in the ID training and test set (blue), in the aggregated OOD data (red), and in the individual OOD test cells, obtained with a MobileNet (see Sec.~\ref{app:models}).
    }
    \label{fig:degrad}
\end{figure*}
We trained models following the procedures we described in Ref.~\cite{kyathanahally:21}, and compare their ID test performance with the OOD performance.
In Fig.~\ref{fig:degrad} we report four performance metrics for a MobileNet,\footnote{The same qualitative conclusions hold for all the architectures we trained, as we will show further in the appendix (Fig.~\ref{fig:single_basic-byOOD}).} 
two of which are based on single-image performance (accuracy and F1-score), and two of which are based on estimations of the abundance of each class (see Sec.~\ref{sec:metrics}).
Both if we use the individual test cells (macro-OOD, orange points), or if we merge all the OOD cells into a single one (micro-OOD, red), the ID test performance is drastically higher than the actual performance when faced with real-life situations.
We stress that the performance strongly varies among the OOD cells. This gives us an indication of the degree of fluctuations in the performances which should be expected: in different days of deployment, the classification performance can be quite different. Furthermore, there can be specific days where the performance is particularly different from usual. One example can be the OOD10 cell, where the accuracy is high, but the F1 score is low. This is because OOD10 comes from a day in which plankton community was dominated by \texttt{filament}. Since the model classifies \texttt{filament} well, the accuracy is high. The F1 score is however low, because the other taxonomic units have very few counts, and the F1 score is typically exaggeratedly low when the number of true positives is close to zero.

\begin{figure}[ht]
    \centering
    \includegraphics[width=\columnwidth]{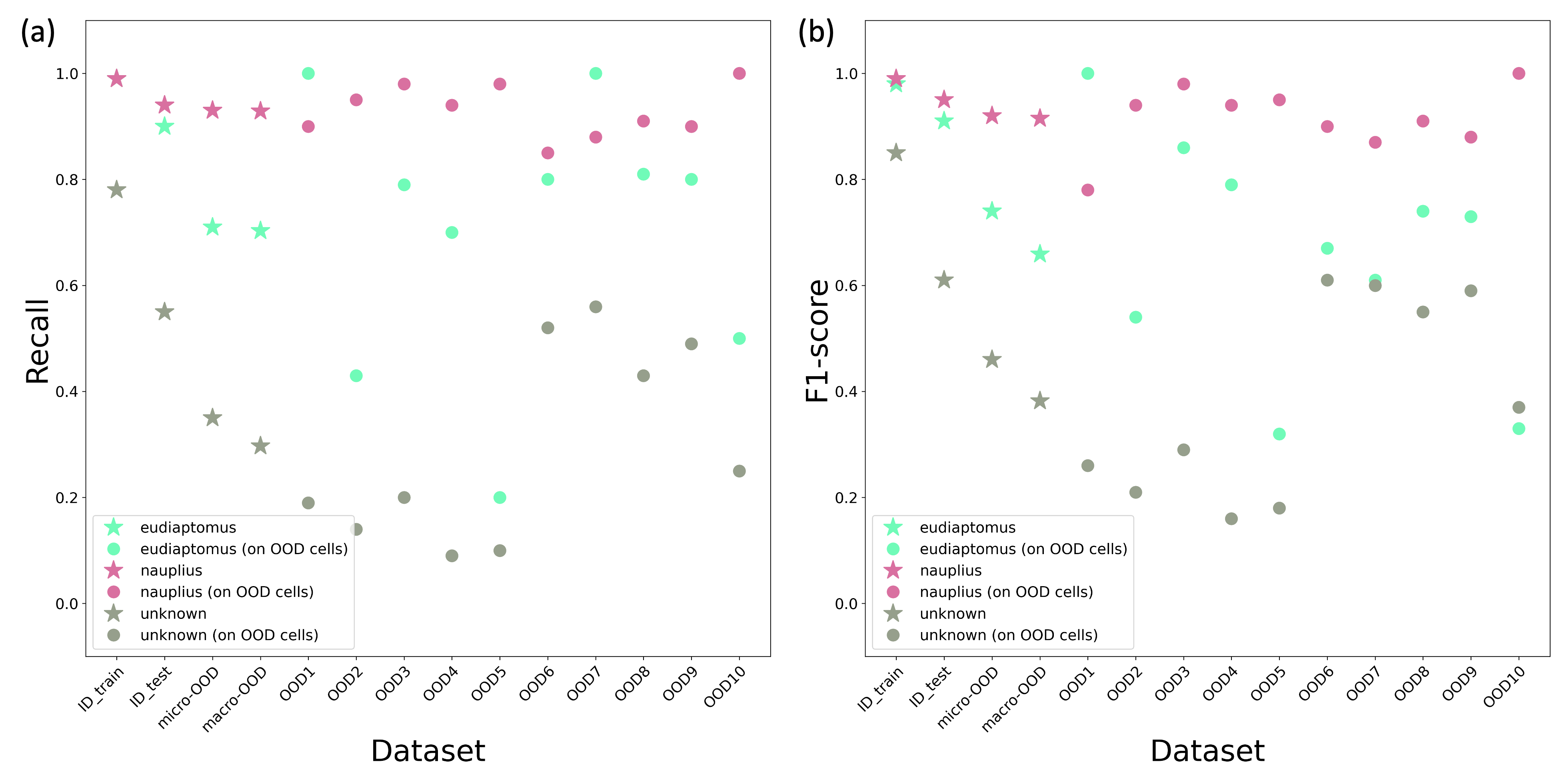}
    \caption{\textbf{(a)} Recall and \textbf{(b)} F1 score in the training set, ID test set, and OOD test sets, obtained by a MobileNet (see Sec.~\ref{app:models}) for three classes: \texttt{nauplius}, \texttt{eudiaptomus} and \texttt{unknown}. More classes are shown in App.~\ref{app:drop-per-class}.
    }
    \label{fig:degrad-3class}
\end{figure}

In Fig.~\ref{fig:degrad-3class}, we show that different classes suffer differently from OOD: while \texttt{nauplius} is practically unaffected, the classification of \texttt{eudiaptomus} deteriorates visibly on OOD data. The \texttt{unknown} class has the greatest OOD drop. This is expected, since it is a container class which gathers all sorts of disparate objects.\footnote{We highlight that the unknown class does not represent instances where the classifier abstained. They instead represent instances where the taxonomists were not able to assign a ground truth.}
In App.~\ref{app:drop-per-class} we show the performance drop for all the classes.

\begin{figure}[t]
    \centering
    \includegraphics[width=0.5\textwidth]{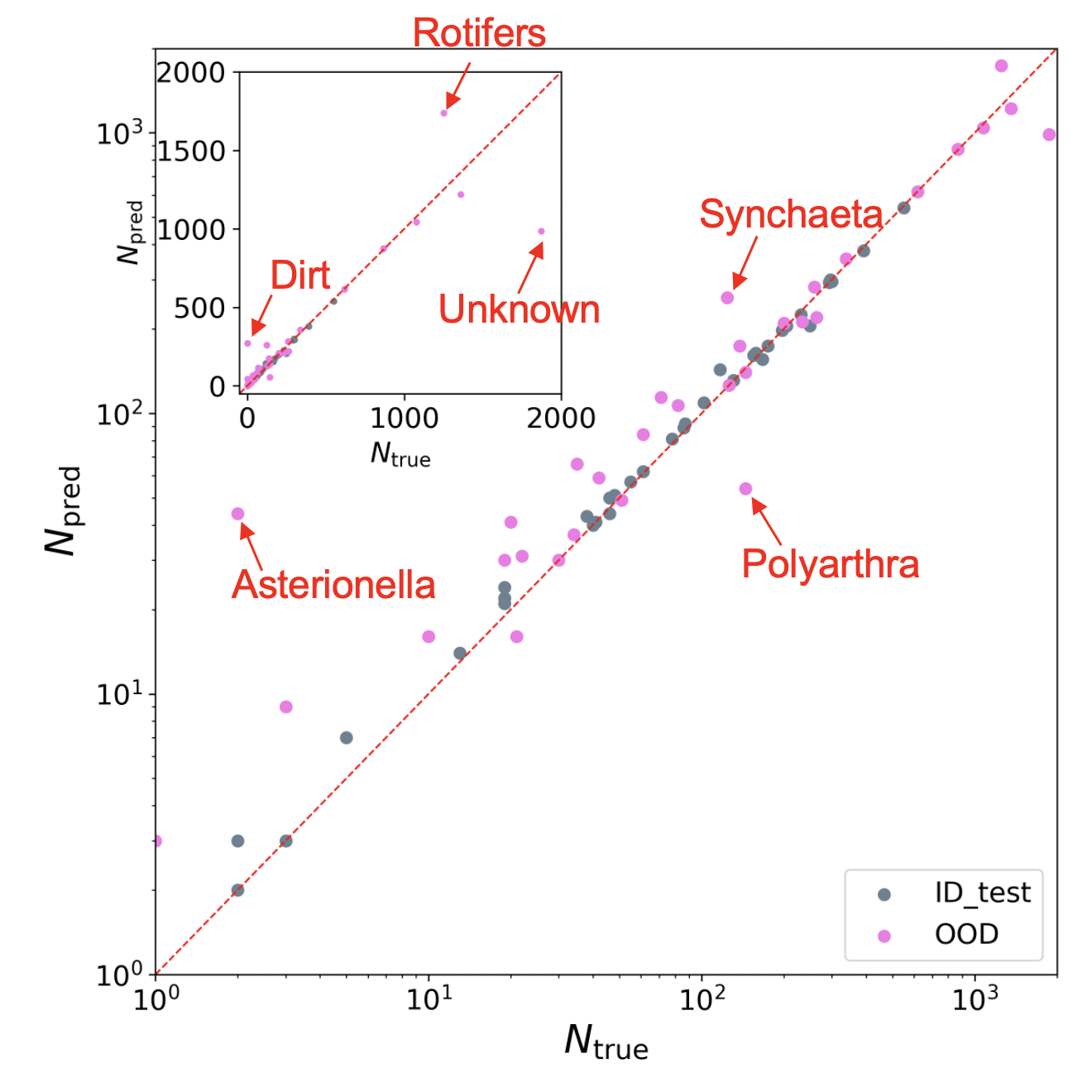}
    \caption{Scatter plot of
    Each point represents a single class, either in the ID test set (grey) or in the aggregated OOD set (pink).
    A perfect classification would fall on the 1:1 line (red line). The inset shows the same data as the main set, but on a linear scale.
    In App.~\ref{app:scatter-ood}, we show this same scatter plot for the individual OOD cells.
    }
    \label{fig:population_scatter}
\end{figure}

In Fig.~\ref{fig:population_scatter} we show a scatter plot of the predicted \textit{vs} true abundance in the ID test and in the overall OOD dataset. The ID points are concentrated around the 1:1 line, and the OOD predictions are further away. While the overall OOD performance is still by many standards acceptable, the abundances of some classes are wrongly estimated.\footnote{
The reader might be interested whether this problem can be overcome by aggregating similar classes together. In App.~\ref{app:aggregated} we show that, though the overall performances go up, a performance drop also persists if we group the classes describing coarser taxonomic units.
}

\subsection{Diagnosing performance degradation }

\subsubsection{Why is performance degrading? A first analysis}\label{sec:warmup}
\begin{figure}[t]
    \centering
    \includegraphics[width=\columnwidth]{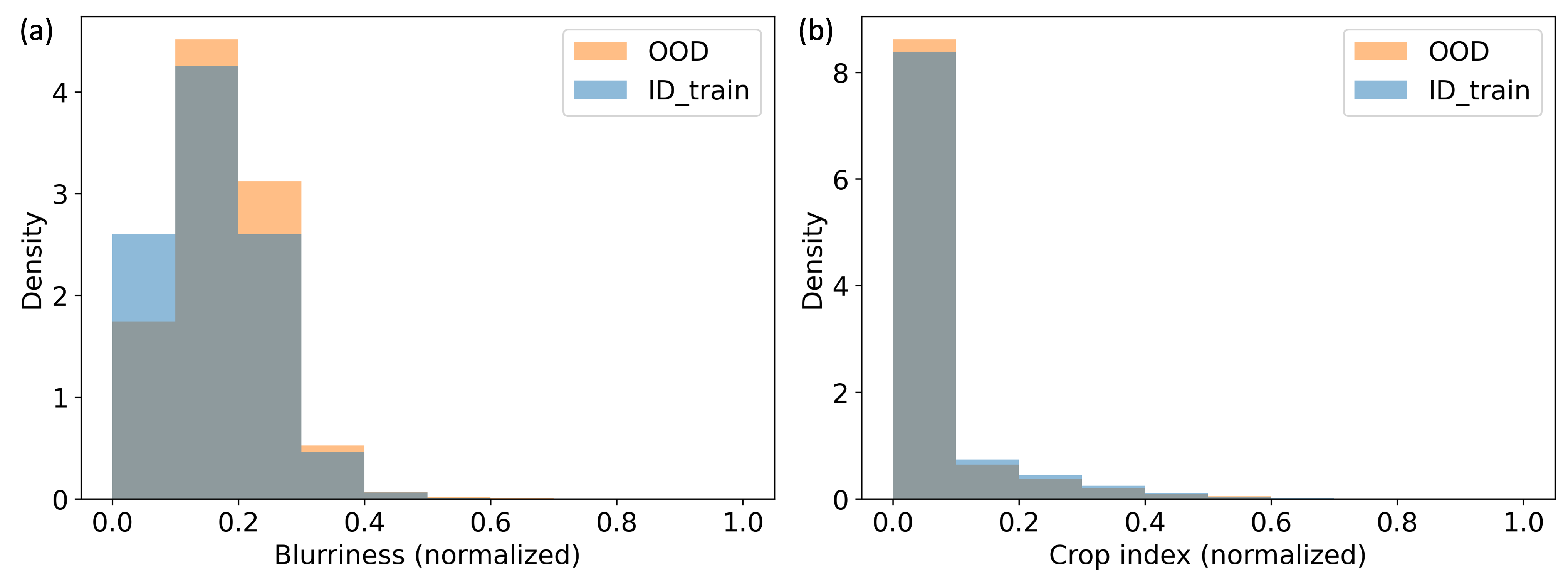}
    \caption{
    \textbf{(a)}: Distribution of the crop index in the ID and in the OOD. The gray part is the overlap of two distributions.
    \textbf{(b)}: Distribution of the blurriness in the ID and in the OOD. Crop index and blurriness are defined in Sec.~\ref{sec:feature}.}
    \label{fig:histo}
\end{figure}
As a start, we can try to explain this phenomenon with the most intuitive explanation possible: there could be more hard-to-classify images in the OOD cells than in the ID, due, for example, to sampling bias. The two most intuitive reasons that we can attribute to image hardness, are that the image was cropped (only a part of the organism appears in the picture), or it was blurry (the organism is out of focus).

In Fig.~\ref{fig:histo}, we compare the ID and OOD data according to these two quantities. In Fig.~\ref{fig:histo}a we see that the crop index (defined in Sec.~\ref{sec:feature}) is equally prevalent in the ID and OOD set, so cropping does not seem to be a driver of the OOD degradation. Similar conclusions are valid for the blurriness (Fig.~\ref{fig:histo}b), which is comparable in the two sets.
Therefore, in order to understand the OOD drop in performance, we need to resort to a more sophisticated analysis.

\begin{figure*}[t]
    \centering
    \includegraphics[width=\textwidth]{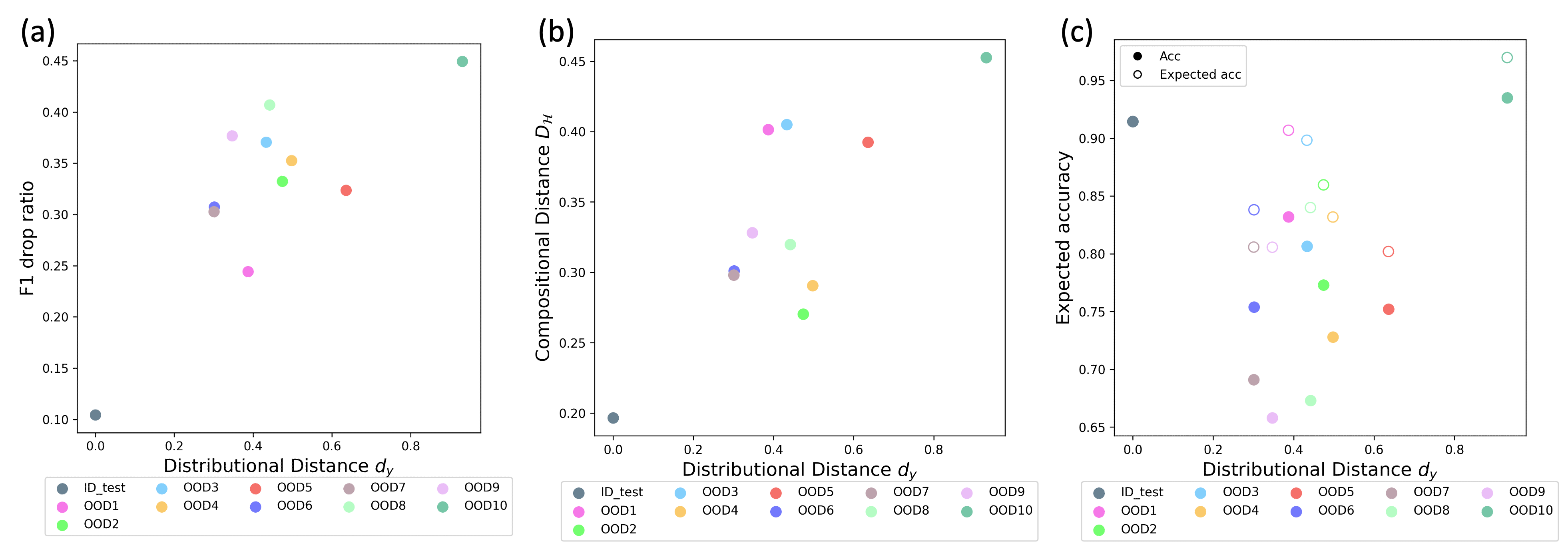}
    \caption{
    \textbf{(a)}: Plot of F1 drop vs $d_y$.
    \textbf{(b)}: Correlation between $d_y$ and $\HD$.
    \textbf{(c)}: Plot of expected accuracy vs $d_y$.
    }
    \label{fig:dy}
\end{figure*}

\subsubsection{Performance in terms of \texorpdfstring{P(y)}{$P(y)$}}
We now explore the dependence of the performance drop on distributional DS, \textit{i.e.} variations in the distribution of classes, $P(y)$, from ID to OOD.
From Fig.~\ref{fig:test-cells}, we already know that distributional DS is present. We quantify it through the distance $d_y$ [Eq.~\eqref{eq:distributional-DS}]. As we show in Fig.~\ref{fig:dy}a, the performance drop has some correlation with the amount of shift. 
However, we show in Fig.~\ref{fig:dy}b that distributional DS correlates with compositional DS, so the two sources of DS cannot be fully untangled. This is reasonable, since for example, the occurring of a bloom will change the relative abundances of taxa, and bring more diversity in plankton morphology.

We can however discard that the main driver of the performance drop is distributional shift. If in fact the performance degradation is driven by dataset shift alone, the OOD expected accuracy [Eq.~\eqref{eq:exp-acc}] and accuracy should match. Instead, the measured accuracy is systematically lower than the expected accuracy (Fig.~\ref{fig:dy}c), and the size of this drop is similar to the overall drop from ID to OOD.

\subsubsection{Performance in terms of $P(\x|y)$}

\paragraph{Overall performance}\label{sec:pxy-overall}
Since $P(y)$ cannot hold the whole responsibility of the degradation, we need to inspect $P(\x|y)$. 
In particular, we want to estimate how much the performance changes when varying $P(\x|y)$ from the training set to each of the (ID or OOD) test sets.
We estimate the change in $P(\x|y)$ from training to tests sets through the distances defined in Sec.~\ref{sec:dist} and App.~\ref{app:dist}. In the main text, we show results for the Hellinger distance $\HD$, and show analogous results for other kinds of distances in App.~\ref{app:dist}.

\begin{figure}[t]
    \centering
    \includegraphics[width=\columnwidth]{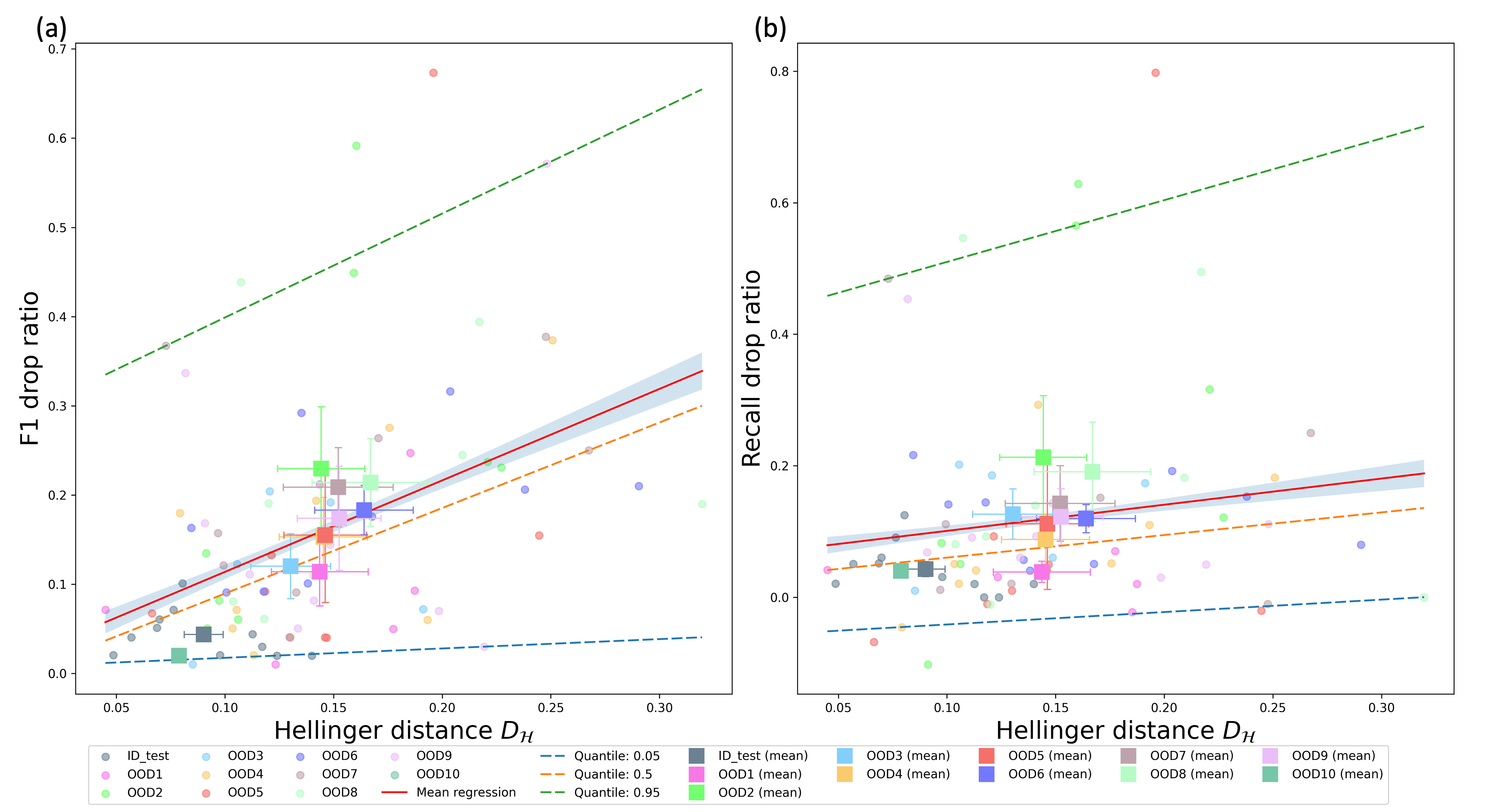}
    \caption{\textbf{(a)}: F1 drop [Eq.~\eqref{eq:drop}] as a function of the Hellinger distance from the training set [Eq.~\eqref{eq:df}]. Each dot indicates a specific class in an OOD cell. The dashed lines are the percentile 5, 50 and 95 regressions. The solid line is the mean regression.    
    Squares indicate a single OOD cell (error bars are the standard error among the different classes in each test cell). 
    \textbf{(b)}: Same, for the recall. Similar plots for other distances and architectures are shown in App.~\ref{app:dist} and App.~\ref{app:sensitivities}. 
    }
    \label{fig:cor_mean}
\end{figure}

In Fig.~\ref{fig:cor_mean} (squares), we show the F1 and accuracy drop in each test cell, as a function $\HD$. We find a visible correlation between performance drop and distance from the training set, with the F1 score correlating better than the accuracy (this is also found with other metrics, App.~\ref{app:dist}, as well as in Sec.~\ref{sec:pxy-unlabeled}). 
The dots represent instead each single class of each dataset. We show the mean regression, and the quantiles 5, 50 and 95 of the distribution. 
All increase steadily with $\HD$.\footnote{In particular, if $\HD$ is not a good indicator of the performance drop (as for example the KL divergence, App.~\ref{app:dist}), one could have that the highest quantiles even decrease with $\HD$, \textit{i.e.} a fraction of the classes become easier, but this is not the case.}
Note that regressions lines on the OOD cells (on the squares instead of the dots) would give an even higher slope and better correlation.

\paragraph{Distinguishing features}\label{sec:pxy-features}
One major question we wish to address about DS degradation is: \textit{what is changing in the images, that makes the performance go down?}. We already saw in Sec.~\ref{sec:warmup} that it is not the cropping nor the blurriness. 
We cannot connect the answer of this question to features such as water temperature or organism health, which cannot be easily inferred from the images. Therefore, we address the explanation of the OOD performance in terms of computer vision features, which can be extracted from the images (App.~\ref{app:feature}). To do so, we compute the Hellinger distance from the ID training set for each single feature, using Eq.~\eqref{eq:hdf}.\footnote{These features are not used for model training, but rather only to quantify the DS.} 

\begin{figure}[t]
    \centering
    \includegraphics[width=\columnwidth]{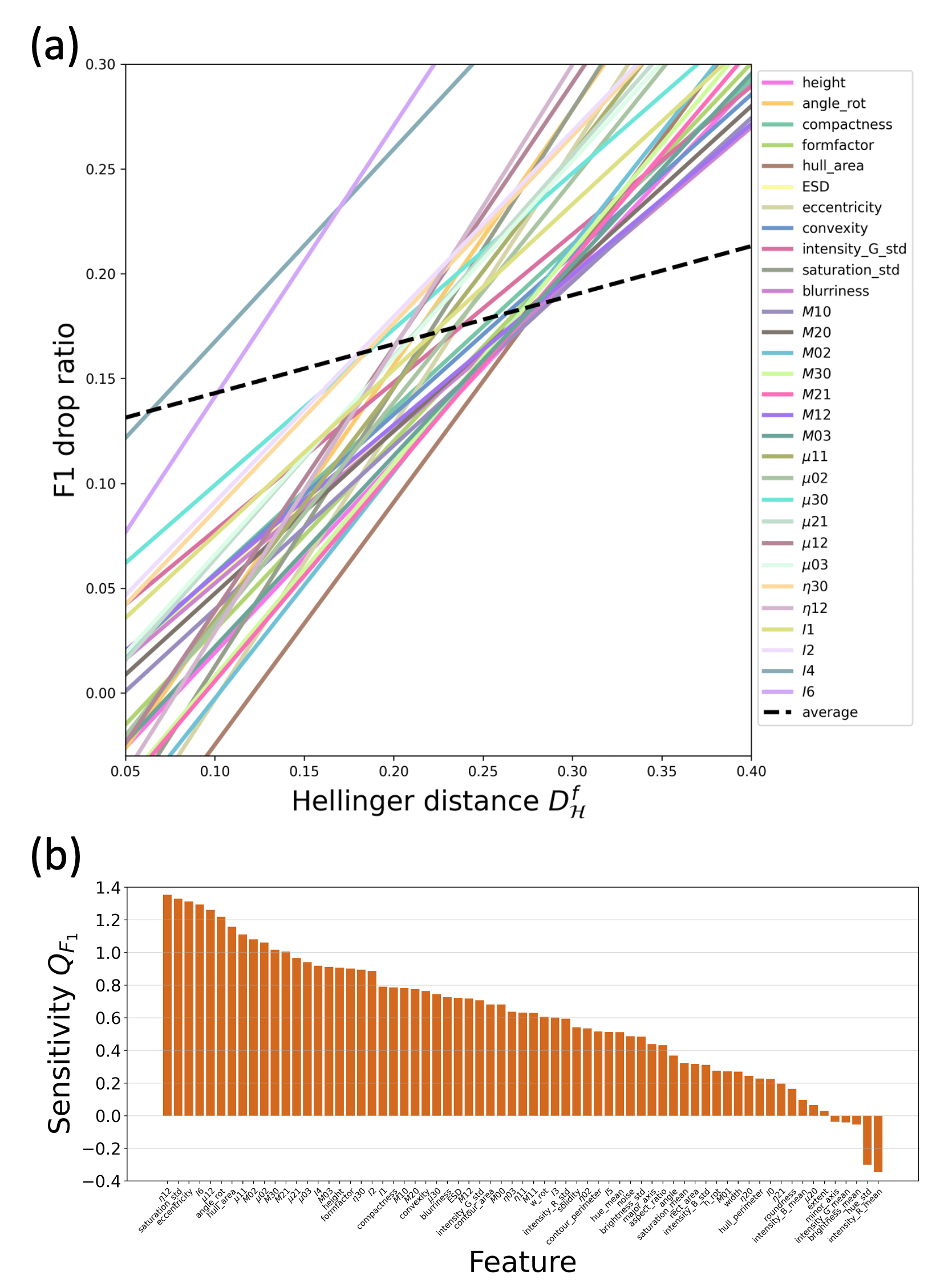}
    \caption{
    \textbf{(a)}: Performance drop as a function of $\HD^f$. For every feature $f$, we calculate regression lines, and show those related to the features with slopes $Q_{F_1}>0.7$.
    \textbf{(b)}: Measured value of $Q_{F_1}$, for each feature.
    All results in this figure are for a MobileNet model, trained with basic augmentations. 
    }
    \label{fig:cor_feature}
\end{figure}

We perform a simple linear regression of the F1 drop as a function of $\HD^f$. This tells us, to first order, how the performance varies while increasing the distance related to $f$. The slope of these regressions is the sensitivity $Q_{F_1}$ defined in Eq.~\eqref{eq:qf1}.
We plot those regression curves in Fig.~\ref{fig:cor_feature}a. For clarity reasons, we only plot the curves with $Q_{F_1}>0.7$, with an additional line representing the average of the remaining features.
We summarize the values of $Q_{F_1}$ in Fig.~\ref{fig:cor_feature}b, where we see that no specific feature stands out.
The most relevant features are (definitions are listed in App.~\ref{app:feature}) saturation\_std, intensity\_G\_std, eccentricity, compactness, formfactor, convexity, hull\_area, height, ESD, blurriness, angle\_rot,  image moments $M_{ij}$, $\mu_{ij}$, $\eta_{ij}$ and $I_i$.
These are related to color, shape, object size, blurriness and orientation, indicating that no single computer vision feature (or family of features) is dominating DS. 

\begin{figure}
    \centering
    \includegraphics[width=\columnwidth]{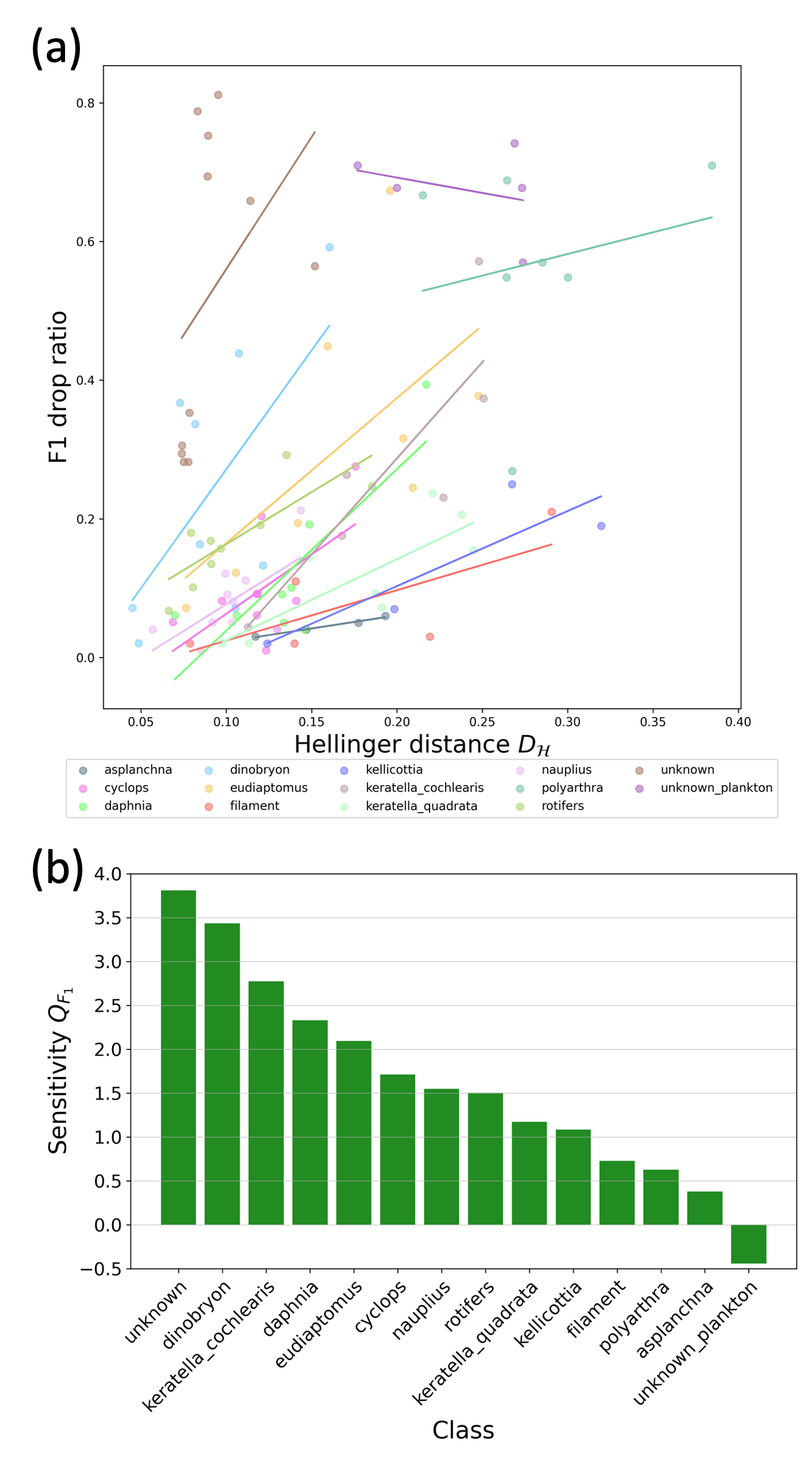}
    \caption{\textbf{(a):} Per-class F1 drop as a function of the Hellinger distance. 
    \textbf{(b):} Sensitivity $Q_{F_1}$ related to each class of those for which enough data is available. In App.~\ref{app:sensitivities} we show the same figure for the \bestmodel model. 
    }
    \label{fig:dist-classes}
\end{figure}

\paragraph{Distinguishing classes}\label{sec:pxy-classes}
Similarly as done for the single features, we can also check which classes are more sensitive to DS. However, we can only do this for those classes which occur at least 10 times in at least 4 OOD cells, to ensure reasonable correlations.  We show them in Fig.~\ref{fig:dist-classes}a.
We summarize the per-class sensitivities in Fig.~\ref{fig:dist-classes}b, where we can see that \texttt{unknown}, \texttt{dinobryon} and \texttt{keratella\_cochlearis} are the classes that is most sensitive to DS.
Note, however, that this result is model-dependent. In fact, we will show that we can produce a model where the sensitivity of \texttt{dinobryon} and \texttt{keratella\_cochlearis} are strongly reduced (App.~\ref{app:sensitivities}).

\paragraph{Unlabeled datasets}\label{sec:pxy-unlabeled}
For new data, we can give a rough estimate of the performance degradation. We do this by estimating the distance in the $P(\x)$ instead of the $P(\x|y)$, as described in Sec.~\ref{sec:pxy-unlabeled}. 
We find that this procedure seems to work reasonably well for the F1 score drop, but not for the accuracy. Details are in App.~\ref{app:per-unlabeled}.

\subsection{Cure}
With the objective of improving the OOD performance, we test seven different methods:
\begin{enumerate}
    \item RGB channel standardization (Sec.~\ref{sec:resRGB})
    \item Targeted augmentations  (Sec.~\ref{sec:resAUG})
    \item Architecture selection (Sec.~\ref{sec:resMODEL})
    \item Ensemble learning (Sec.~\ref{sec:resENS})
    \item Test-time augmentation (Sec.~\ref{sec:resTTA})
    \item Adjusted counts (Sec.~\ref{sec:resAC})
    \item Abstention    (Sec.~\ref{sec:resABS})
\end{enumerate}
In the following, we describe the effectiveness of each of those methods for our application case.

\subsubsection{RGB channel standardization}\label{sec:resRGB}

Since, as noted in Sec.~\ref{sec:pxy-features}, saturation\_std appears as one of the most sensitive features to DS. Therefore, we standardize the RGB color distributions, in such a way that those of the OOD cells match those of the training set, as described in Sec.~\ref{sec:rgb}. However, as we show in App.~\ref{app:rgb}, this procedure does not improve the performances.

\subsubsection{Targeted augmentations}\label{sec:resAUG}
\begin{figure*}[t]
    \centering
    \includegraphics[width=\textwidth]{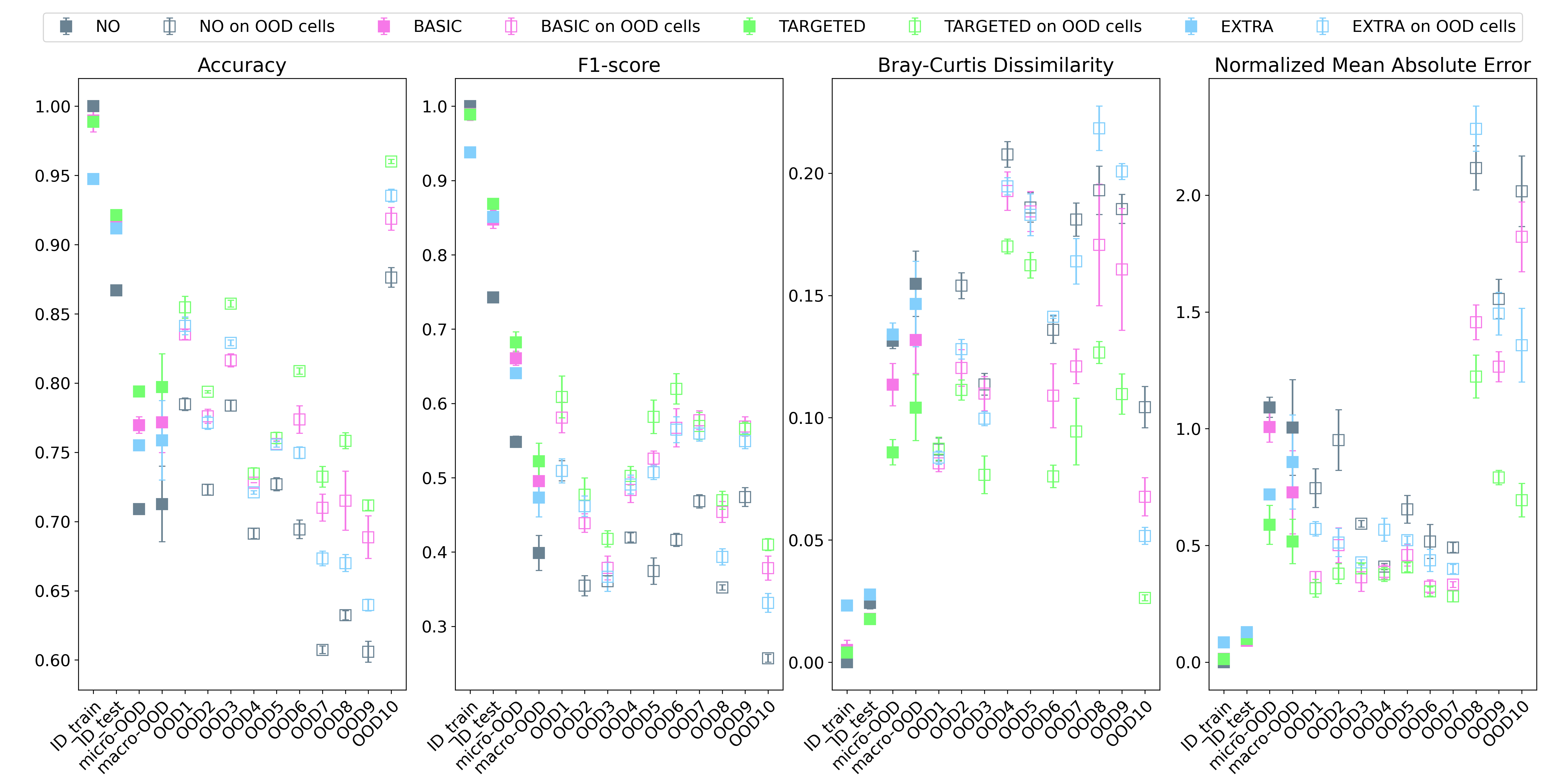}
    \caption{Comparison between the different augmentation schemes (Sec.~\ref{sec:augment}, App,~\ref{app:augment}), for a MobileNet model. Note that, although the ID performances are similar for any of the three applied augmentations, their OOD performances spread.
    }
    \label{fig:aug_ood}
\end{figure*}

We compare the four strategies of image augmentation in Fig.~\ref{fig:aug_ood}.
The targeted augmentation (the choosing of targeted augmentations is discussed in detail in App.~\ref{app:augment}) emerges as a clear winner (the green points have the best performance across all OOD cells). Note that, while the difference between different augmentation schemes is similar in the ID test set, it becomes larger in the OOD cells. We also highlight that training a model with targeted takes only slightly longer than with basic augmentation, but much less than the extra augmentation (Sec.~\ref{sec:augment}).

\subsubsection{Architecture selection}\label{sec:resMODEL}
We now compare the performance of different architectures: MobileNetV3~\cite{howard2019searching}, EfficientNet-B2~\cite{tan2019efficientnet}, EfficientNet-B7~\cite{tan2019efficientnet}, DenseNet~\cite{huang2017densely}, ViT~\cite{dosovitskiy2020image}, DeiT~\cite{touvron2021training}, Swin-T~\cite{liu2021swin} and BEiT~\cite{bao2021beit}. 
The specific choice of the architectures is motivated by previous literature on plankton classification: DenseNets were found to be well-performing in Ref.~\cite{lumini:19}, EfficientNets in Ref.~\cite{kyathanahally:21}, DeiTs in Ref.~\cite{kyathanahally:22}, Swin transformers in Ref.~\cite{yue:23}, and BEiTs in Ref.~\cite{maracani:23}.

\begin{figure}[t]
    \centering
    \includegraphics[width=\columnwidth]{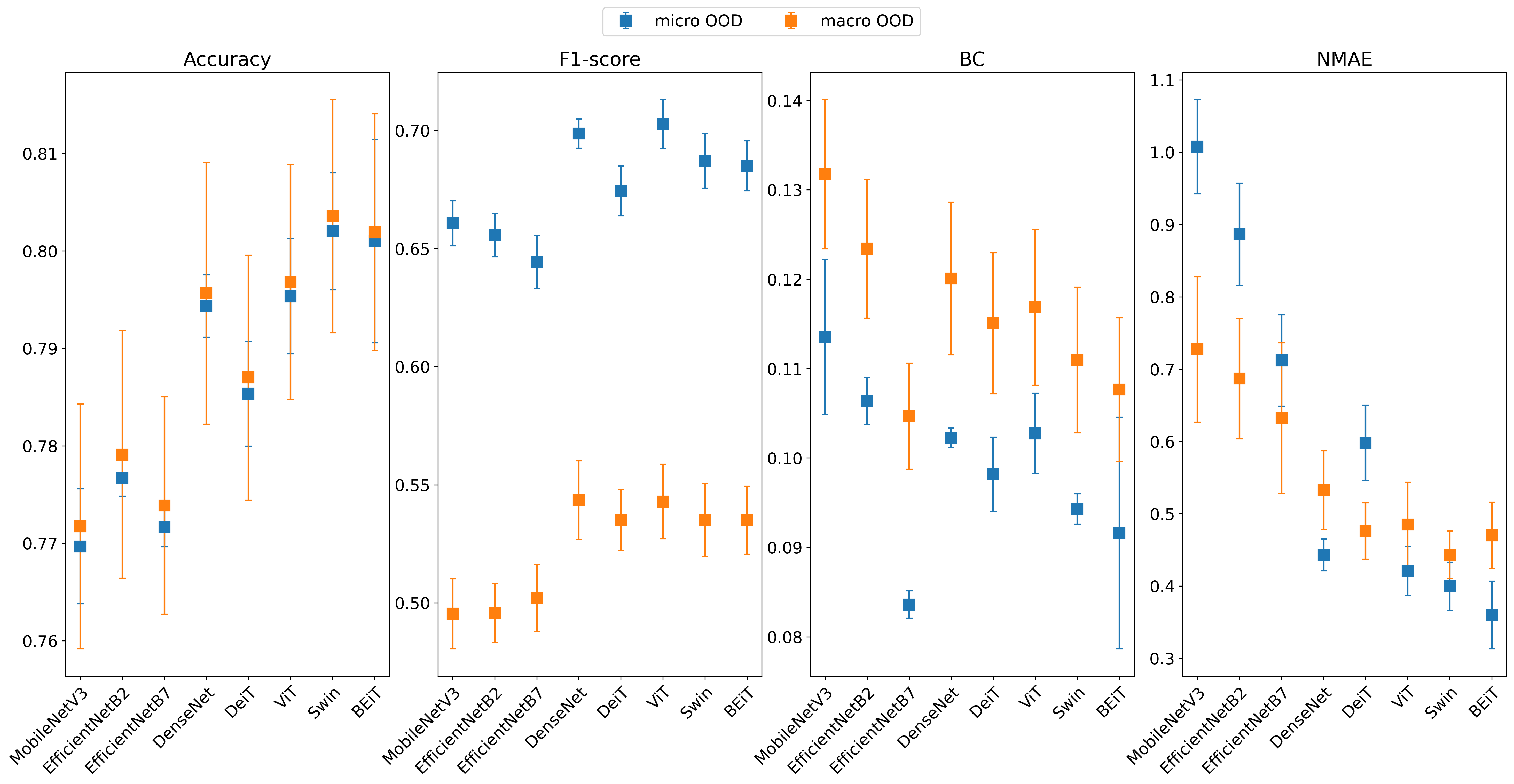}
    \caption{
    The classification results of basic models using eight architectures, tested on micro and macro OOD. 
    }
    \label{fig:single_basic-byArchitecture}
\end{figure}

We compare the OOD performance of those architectures in Fig.~\ref{fig:single_basic-byArchitecture}, and show the performances on the single OOD cells in App.~\ref{app:arch}.
The best CNN is EfficientNet-B7, which has high F1 score and BC, and low Accuracy and NMAE. The transformers perform generally better (except for the BC), and in particular the BEiT has the best performances in terms of population metrics (BC and NMAE). Therefore we select the BEiT architecture.\footnote{We show in App.~\ref{app:ensemble} that similar conclusions hold when using ensemble models.}

\subsubsection{Ensemble learning}\label{sec:resENS}
As shown several times before, using several models to increase the quality of the predictions can strongly increase the quality of the classification~\cite{lumini:19,kyathanahally:21,kyathanahally:22,maracani:23}. 

Here, we focus on ensembling over several replicates of the same architecture: we train the same model several times, with different initial weight configurations. This is more convenient than ensembling over different architectures, since the targeted augmentations are chosen on a per-model basis, and the overall performance is similar (App.~\ref{app:ensemble}). 

\begin{figure}
    \centering
    \includegraphics[width=\columnwidth]{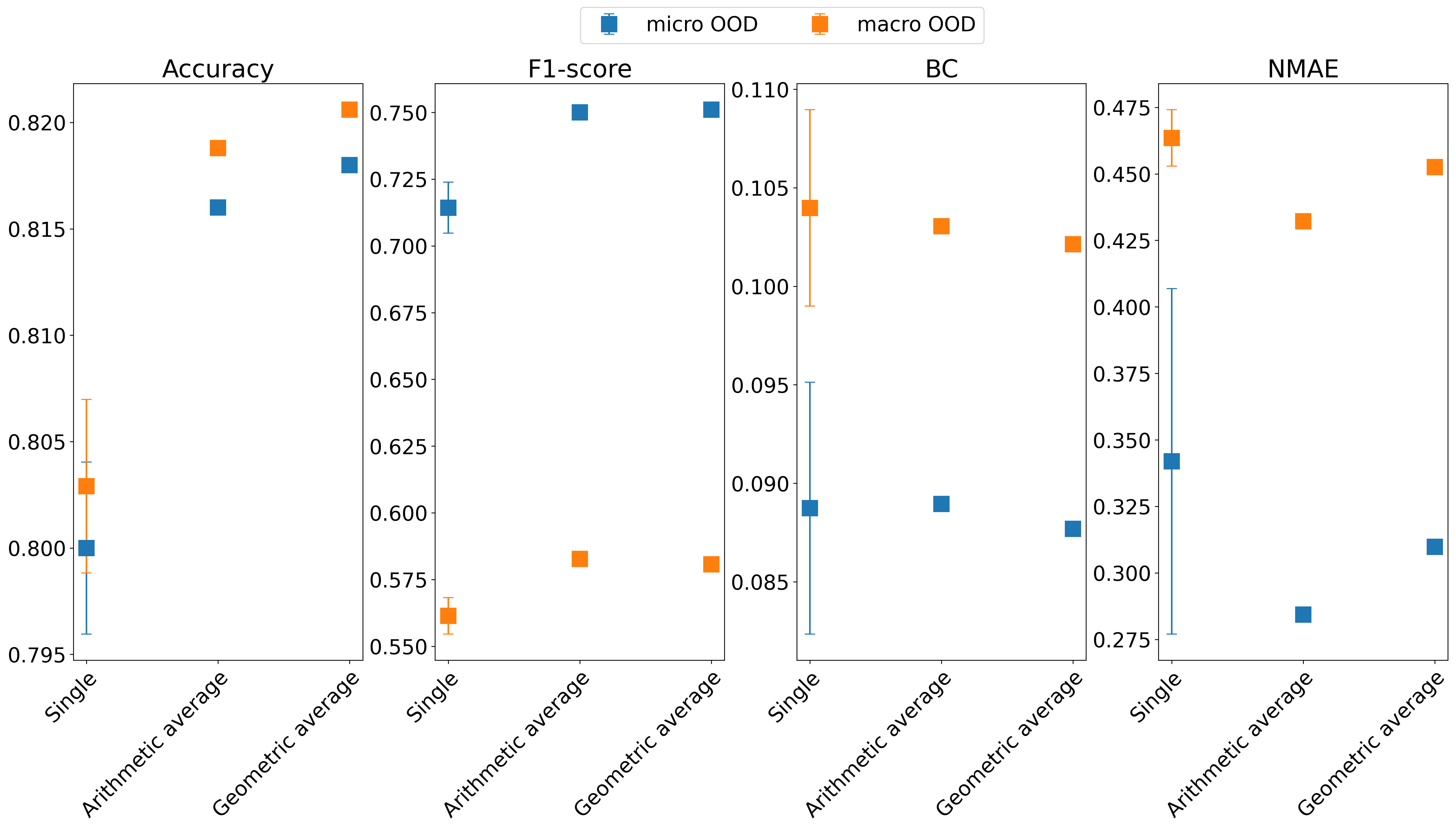}
    \caption{OOD performance without ensembling, and with two different kinds of ensembling, as defined in Sec.~\ref{sec:ensemble}.
    We report two OOD metrics: micro OOD is the performance over the aggregated OOD cell. Macro OOD is the average of the performances across test cells. }
    \label{fig:ensembling_ood}
\end{figure}

In Fig.~\ref{fig:ensembling_ood} we show the result of ensembling on the OOD cells through 3 single BEiT learners, trained with targeted augmentation. 
The improvement is sizable, regardless of the chosen performance metric, and geometric averaging seems slightly better. A reason can be that, with geometric averaging, if one of the learners gives a low score to a specific class, this will be more relevant than the possibly high scores given by the other learners.

\subsubsection{Test-time augmentation}\label{sec:resTTA}
We now test the efficacy of test-time augmentation (TTA), consisting in producing predictions on augmented versions of each test image, in addition to the images themselves.
With TTA, it is a good practice that the augmented images are true positives, and not deformed versions of the image that may never occur in reality. This reduces the number of possible augmentations available.
As we show in Fig.~\ref{fig:aug_ood}, despite the targeted data augmentation, models still suffer OOD degradation, and apparently one of the strongest factors is rotation (as shown in Fig.~\ref{fig:cor_feature}b). Therefore, we use rotated images for the TTA (see Sec.~\ref{sec:tta}).
\begin{figure}[t]
    \centering
    \includegraphics[width=\columnwidth]{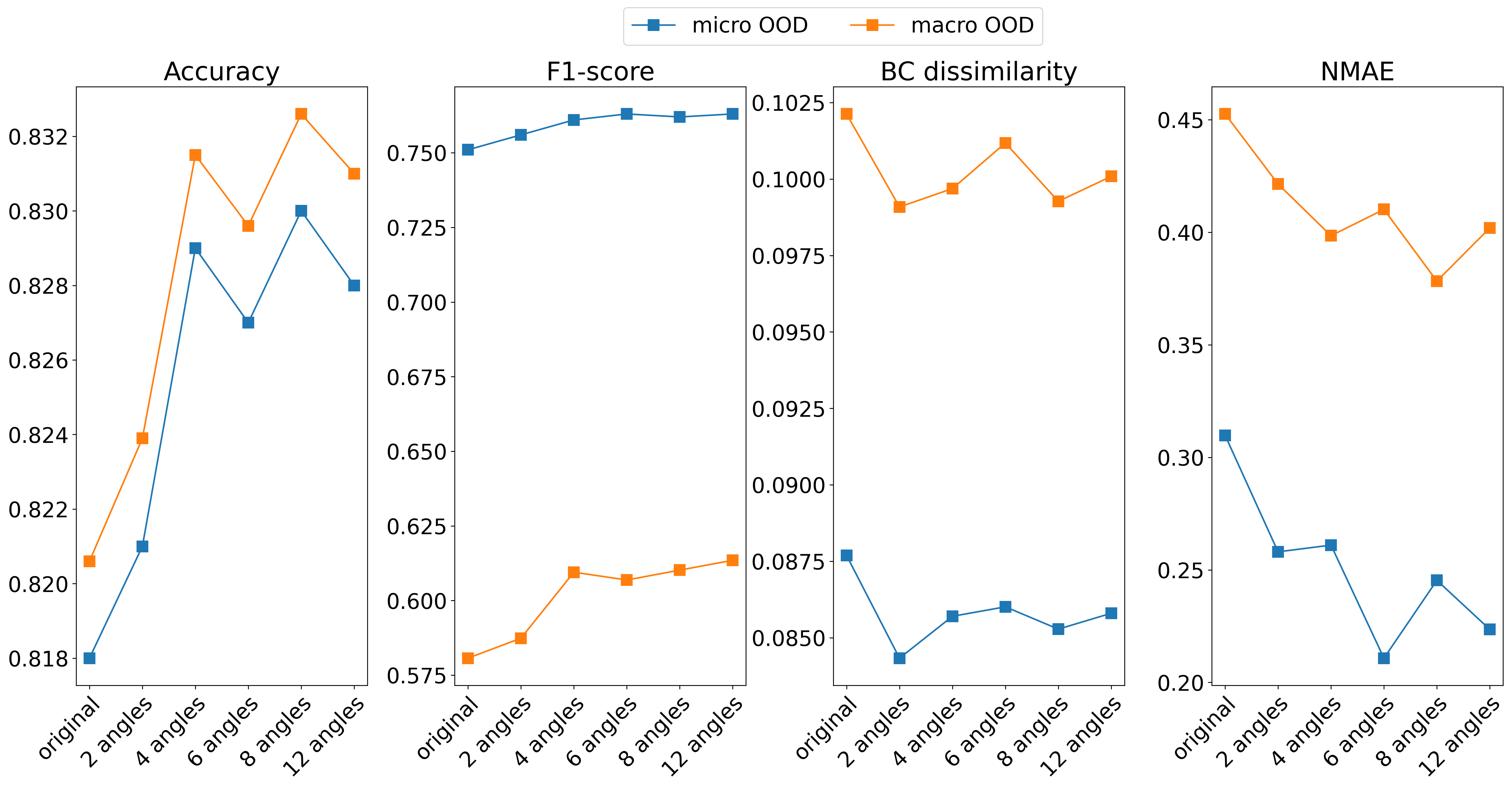}
    \caption{Performance on aggregated OOD dataset under various types of TTA, of a 3-BEiT geometric ensemble trained with targeted data augmentation.}
    \label{fig:tta}
\end{figure}
From Fig.~\ref{fig:tta}, we see that increasing the number of angles generally increases the performance.\footnote{In App.~\ref{app:tta} we also try flipping, but we find no advantage with respect to only doing rotations.}
We must however consider that the prediction time increases proportionally to the number of TTAs, so a lower number of angles is better in terms of wallclock time. Therefore, we choose to limit the number of angles $k$ to 4, since the gain is small after that.

\subsubsection{Adjusted counts}\label{sec:resAC}
\begin{figure}
    \centering
    \includegraphics[width=\columnwidth]{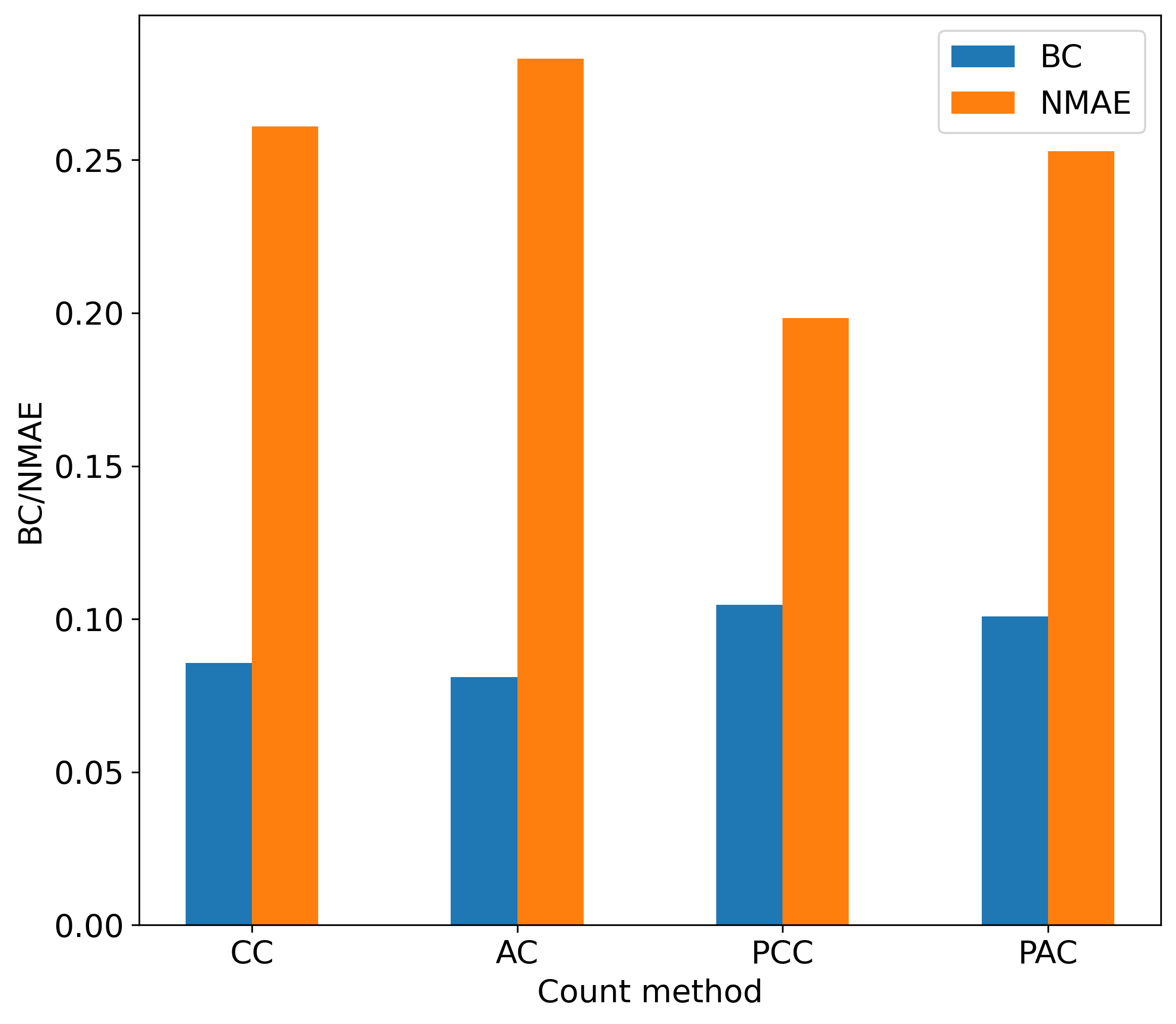}
    \caption{Comparison of different ways of performing adjusted counts. Methods of calculating these adjusted count are described in Sec.~\ref{sec:adjust_count}.}
    \label{fig:adjusted-counts}
\end{figure}

We also test the efficacy of methods that adjust the counts of the populations based on the ID metrics, since these methods seemed to have an effect in improving the classification performances in previous work~\cite{beijbom:15,gonzalez:19,orenstein:20b}. However, as we see from Fig.~\ref{fig:adjusted-counts}, none of the adjusted counts methods outperforms the vanilla classify and count procedure. In fact, although it is never the best, it is the second best both in terms of Bray-Curtis dissimilarity and of Normalized Mean Absolute Error. Therefore, we do not adjust counts in our final model.
Since these methods are thought to balance the effects of distributional (not compositional) DS, the failure of count adjusting is consistent with our previous observations that DS is a mainly compositional (not distributional) problem.

\subsubsection{Confidences and Abstention}\label{sec:resABS}
\begin{figure*}[t]
    \centering
    \includegraphics[width=\textwidth]{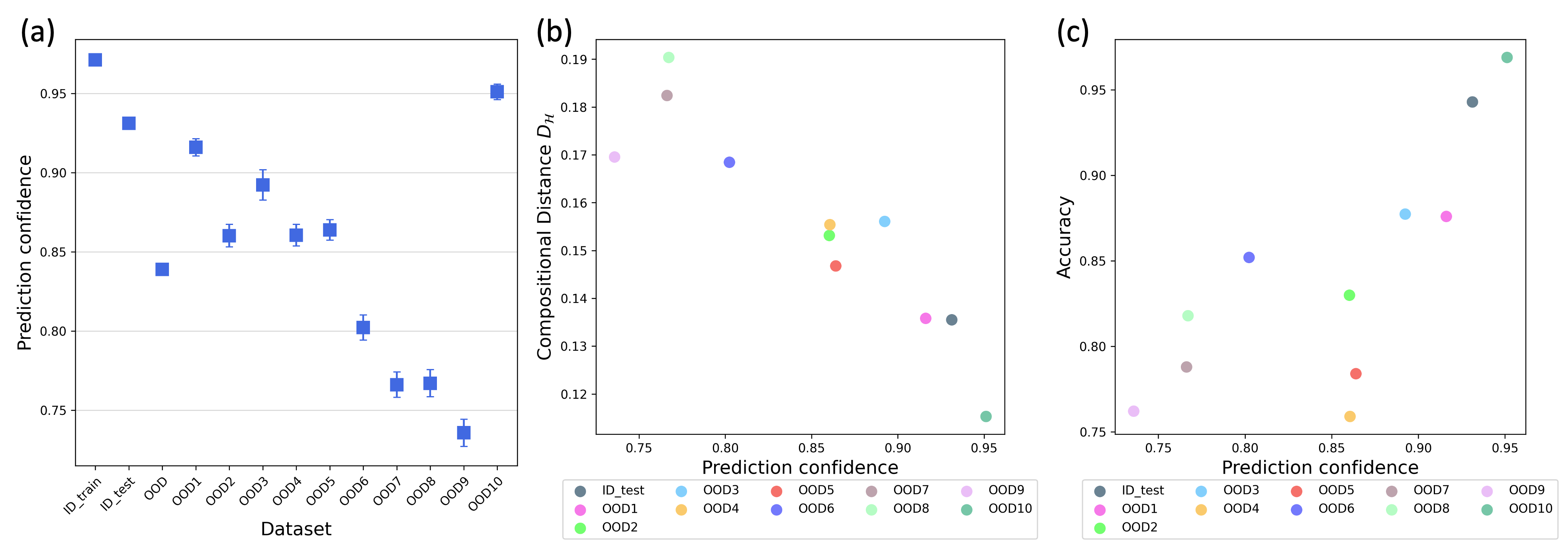}
    \caption{
    \textbf{(a)}: Prediction confidence of the \bestmodel model.
    \textbf{(b)}: Average confidence on each OOD cell, as a function of their Hellinger distance from the ID training set.
    \textbf{(c)}: Average accuracy in each OOD cell, as a function of the prediction confidence.
    }
    \label{fig:confidences}
\end{figure*}

We show in Fig.~\ref{fig:confidences}a, that the model confidences are high when ID, and lower when OOD. We see that the average confidences correlate well with $\HD$ (Fig.~\ref{fig:confidences}b), indicating that both quantities are good indicators of how out-of-distribution the OOD test cells are.

This can be exploited, as already previously done~\cite{ellen:15}, by utilizing abstention, which consists in asking the models to abstain from classifying when the confidence is too low. This increases the precision, and decreases the recall. Depending on the use case, it might be beneficial, when higher confidences correlate with better guesses, as it happens in our case (Fig.~\ref{fig:confidences}c). 
Since the performance under abstention depends on the chosen abstention threshold $\theta$, instead of providing performances at a given threshold, we show in Table.~\ref{tab:abstention} how precision and recall vary when changing this threshold. 
For the performances reported in the rest of the paper, we do not use abstention (this is equivalent to using a threshold $\theta=0$).

\begin{table}[t]
\centering
\caption{Abstention result, macro-averaged precision and recall with applying different confidence threshold during prediction on OOD.}
\begin{tabular}{cccc}
\toprule
\textbf{Threshold} & \textbf{\%classified} & \textbf{Precision} & \textbf{Recall} \\ \midrule
0 & 100 & 0.78 & 0.82 \\
0.1 & 99.72 & 0.78 & 0.81 \\
0.2 & 98.19 & 0.80 & 0.81 \\
0.3 & 95.49 & 0.81 & 0.79 \\
0.4 & 91.20 & 0.83 & 0.78 \\
0.5 & 86.44 & 0.85 & 0.76 \\
0.6 & 81.84 & 0.87 & 0.74 \\
0.7 & 77.06 & 0.87 & 0.72 \\
0.8 & 71.10 & 0.90 & 0.68 \\
0.9 & 63.41 & 0.92 & 0.64 \\
0.95 & 56.95 & 0.93 & 0.61 \\
0.99 & 45.87 & 0.98 & 0.56 \\ 
0.999 & 33.41 & 0.98 & 0.48 \\ \bottomrule
\end{tabular}
\label{tab:abstention}
\end{table}

\subsubsection{Ablation study}
We show several methods that can be used together, and show that they improve the OOD performance. The reader can still be interested in knowing whether each of these methods is useful, with and without the presence of the others. Therefore, we perform an ablation study on BEiT models, comparing the performances that are obtained with each combination of model improvements (Tab.~\ref{tab:ablation}). We can see that though the most crucial one is ensembling, best performances are obtained by mixing all the approaches. On the right part of the table, we compare how each methodology slows down the training and the inference times, since this needs to be taken into account as an additional factor, on top of sheer performance. 
\begin{table*}[t]
    \centering
    \caption{BEiT model performance metrics on different ablation settings.}
    \begin{tabular}{ccc|cccc|cc}
    \toprule
        \multirow{2}{*}{\textbf{Ensemble}} & \multirow{2}{*}{\textbf{TTA}} & \multirow{2}{*}{\textbf{Targeted aug.}} & \textbf{Accuracy} $\uparrow$ & \textbf{F1-score} $\uparrow$ & \textbf{BC} $\downarrow$ & \textbf{NMAE} $\downarrow$ & \multirow{2}{*}{\textbf{Training cost}} & \multirow{2}{*}{\textbf{Inference cost}} \\
        &  &  & \multicolumn{4}{c|}{(Micro-OOD/Macro-OOD)} &  & \\
    \midrule
         \ding{55} & \ding{55} & \ding{55} & 0.782/0.784 & 0.673/0.516 & 0.117/0.132 & 0.441/0.506 & $\mathcal{O}(1)$ & $\mathcal{O}(1)$ \\
         \ding{51} & \ding{55} & \ding{55} & 0.821/0.822 & 0.704/0.569 & \textbf{0.085}/\textbf{0.100} & \textbf{0.255}/0.413 & $\mathcal{O}(n)$ & $\mathcal{O}(n)$ \\
         \ding{55} & \ding{51} & \ding{55} & 0.795/0.797 & 0.702/0.557 & 0.111/0.127 & 0.363/0.474 & $\mathcal{O}(1)$ & $\mathcal{O}(m)$ \\
         \ding{55} & \ding{55} & \ding{51} & 0.793/0.796 & 0.716/0.556 & 0.100/0.113 & \textbf{0.256}/0.464 & $\mathcal{O}(1)$ & $\mathcal{O}(1)$ \\
         \ding{51} & \ding{51} & \ding{55} & 0.824/0.825 & 0.714/0.588 & \textbf{0.085}/0.102 & \textbf{0.258}/\textbf{0.399} & $\mathcal{O}(n)$ & $\mathcal{O}(nm)$ \\
         \ding{51} & \ding{55} & \ding{51} & 0.818/0.821 & 0.751/0.581 & 0.088/0.102 & 0.310/0.453 & $\mathcal{O}(n)$ & $\mathcal{O}(n)$ \\
         \ding{55} & \ding{51} & \ding{51} & 0.812/0.815 & 0.746/0.581 & 0.096/0.109 & 0.269/0.412 & $\mathcal{O}(1)$ & $\mathcal{O}(m)$ \\
         \ding{51} & \ding{51} & \ding{51} & \textbf{0.829}/\textbf{0.832} & \textbf{0.761}/\textbf{0.610} & \textbf{0.086}/\textbf{0.100} & \textbf{0.261}/\textbf{0.398} & $\mathcal{O}(n)$ & $\mathcal{O}(nm)$ \\
    \bottomrule
    \end{tabular}
    \label{tab:ablation}
\end{table*}

\subsection{The final model}
\begin{figure}
    \centering
    \includegraphics[width=\columnwidth]{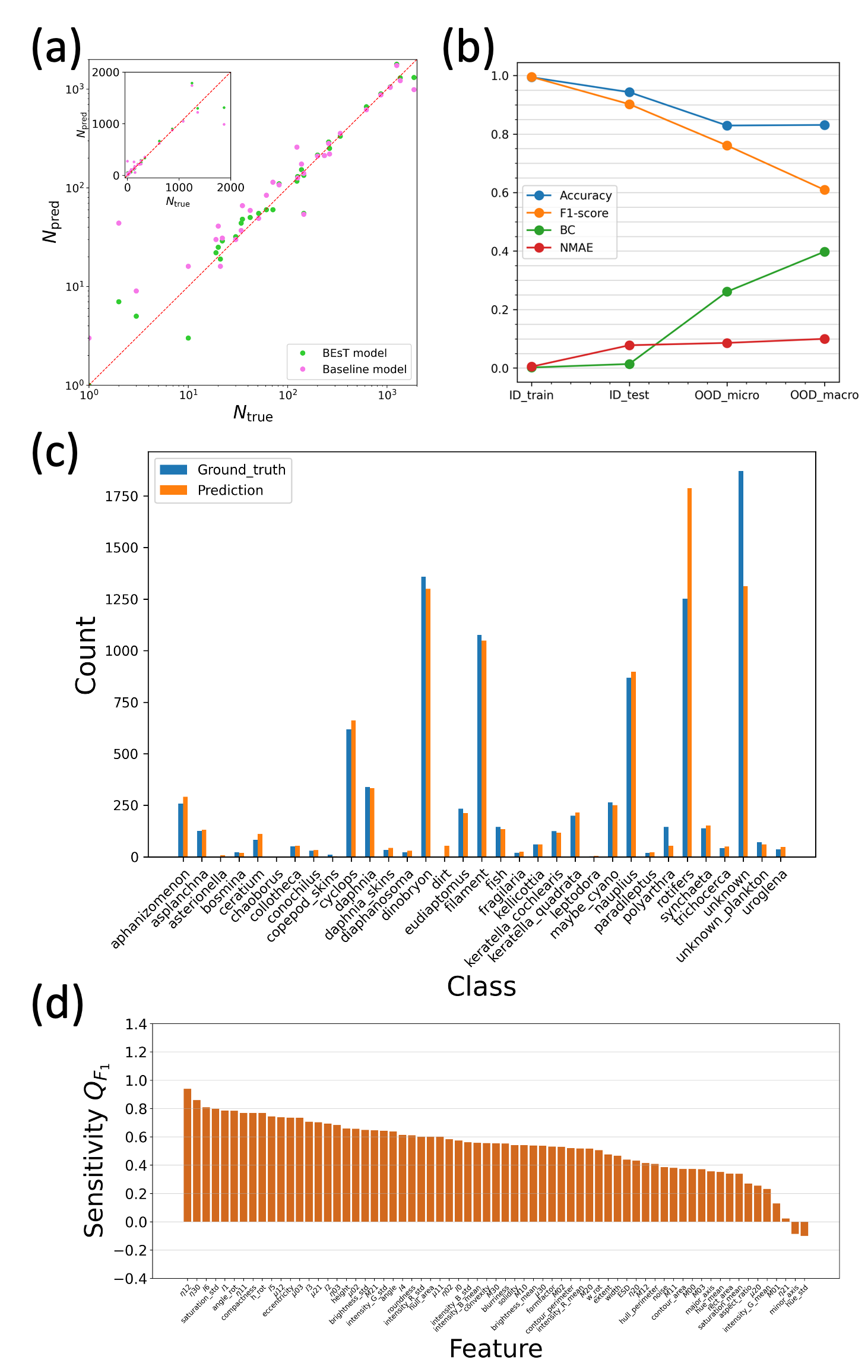}
    \caption{
    \textbf{(a)}: Scatter plot of true \vs predicted abundance, for the baseline model, and the \bestmodel model.
    \textbf{(b)}:     
    Train, test, micro-OOD and macro-OOD performances of the \bestmodel model. 
    \textbf{(c)}:     
    Bar plot showing the true and the predicted plankton abundances, for the \bestmodel model, on the aggregated OOD data.
    \textbf{(d)}: 
    Sensitivities $Q_{F_1}$ for the \bestmodel model. 
    }
    \label{fig:final}
\end{figure}

In Fig.~\ref{fig:final}a we show a scatter plot of the true \vs predicted OOD counts, for each population. We show two models: the baseline model and the final selected model, which we call \bestmodel model. The improvement is steady across more or less all the classes. Fig.~\ref{fig:final}c displays how well the final model performs on OOD data. 

The only two classes with a large number of misclassified images are \texttt{rotifers} and \texttt{unknown}, which get mixed one with the other. These are container classes, which gather in themselves a large number of different kinds of objects. One way of increasing this performance would be to distinguish different kinds of subclasses among them. Since, however, one of these classes expresses objects that were not identifiable by the taxonomists, there is a possibility that the \texttt{unknown} images classified as \texttt{rotifers} are actually rotifers.

In Fig.~\ref{fig:final}b, we show the performances of our final model, both ID and OOD.
The reader might argue that, while it is nice that the new performances are better than the baseline ones, the baseline performances are not that bad either. However, what we are testing here, is the solidity of our models to unknown changes in the future. The new models are also more robust to DS, as is demonstrated by the smaller sensitivities depicted in Fig.~\ref{fig:final}d and in App.~\ref{app:sensitivities}.
In fact, the performances that we are reporting are obtained on the same the OOD cells that were used to perform the model selection. Therefore, these are validation and not test performances. True performances should be given on new OOD data, so models with a lower sensitivity give us higher trust that they will be robust to varying conditions. 

\section{Discussion}
\paragraph{Summary}
Deploying classifiers without accounting for DS has the risk of providing highly deceptive results. Here, we show that DS is indeed a problem with plankton classification, we display a formalism to describe it, and a procedure to address it.

\begin{figure}
    \centering
    \includegraphics[width=\columnwidth]{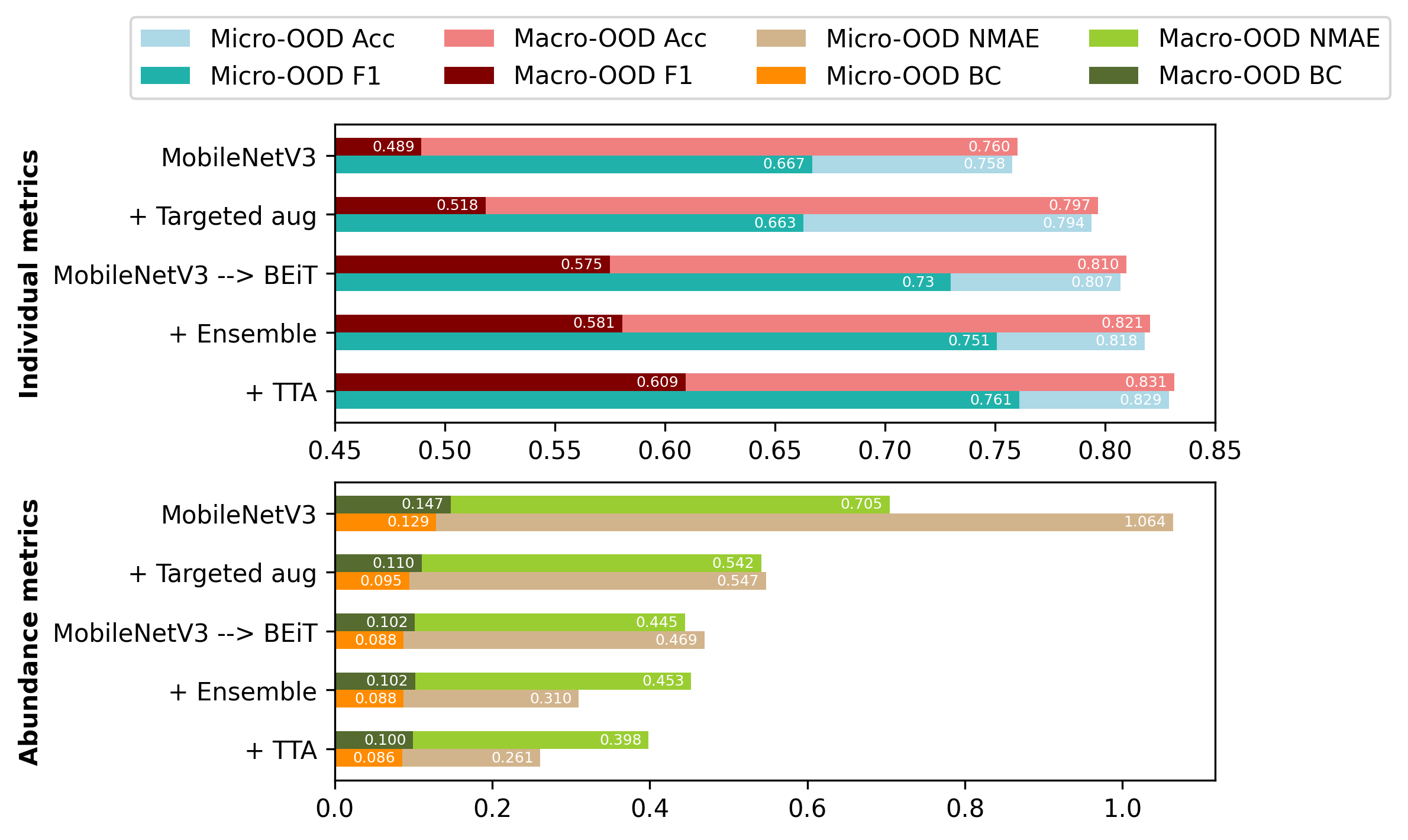}
    \caption{OOD performances after each step of model improvement. The reader interested in these steps applied in a different order can consult Tab.~\ref{tab:ablation}.}
    \label{fig:process}
\end{figure}

In particular, we address classification of lake plankton from the DSPC installed in lake Greifensee in Switzerland, focusing on dataset shift (DS). Our main aim is to assess whether DS can be a hindrance, and to find a classifier which is robust to DS. 
We see that the dataset shift is present, that it negatively influences model performances, and that compositional dataset shift is driving the performance decrease.
We then study the sensitivity of our models to DS. 
This depends on the specific architectures and how they are trained. 
The model performance under DS is mostly affected by color, rotation and shape features and less, for example, by noise in the image. Due to the imbalance of class abundances, we cannot safely assess the sensitivity of all classes. Among those with enough available data, the most sensitive to DS are \texttt{dinobryon}, \texttt{keratella\_cochlearis} and \texttt{unknown}. 
Finally, we evaluate a series of methods which can improve the model robustness to DS. Some of those, such as adjusted counting and RGB channel standardization, do not work, while others, such as ensembling and test-time augmentation, increase robustness. In Fig.~\ref{fig:process}, we summarize the steps that we took to get our best model, in terms of OOD performance.

\paragraph{Other possible strategies to address dataset shift}
One way to increase OOD performance is to increase ID performance.
In fact, there is evidence that models that have better in-domain performance will also have a better OOD performance~\cite{recht:18,recht:19}. However, it is important to keep in mind that the ID performance is not representative of the true performance of the model, and one should always have an estimate of the OOD performance.

Beyond those that we already tested, there are many other ways that one could address DS in our plankton classification problem. 
For example, one could use Domain Adaptation~\cite{wang:18b,patel:15} to infer the final $P(\x,y)$ and consequently reweigh the loss function.\footnote{Software for this already exists, \textit{e.g.} \url{https://domainadaptation.org}}
This is doable in our case, because $P(\x,y)=P(y|\x)P(\x)$ and, since one potentially has access to a large number of unlabeled images (all the images that need to be classified), we have access to the $P(\x)$.

More related to the results of this specific work, we see that rotations still play an important role, while translations are marginal, since our images are usually centered. Rotation-invariant CNNs might therefore be of help~\cite{mo:24}. Also, we find that the main source of OOD error in our tests is that the \texttt{rotifer} and the \texttt{unknown} classes get mixed. Since these classes include many kinds of different objects, robustness would likely increase if one was able to create reasonable sub-classes out of them. For example, unknown with some specific shapes and unknown with some specific colors. 

\paragraph{How serious is DS?}
Although we find that DS is present, and it strongly affects accuracy and F1 scores, even with the worse models it does not seem that the overall populations are so strikingly different. One could therefore be tempted, in future studies, from dismissing an analysis of model performance under DS.
We highlight, however, that we are working in a setting where the sources of DS are minimized (no temporal dependence, no varying external conditions, stable lake environment), so we expect that in generic plankton classification settings OOD will be larger than the one we find.
Ultimately, in order to really see whether DS shift is present, and to assess the utility of different architectures addressing it, one should compare the effect of different time series on the data that is generated through these classifiers. For example, one should compare the time series created by the two different models, and check whether they are quantitatively similar, through the measurement of autocorrelation functions and related quantities.

\paragraph{On the size of the OOD cells}
The size of our OOD cells meets a trade-off between data availability, and expert work. 
OOD cells of size 1000, with 35 classes, do not ensure a fully representative distribution of the data, but creating extremely large and numerous OOD cells is beyond the scope of this work. This reflects on an increased measurement uncertainty on Hellinger distances, performances, and sensitivities, which is exacerbated with rare classes.
However, when engaging in a large plankton classification campaign, one must likely regularly manually check that the classification is stable and to perform active learning~\cite{bochinski:19}. This ensures the availability of increasing amounts of data that can be used as OOD cells, thus providing progressively more interpretable results.

\paragraph{On data leakage}
The procedure we propose performs model selection based on the OOD cells. Since the OOD cells are used for model selection, they should not be used to report performances~\cite{kapoor:22}. 
In future work, we will retrain our selected model on a combination of all ID and OOD datasets, and deploy it on a very large amount of unlabeled data, to construct plankton abundance time series. We will test the classifications on random days, and these will be used as an assessment of the performance.

\appendix

\section{Related Work}\label{app:related-work}
Here, we provide a brief summary of recent literature related to DS in the context of plankton monitoring.\\[1ex]

Ref.~\cite{beijbom:15} studies DS with an AlexNet on the WHOI dataset~\cite{orenstein:15}, which contains millions of annotated black \& white images from a Flow Cytobot deployed in a sea environment.
They build a training set plus 21 test cells, which are small datasets used to assess the OODG, from where they clearly show the presence of class-distribution shift. They conclude that domain adaptation methods for quantification outperform random sampling.

Ref.~\cite{gonzalez:17} poses emphasis on how to test classifiers in the presence of DS, since with DS the test performance uses significance.
Instead of testing on hold-out data as done in Ref.~\cite{beijbom:15}, Ref.~\cite{gonzalez:17} proposes to use a cross-validation (CV) scheme, where each of 60 CV folds corresponds to a day of sampling. However, CV is viable because they work with Support Vector Machines and Random Forests, but it is generally not viable with deep neural networks. Further, the features that we use for calculating Hellinger distances are also used as input features, whereas in our case we use pixel values as an input for the models, and features for calculating Hellinger distances.
Similarly to us, they assess dissimilarity among datasets through the Hellinger Distance $\HD$, and find a weak correlation between accuracy and $\HD$.

Ref.~\cite{orenstein:20b} studies corrections to quantification in the context of binary classification of diatom chains, proposing to manually assess the performance on the target domain, and to use this information to correct the quantification. However, while viable for binary classification, it is not clear whether this would also be efficient in the multi-class case, since reconstructing the probabilities on the target domain would require the manual annotation of large volumes of images.

One alternative is to only focus on identifying rare classes~\cite{walker:21}, and treating the rest as a background. Treating most of the images as a background allows to use a hard-negative mining scheme, which consists of retraining the model with images that would belong to the background and were wrongly classified. However, while useful for specific target studies, this does not help when one is interested in the entire taxonomic distribution of the target ecosystem.

Ref.~\cite{yang:22} studies out-of-domain generalization with particular emphasis on the appearing of new classes, and on a radical domain shift such as sampling from a different monitoring station. 
For example, they train on data coming from one station, and test on data coming from another. 
It does however not address the DS appearing within a same station, \textit{i.e.} the difference in performance from the same station, just because the data are sampled at a different point in time. 
In particular, the focus is on differing stations, without specifying whether the OOD cells belong to a specific day of sampling, thus representing the data distribution in that particular point in time.

\section{Data}\label{app:data}
\begin{figure*}
    \centering
    \includegraphics[width=\textwidth]{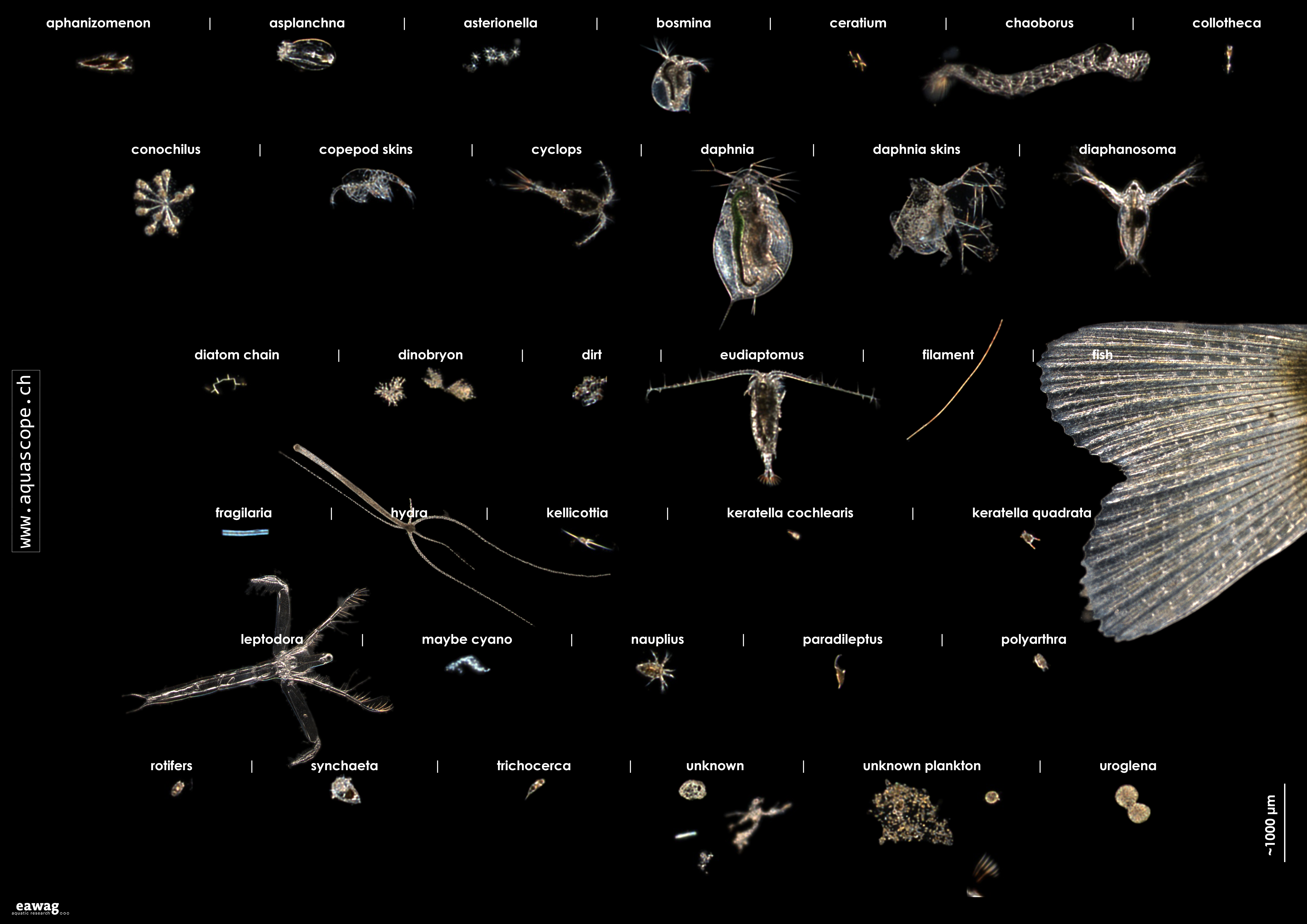}
    \caption{Sample images from each class in the ZooLake2.0 dataset.}
    \label{fig:ZooLake2.0}
\end{figure*}

In Fig.~\ref{fig:ZooLake2.0}, we show a collection of images from each class. 
For a detailed description of all the classes, we refer the reader to the appendix of Ref.~\cite{kyathanahally:21}.

\section{Models}\label{app:models}
Transfer learning~\cite{tan2018survey} is used to save time and computing resources. We import the models and weights pre-trained on the ImageNet dataset~\cite{deng2009imagenet} from the \textit{TIMM} Python package~\cite{rw2019timm}, and replace the final layer with a dense layer which connects to $N_c$ outputs. The whole model training scheme is organized into two stages. In the first stage, all layers of the pre-trained model but the last layer are frozen, only the parameters in the last layer are trained. The training optimizer is AdamW \cite{loshchilov2017decoupled}. The two hyperparameters in AdamW, learning rate ($\eta$) and weight decay are tuned by Bayesian Optimization search \cite{snoek2012practical}, using the \textit{Ray.Tune} package \cite{liaw2018tune}. All images of ZooLake2.0 are resized to a size of $224\times224$, then 25 trials with different hyperparameter combinations are trained for 50 epochs, to find the best combination of hyperparameters by Bayesian Optimization search. The best hyperparameter set is then saved, and used for the second stage of training. In the second stage, by unfreezing all layers, all parameters of the model are trained for 100 epochs, with a low learning rate of $\eta = 10^{-5}$. During the training, the real-time training curve is saved after every epoch. The model is saved as a checkpoint whenever the F1 score on the validation set improves after a new epoch. Once the entire training is completed, the best model and the last epoch model are saved. The training of all models are performed in the \textit{PyTorch} framework \cite{paszke2019pytorch}, with two Nvidia GTX 2080Ti GPUs.

\section{Feature description}\label{app:feature}
In order to describe the morphological and color feature of image, 67 descriptors are extracted for each image. The explanations of these feature descriptors are given below. 
\begin{description}
    \item[width, height] the width and height of bounding rectangle of the object in the image.
    \item[w\_rot, h\_rot] the width and height of rotated bounding rectangle of the minimum area.
    \item[angle\_rot] the angle between rotated and original bounding boxes, ranging from 0 to 90.
    \item[aspect\_ratio] the ratio of width to height.
    \item[rect\_area] the area of bounding box, i.e. the product of width and height.
    \item[contour\_area, contour\_perimeter] the area and perimeter of the contour of the object.
    \item[extent] the ratio of contour area to bounding rectangle area.
    \item[compactness] the squared contour perimeter, divided by $4 \pi \times$ contour area. The circle has a compactness of 1.
    \item[formfactor] the multiplicative inverse of compactness.
    \item[hull\_area, hull\_perimeter] the area and perimeter of the convex hull. Convex hull is the smallest convex polygon that can fit in the object.
    \item[solidity] the ratio of the object area to the convex hull area. This measures the density of an object.
    \item[ESD] equivalent spherical diameter, the diameter of the circle whose area is same as the contour area.
    \item[major\_axis, minor\_axis] the length of major and minor axis of the fitted ellipse.
    \item[angle] the angle between major axis and vertical axis, ranging from 0 to 180.
    \item[eccentricity] the ratio of minor axis to major axis.
    \item[convexity] the ratio of hull perimeter to contour perimeter.
    \item[roundness] the squared hull perimeter, divided by $4 \pi \times$ contour area. The circle has a roundness of 1.
    \item[{\footnotesize intensity\_R\_mean, intensity\_G\_mean, intensity\_B\_mean}] the mean pixel value of 3 color channels respectively.
    \item[intensity\_R\_std, intensity\_G\_std, intensity\_B\_std] the standard deviation of pixel values of 3 color channels respectively.
    \item[hue\_mean, saturation\_mean, brightness\_mean] the mean value of hue, saturation and brightness respectively.
    \item[hue\_std, saturation\_std, brightness\_std] the standard deviation of hue, saturation and brightness respectively.
    \item[blurriness] the mean of absolute values of Laplacian.
    \item[noise] the standard deviation of all pixel values of an image.
    \item[$M_{ij}, \mu_{ij}, \eta_{ij}, I_{ij}$] these are the moments of object contour, up to the third order, as described in Ref.~\cite{hu1962visual}.\footnote{A simple description of these moments can be found at \url{https://en.wikipedia.org/wiki/Image_moment} (visited on Jan. $1^\mathrm{st}$, 2024).} These moments are weighted sums of the pixels, which can be used to characterize images.
    The $M_{ij}$ are raw moments - they do not have any specific property beyond characterizing the image; the $\mu_{ij}$ are the central moments - these are translation invariant and can provide information about the orientation and elongation of the object in the image; the $\eta_{ij}$
    describe the distribution of pixel intensities with respect to the horizontal and vertical axes, which is invariant to both translation and scale, they so can be related to the orientation of the image; the $I_{i}$ are invariant with respect to translation, scale, and rotation, and are therefore useful for characterizing and recognizing shape in image. 
\end{description}

\section{Targeted Augmentations}\label{app:augment}

\begin{figure}[t]
    \centering
    \includegraphics[width=\columnwidth]{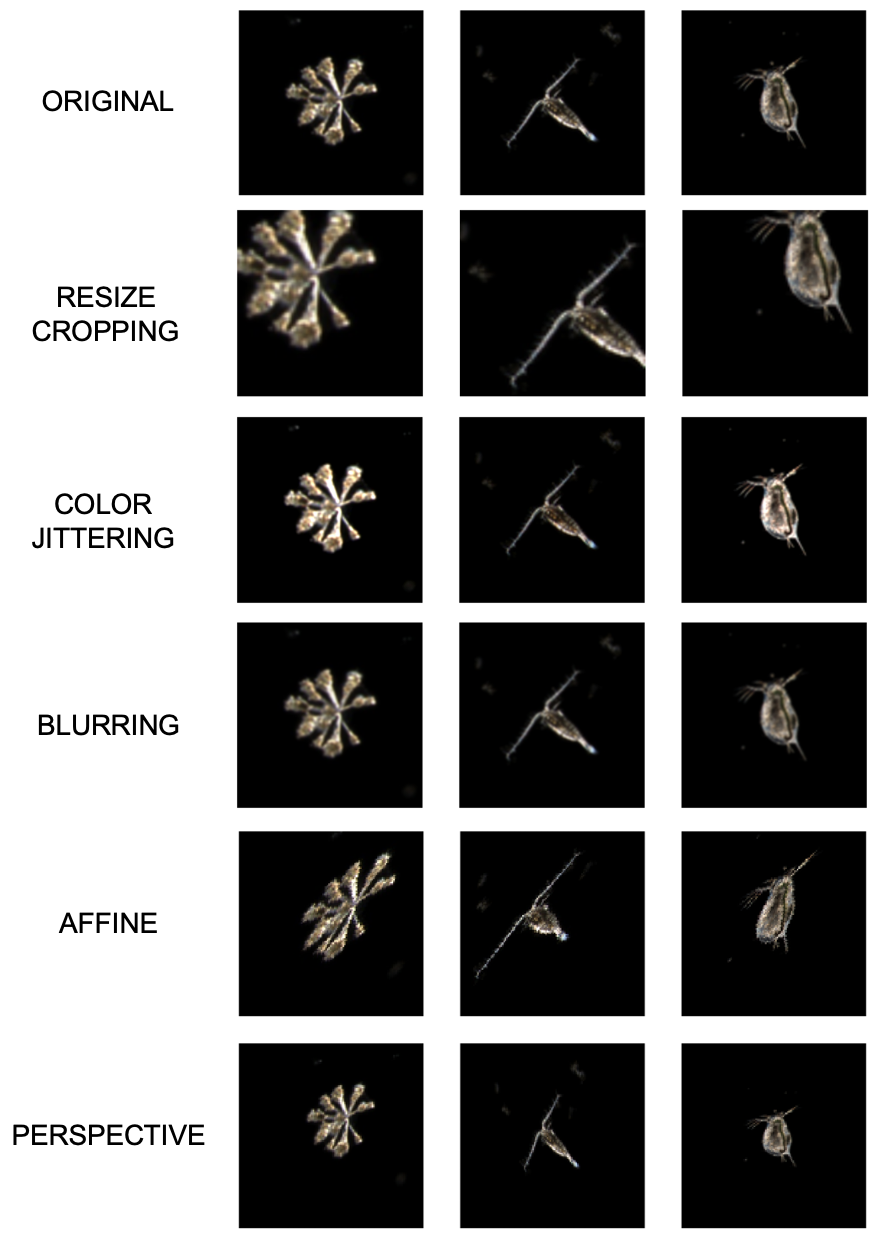}
    \caption{
    Graphical examples of targeted augmentations compared to original images.
    }
    \label{fig:aug_targeted}
\end{figure}

According to the sensitivity analysis in Sec.~\ref{sec:pxy-features}
the most relevant features are saturation\_std, intensity\_G\_std, eccentricity, compactness, formfactor, convexity, hull\_area, height, ESD, blurriness, angle\_rot and some image moments, including raw moments, translation invariant moments, scale invariant moments and rotation invariant moments. 
These are related with color, shape, object size, blurriness and orientation. 
These results do not highlight a single distinguishable feature, possibly due to the small size and number of OOD cells, which give us uncertain estimates of the sensitivities.
However, it still indicates that robustness towards changes in shape, color, size, orientation and blurriness should be addressed. 
Therefore, we deploy the following augmentations: 
\begin{itemize}
    \item \textbf{Color jittering} includes the tuning of brightness, contrast, saturation and hue. This is the targeted augmentation for saturation\_std, intensity\_G\_std. 
    \item \textbf{Gaussian blur} blurs image with randomly chosen Gaussian blur. This is the targeted augmentation for blurriness.
    \item \textbf{Random affine} includes the translation and shearing of original images. This is the targeted augmentation for shape-dependent features, \eg compactness, form factor, eccentricity, convexity and moments. 
    \item \textbf{Random resized crop} crops a random portion of image and resizes it to a given size. This is useful to handle the fluctuation in size-dependent features, \eg height, hull\_area, ESD, raw moments and translation invariant moments. 
    \item \textbf{Random perspective} performs a random perspective transformation of the given image with a given probability. This effectively adjusts the shape of the object and helps to enhance the model robustness against the variation of shape-dependent features, including compactness, formfactor, eccentricity, convexity and all moments. 
\end{itemize}
Examples of the augmented image are shown in Fig.~\ref{fig:aug_targeted}.

\section{RGB channel standardization}\label{app:rgb}

\begin{figure}
    \centering
    \includegraphics[width=\columnwidth]{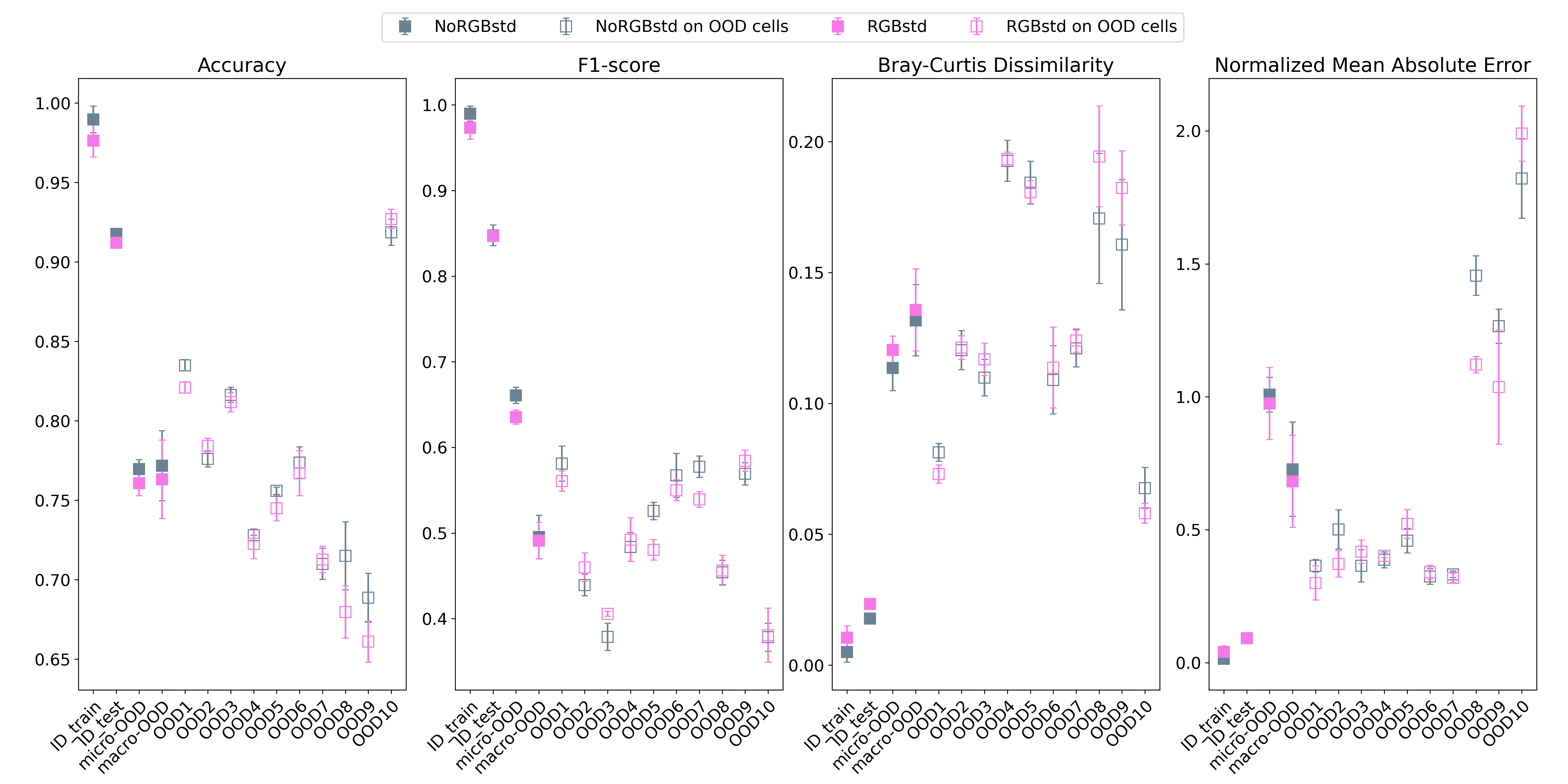}
    \caption{ID and OOD Performances of a MobileNet model, with and without the RGB channel standardization. }
    \label{fig:rgbstandard}
\end{figure}

As we report in Fig.~\ref{fig:rgbstandard}, standardizing the RGB channels does not help to improve the OOD performance of our models.

\section{Other ways of estimating the distribution shift}\label{app:dist}
While the Hellinger distance is the one that best correlates with the performance drop in our data, we also try several other dissimilarity measures as estimator of dataset compositional shift. We report on the following:
\begin{itemize}
    \item Hellinger distance (described in the main text)
    \item Wasserstein distance
    \item Kullback-Leibler divergence
\end{itemize}

\paragraph*{Wasserstein Distance}
The Wasserstein distance, also known as Earth Mover's Distance (EMD), is a mathematical measure of the distance between two probability distributions \cite{kantorovich1960mathematical}. It measures the minimum amount of work required to transform one distribution into the other, where work is defined as the amount of "mass" moved multiplied by the distance it is moved. Given two normalized distributions $p$ and $q$ defined on a discrete one-dimensional support, the Wasserstein distance is 
\begin{equation}\label{eq:wd}
    D_\mathcal{W}^{f} = \frac{1}{n_\mathrm{bins}} \sum_{i=1}^{n_\mathrm{bins}} \left|p_i-q_i\right|.
\end{equation}

\paragraph*{Kullback-Leibler Divergence}
Kullback-Leibler (KL) divergence, also known as relative entropy, is a measure of how one probability distribution differs from a second, reference distribution~\cite{kullback1951information}. Different from the Hellinger and Wasserstein distance, the Kullback-Leibler divergence is an asymmetric measure.\footnote{So, strictly speaking, it is not a distance.} The divergence from $q$ to $p$ $D_\mathcal{KL}^{f}(p \parallel q)$ and the divergence from $p$ to $q$ $D_\mathcal{KL}^{f}(q \parallel p)$ are different. For discrete probability distributions $p$ and $q$ defined on the same sample space $\mathcal{X}$, the KL divergence from $q$ to $p$ is 
\begin{equation}\label{eq:kl}
    D_\mathcal{KL}^{f}(p \parallel q) = \sum_{x\in\mathcal{X}}p(x)\log{\frac{p(x)}{q(x)}}\,.
\end{equation}

In our study, we keep the in-domain feature distribution always as reference distribution, calculate the divergence of out-of-domain distribution to the reference. 

\begin{figure}[t]
    \centering
    \includegraphics[width=\columnwidth]{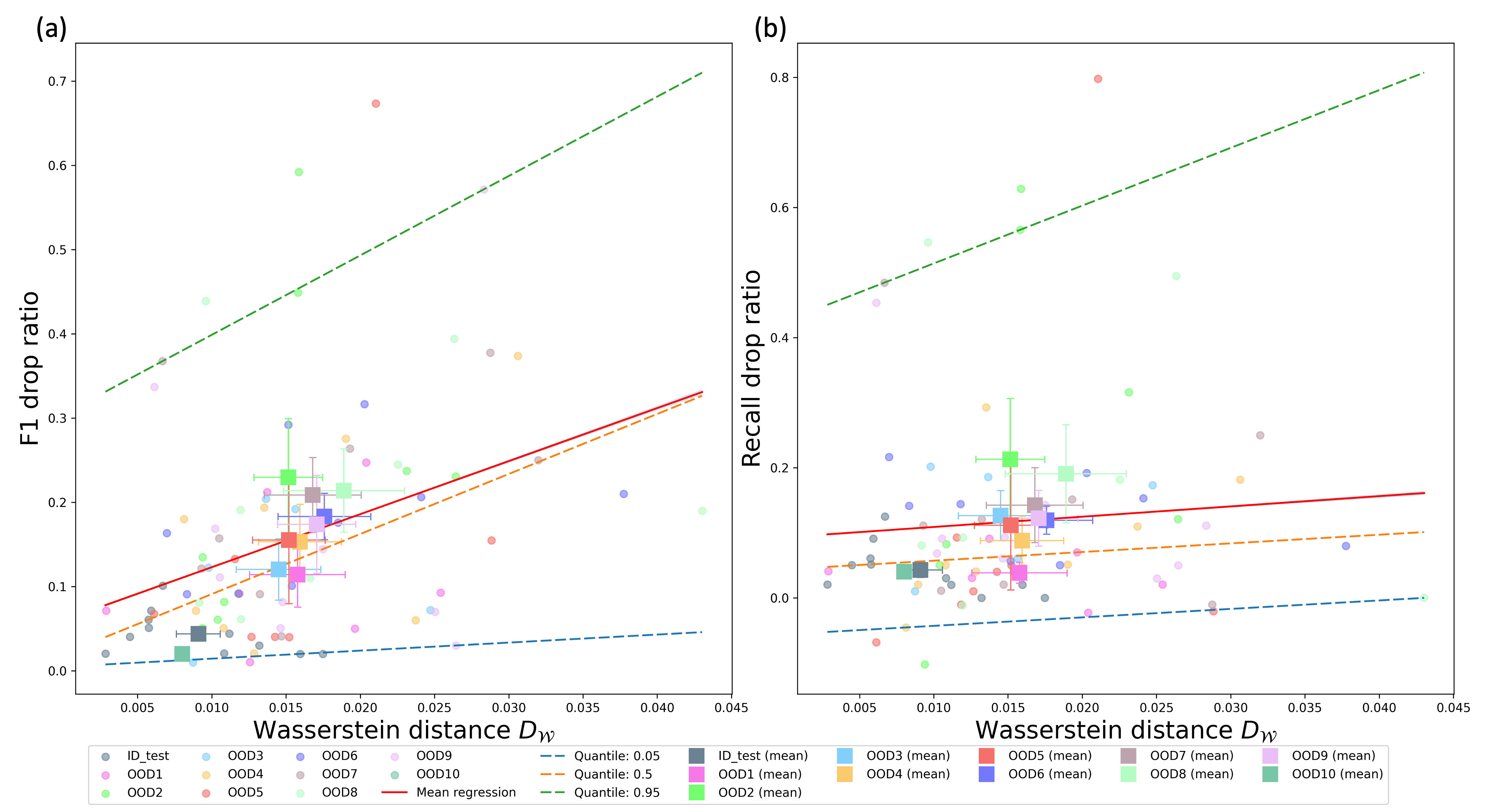}

    \caption{\textbf{(a)}: F1 drop [Eq.~\eqref{eq:drop}] as a function of the Wasserstein distance from the training set [Eq.~\eqref{eq:df}]. The error bars are calculated by taking the fluctuations among the different classes in each test cell. \textbf{(b)}: Same, for the recall. 
    }
    \label{fig:cor_mean_wasserstein}
\end{figure}

\begin{figure}[t]
    \centering
    \includegraphics[width=\columnwidth]{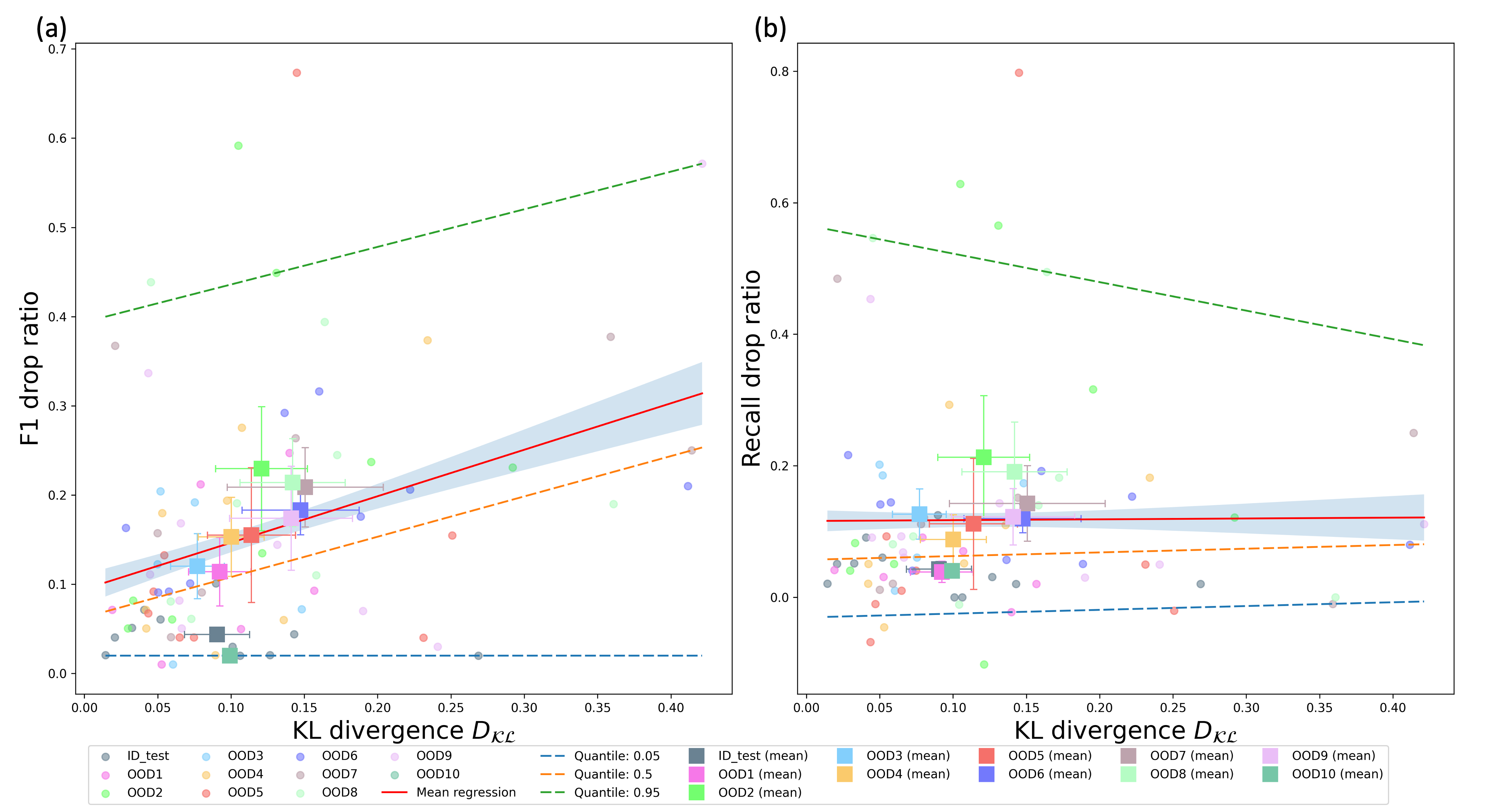}

    \caption{\textbf{(a)}: F1 drop [Eq.~\eqref{eq:drop}] as a function of the Kullback-Leibler divergence from the training set [Eq.~\eqref{eq:df}]. The error bars are calculated by taking the fluctuations among the different classes in each test cell. \textbf{(b)}: Same, for the recall. 
    }
    \label{fig:cor_mean_KL}
\end{figure}

In Fig.~\ref{fig:cor_mean_wasserstein} and Fig.~\ref{fig:cor_mean_KL}, we show the equivalent of Fig.~\ref{fig:cor_mean}, but using $D_\mathcal{W}$ and  $D_\mathcal{KL}$. The correlation between distance and performance drop is lower than using the Hellinger distance.

\section{Per-class performance drop}\label{app:drop-per-class}
\begin{figure}[t]
    \centering
    \includegraphics[width=0.5\textwidth]{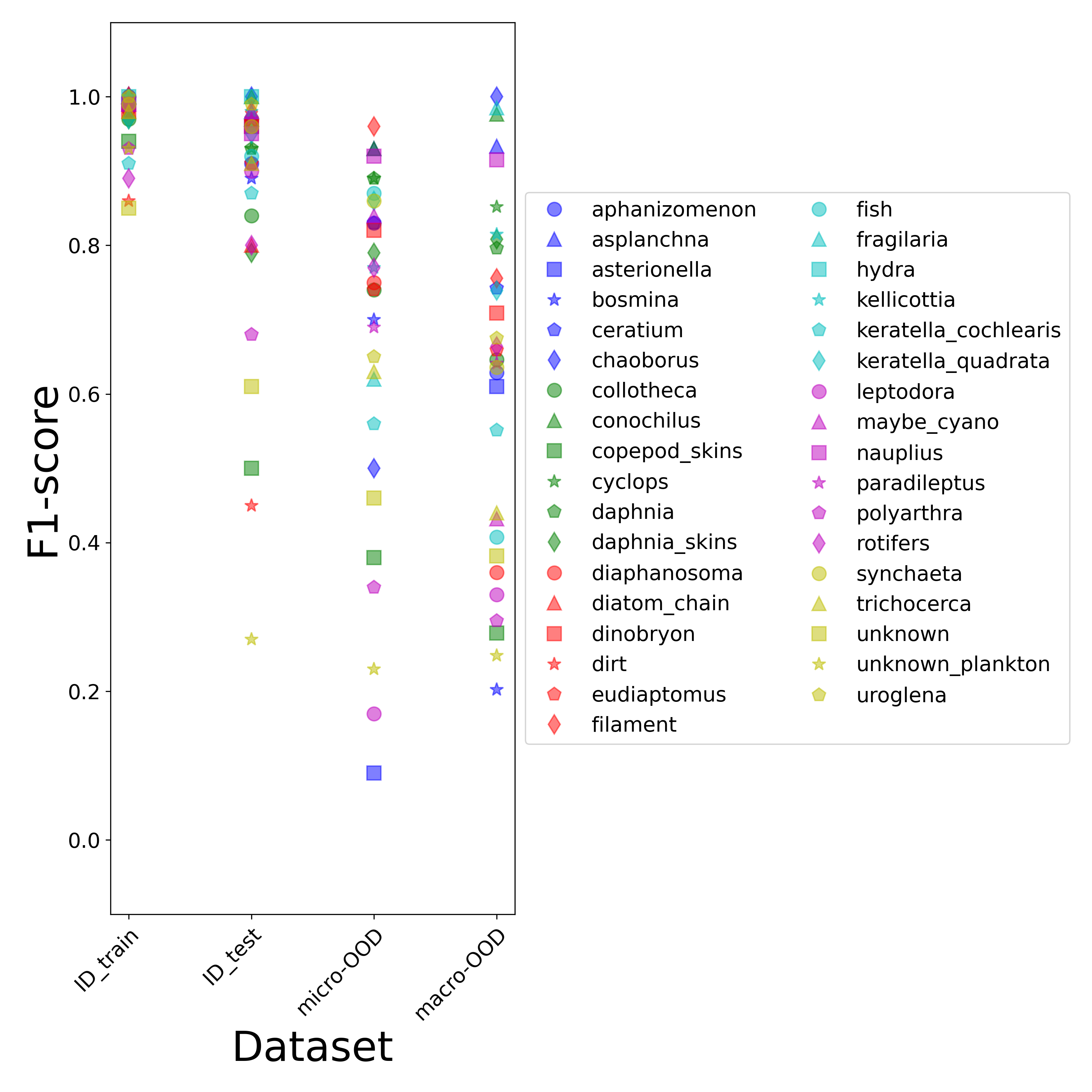}
    \caption{Performance drop for each class, with baseline model. 
    }
    \label{fig:degrad_baseline}
\end{figure}

\begin{figure}[t]
    \centering
    \includegraphics[width=0.5\textwidth]{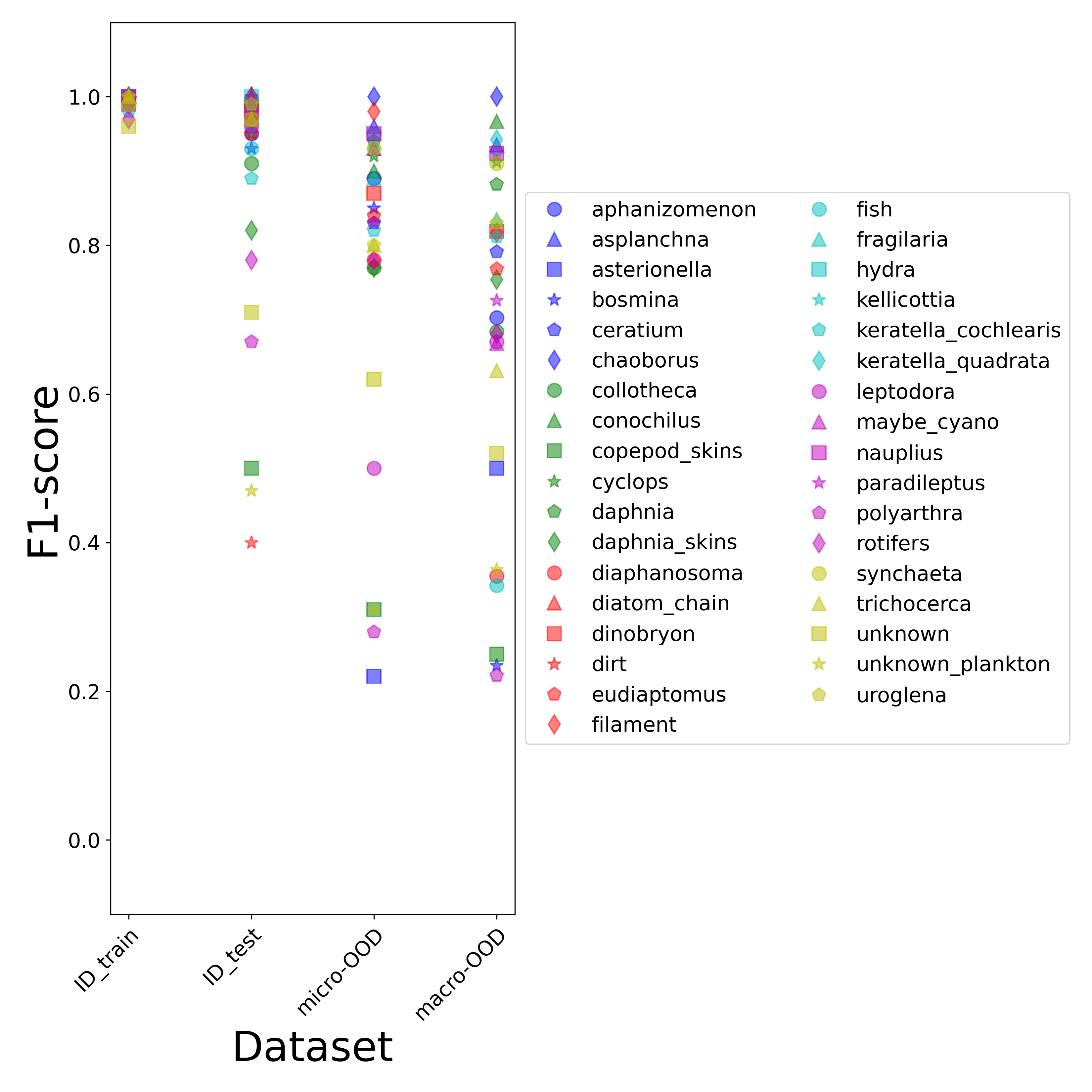}
    \caption{Performance drop for each class, with \bestmodel model.}
    \label{fig:degrad_best}
\end{figure}

Fig.~\ref{fig:degrad_baseline} shows the F1-score of MobileNet model on ID\_train, ID\_test and OOD (micro and macro) sets, for all 35 classes. Some classes such as \texttt{asterionella}, \texttt{bosmina}, \texttt{leptodora} and \texttt{polyarthra} have worse OOD performance compared to other classes. Fig.~\ref{fig:degrad_best} displays the similar results, but using the \bestmodel model. In general, the OOD performance of each single class improves. Note, also, that most of the low performances are not associated with living organisms.

\section{Performance drop with aggregated classes}\label{app:aggregated}
\begin{figure}[t]
    \centering
    \includegraphics[width=\columnwidth]{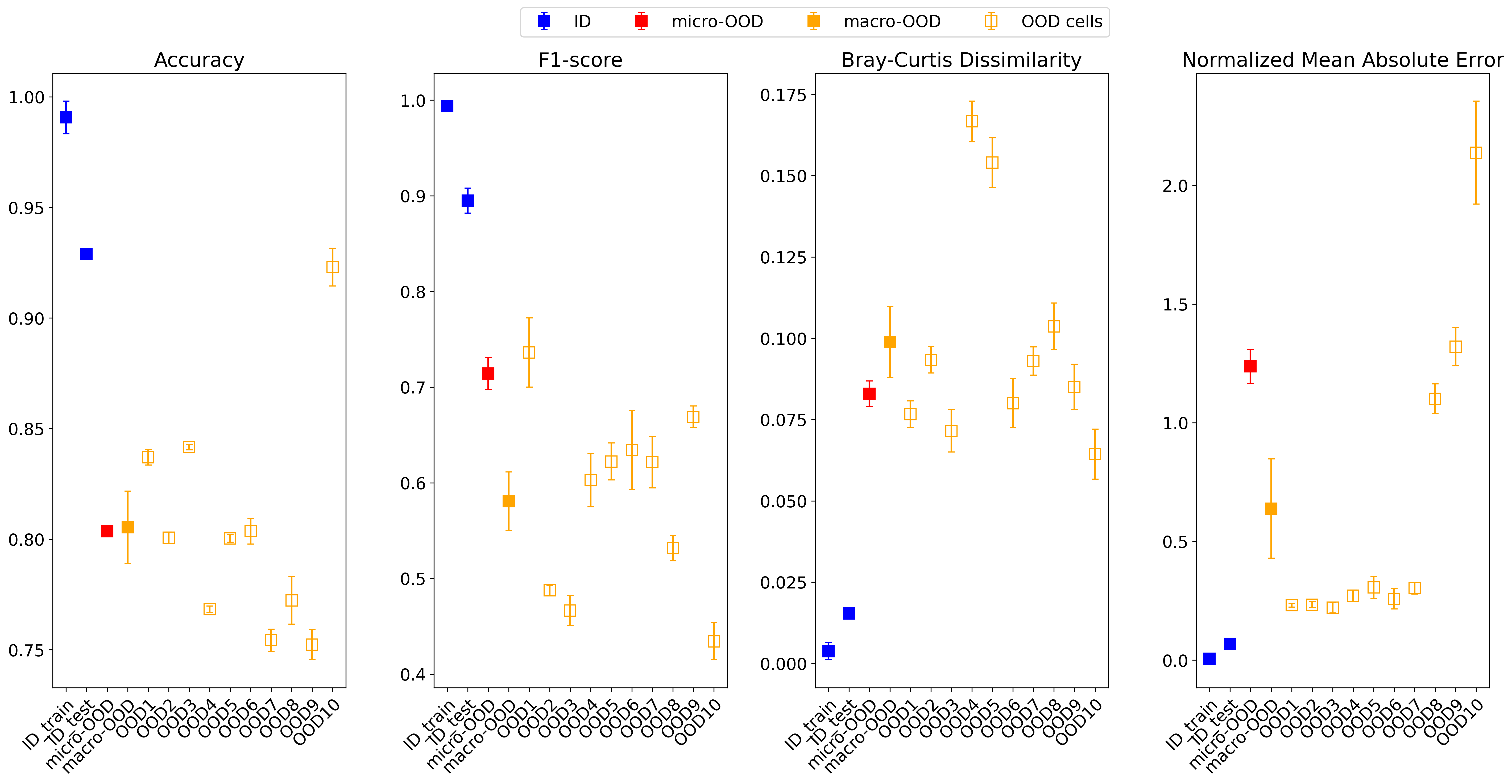}
    \caption{
    Same figure as Fig.~\ref{fig:degrad}, but using superclasses.
    }
    \label{fig:degrad_supra}
\end{figure}

The classes in the ZooLake2.0 dataset can be coarsened in superclasses. As a result, the classes \texttt{hydra} and \texttt{kellicottia} are merged into a new superclass \texttt{hydra}, classes \texttt{keratella\_cochlearis} and \texttt{keratella\_quadrata} are combined as \texttt{keratella}, classes \texttt{nauplius} and \texttt{paradileptus} become a new \texttt{nauplius}, classes \texttt{collotheca}, \texttt{conochilus}, \texttt{polyarthra}, \texttt{rotifers}, \texttt{synchaeta} and \texttt{trichocerca} are merged into a single superclass \texttt{rotifer}, classes \texttt{dirt}, \texttt{unknown} and \texttt{unknown\_plankton} go to the superclass \texttt{unknown}.
In Fig.~\ref{fig:degrad_supra}, we show the OOD performance drop, when using superclasses. 

\begin{figure}[t]
    \centering
    \includegraphics[width=\columnwidth]{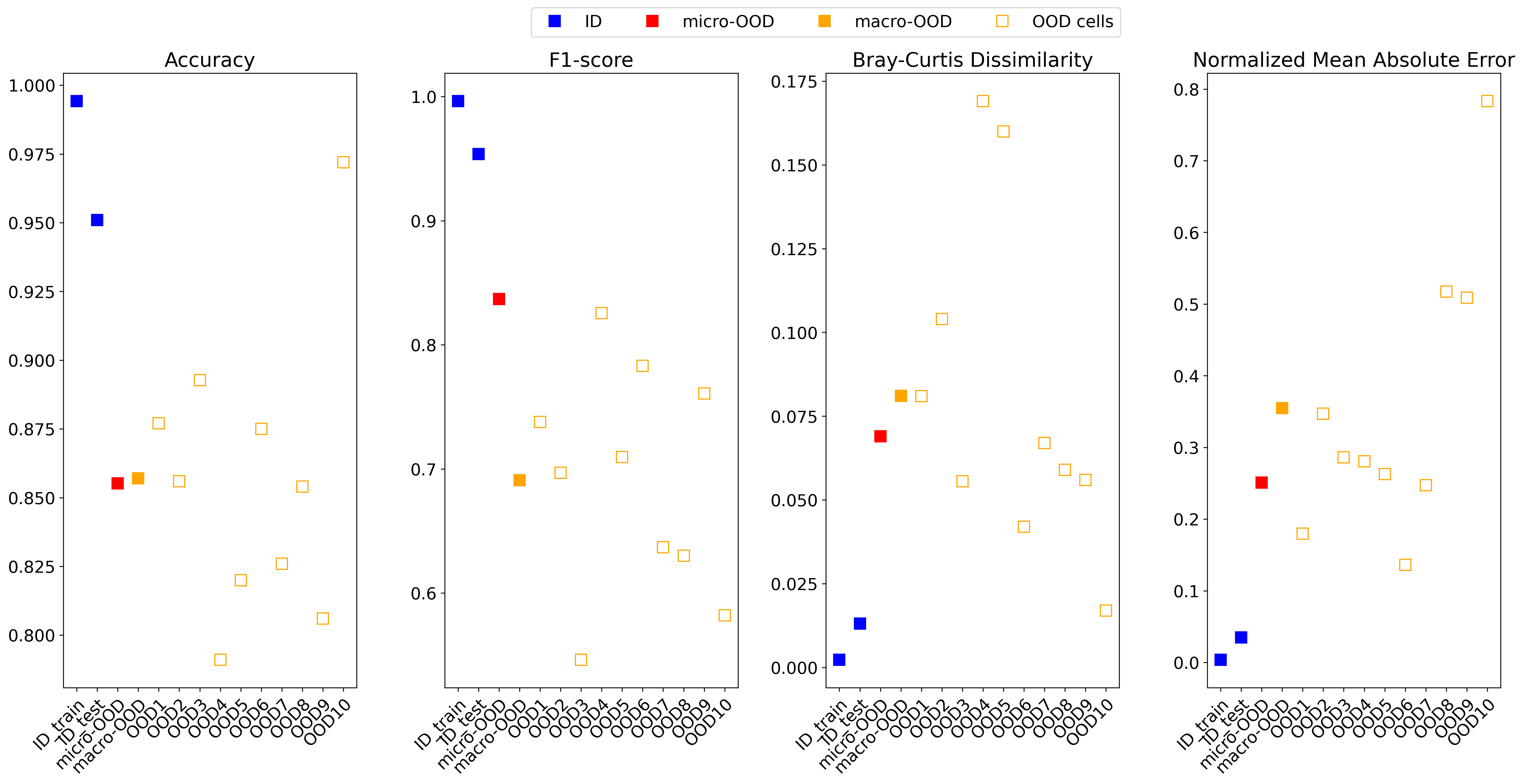}
    \caption{
    Same figure as Fig.~\ref{fig:degrad_supra}, but using \bestmodel model.
    }
    \label{fig:degrad_supra_best}
\end{figure}

In Fig.~\ref{fig:degrad_supra_best}, we show the classification performance on the \bestmodel model using superclasses. Although all performance metrics show improvement over the baseline model, the OOD performance still lags far behind that of the ID.

\begin{figure}
    \centering
    \includegraphics[width=\columnwidth]{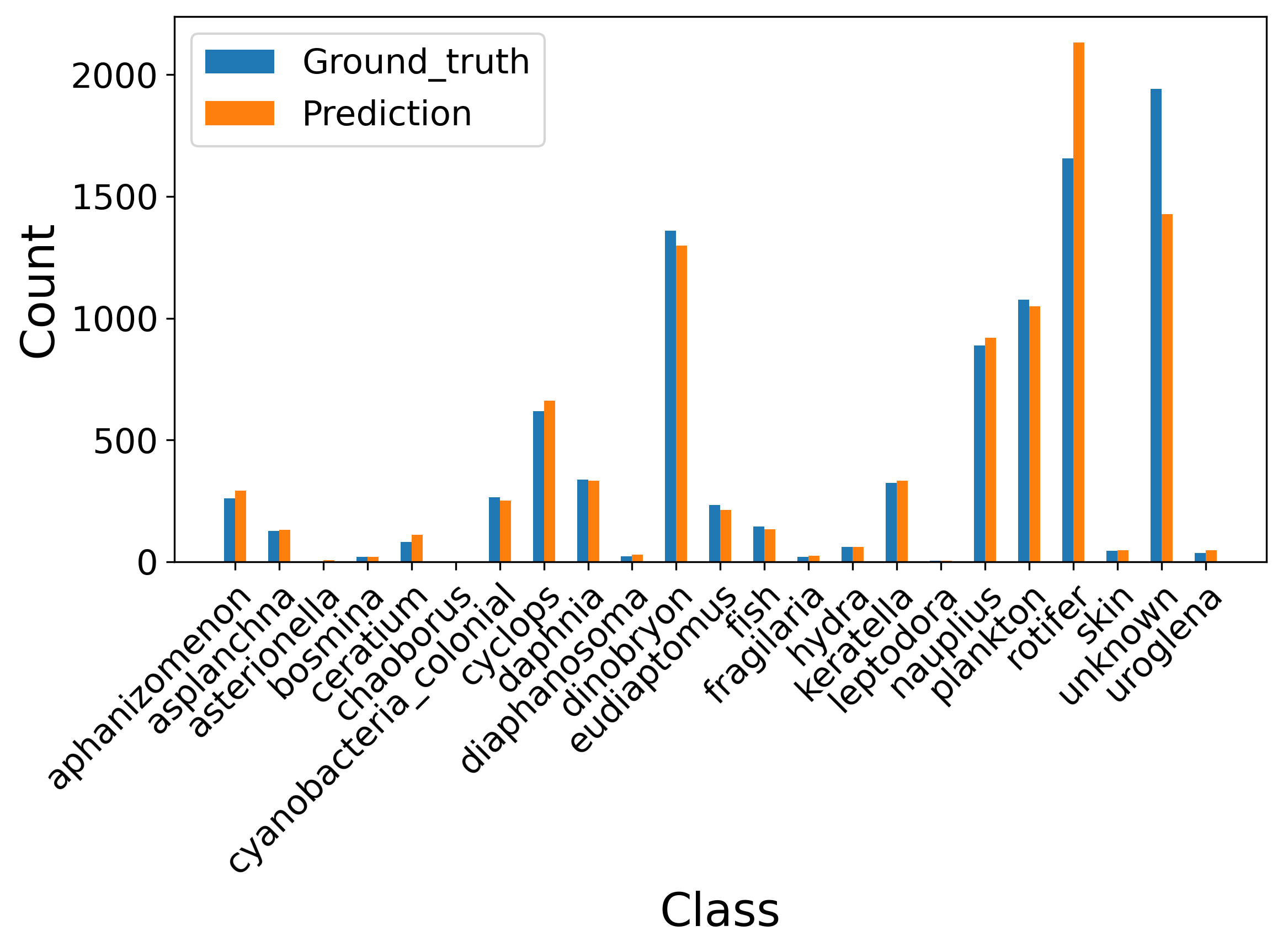}
    \caption{     
    Bar plot showing the true and the predicted plankton abundances, for the \bestmodel model, on the aggregated OOD data on superclass level.
    }
    \label{fig:bar_best_supra}
\end{figure}

In Fig.~\ref{fig:bar_best_supra}, we show the equivalent of Fig.~\ref{fig:final}c, updating the categories with superclasses.

\section{OOD degradation on the single OOD cells}
\label{app:scatter-ood}

\begin{figure*}[t]
    \centering
    \includegraphics[width=\textwidth]{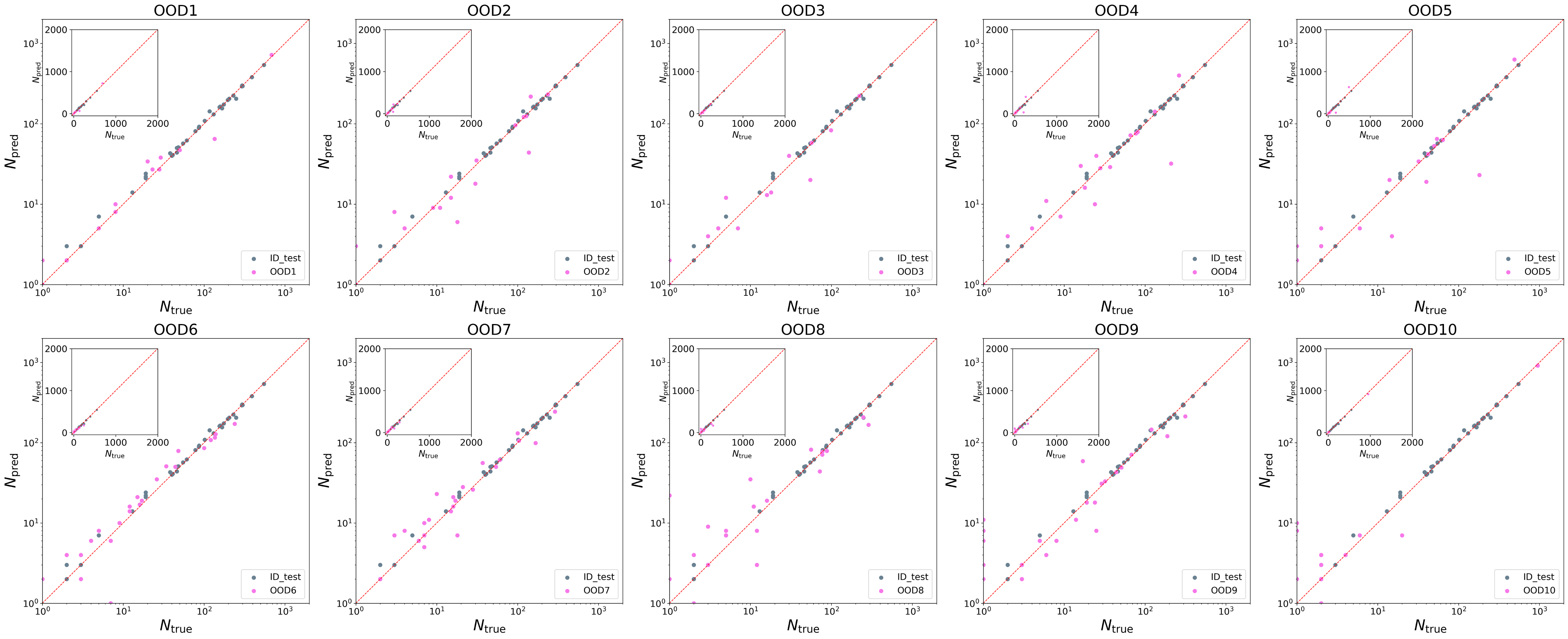}
    \caption{Same figures as Fig.~\ref{fig:population_scatter}, but for the individual OOD cells.}
    \label{fig:population_scatter_cells}
\end{figure*}

\begin{figure*}[t]
    \centering
    \includegraphics[width=\textwidth]{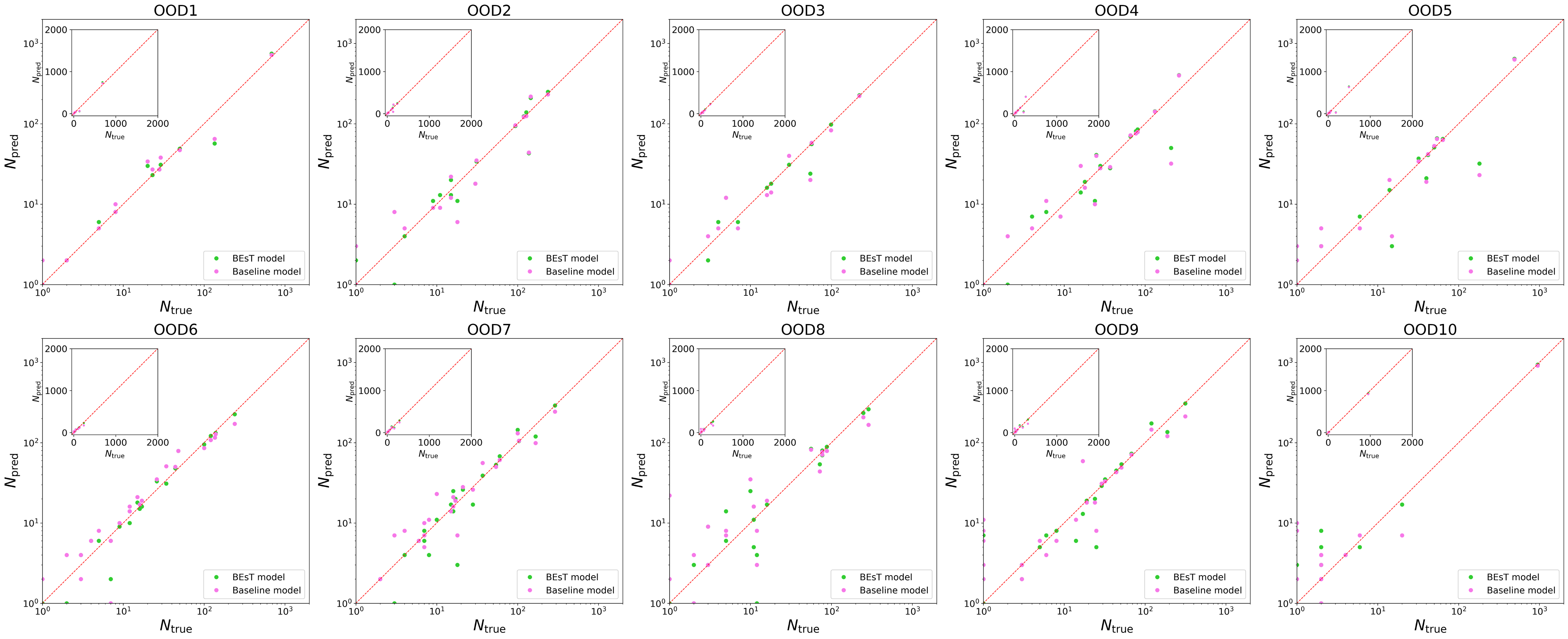}
    \caption{Same figures as Fig.~\ref{fig:final}a, but for the individual OOD cells.}
    \label{fig:population_scatter_cells_bestmodel}
\end{figure*}

In Fig.~\ref{fig:population_scatter_cells}, we show the population scatter plot for 10 individual OOD cells, comparing the OOD performance of the baseline MobileNet model with the ID test performance. The diagonal lines represent the 1:1 relationship between predicted and true abundances in each OOD cell. 
In Fig.~\ref{fig:population_scatter_cells_bestmodel} we show the same figure, but comparing the MobileNet with the \bestmodel model.


\section{Sensitivities for all models and classes}\label{app:allresults}

\subsection{Estimating performance drop on new data}\label{app:per-unlabeled}

\begin{figure}[t]
    \centering
    \includegraphics[width=\columnwidth]{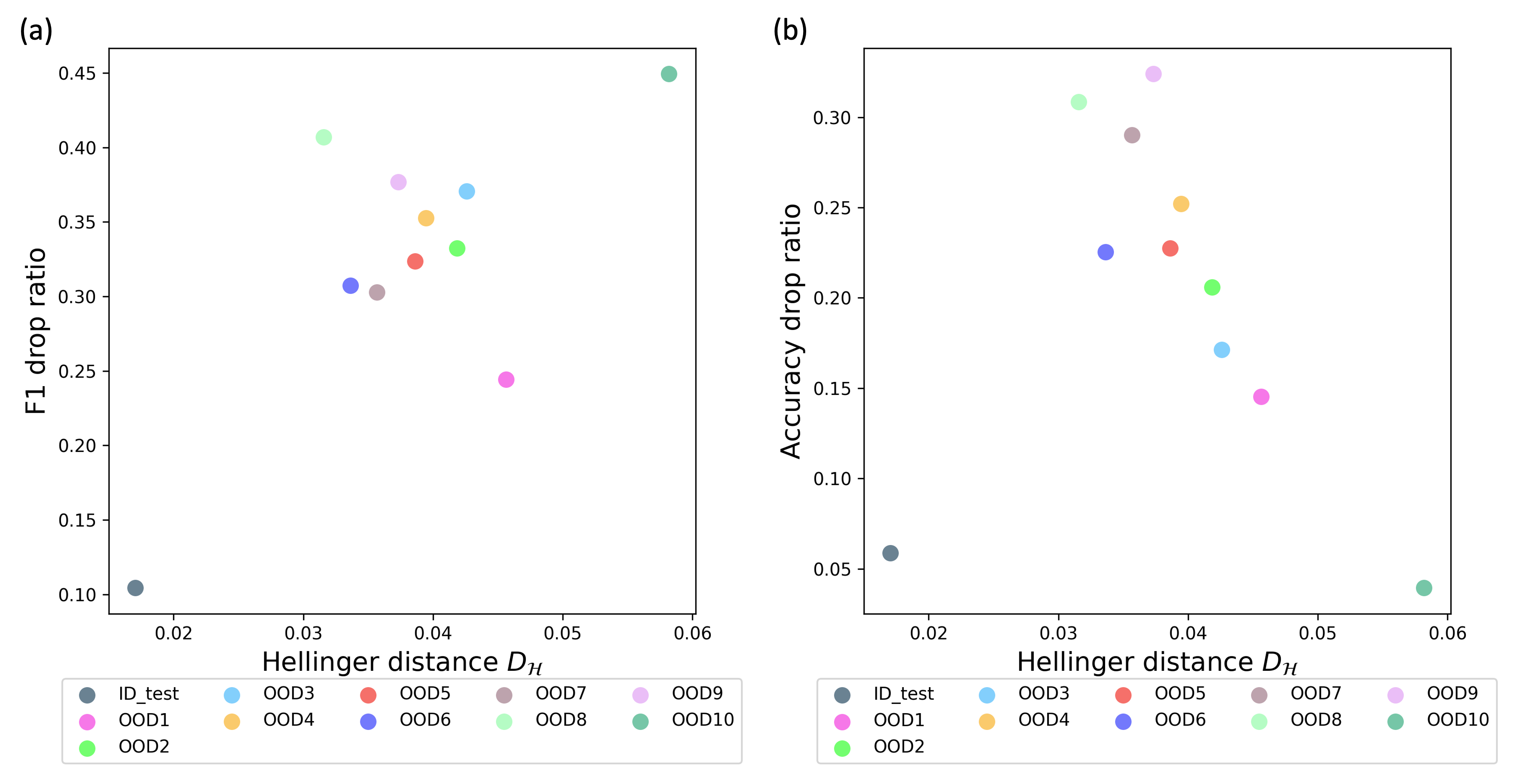}
    \caption{\textbf{(a)} F1 and \textbf{(b)} accuracy drop as a function of the Hellinger distance between the ID and OOD $P(\x)$, for a MobileNet model. An equivalent figure for the \bestmodel architecture is provided in Sec.~\ref{app:per-unlabeled}.
    }
    \label{fig:hellinger-unlabeled}
\end{figure}

\begin{figure}[t]
    \centering
    \includegraphics[width=\columnwidth]{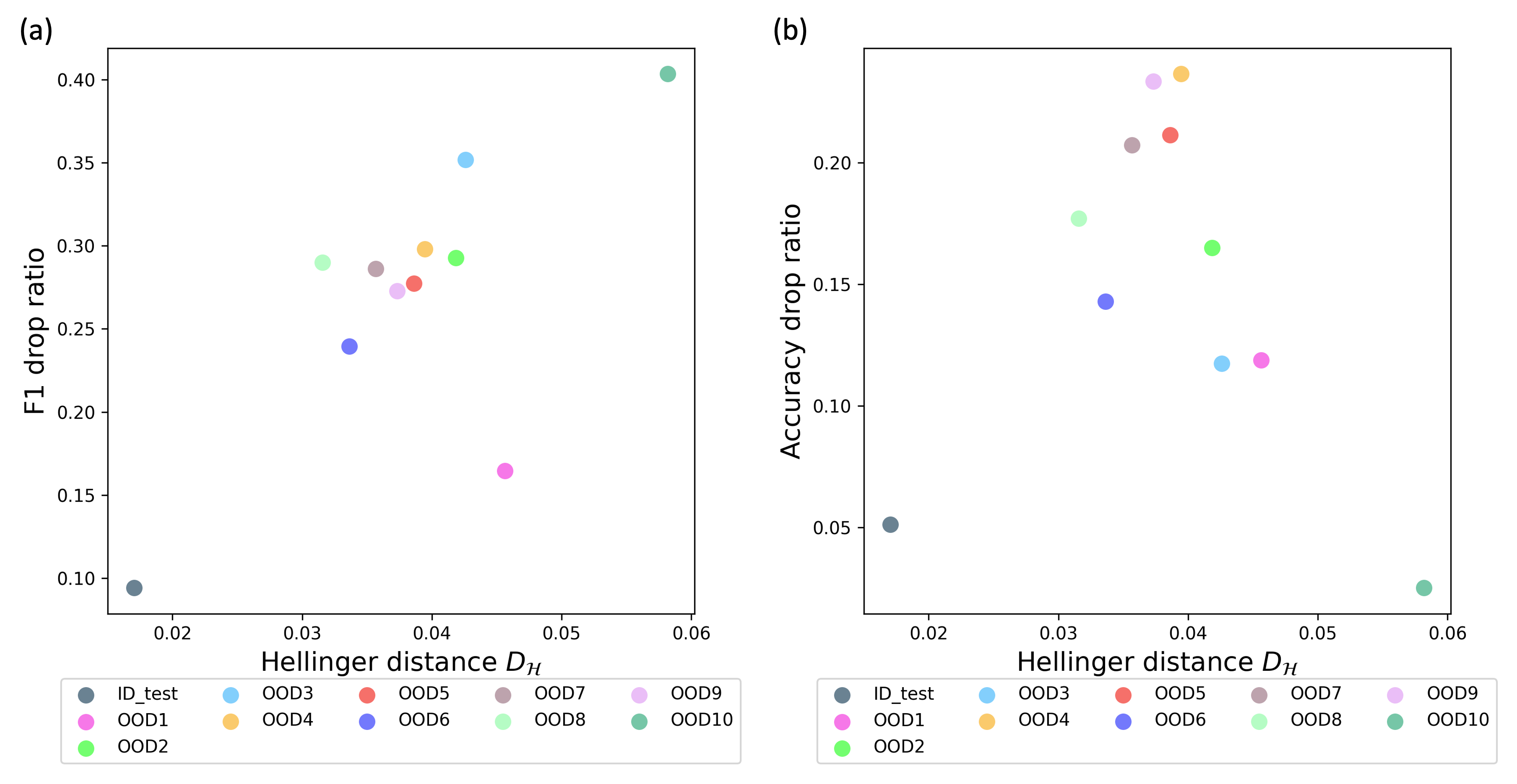}
    \caption{\textbf{(a)} F1 drop and \textbf{(b)} accuracy drop as a function of the Hellinger distance between the ID and OOD $P(\x)$, for the \bestmodel model. Same figure as Fig.~\ref{fig:hellinger-unlabeled}, which is for the baseline model.
    }
    \label{fig:hellinger-unlabeled_best}
\end{figure}

We show the F1 drop ratio as a function of $\HD$ in Fig.~\ref{fig:hellinger-unlabeled}a, for a MobileNet model (we also show this in Fig.~\ref{fig:hellinger-unlabeled_best}, for the \bestmodel model). This could allow us to estimate the F1 drop in a generic OOD dataset. For example, if a new unlabeled dataset has $\HD\simeq0.12$, we can expect an F1 drop of around 10\%. 
This estimate must however take into account the reduced size of the OOD cells: as highlighted in Sec.~\ref{sec:metrics}, the F1 score tends to be artificially lower in the presence of small abundances, so we expect the drop to be lower in the presence of bigger OOD datasets.\footnote{This is, for example, confirmed by the fact that micro OOD F1 scores (calculated on the aggregated OOD data) are larger than macro OOD F1 scores (calculated on the OOD cells one by one, and only later averaged). We see this \textit{e.g.} in Fig.~\ref{fig:single_basic-byArchitecture}.}
Moreover, in this case, since neither the F1 score nor the accuracy drop correlate well with the Hellinger distance, so more work/data is needed to confidently estimate the performance degradation with completely new unlabeled data. 

\subsection{Sensitivities for all models and classes}\label{app:sensitivities}
\begin{figure}[t]
    \centering
    \includegraphics[width=\columnwidth]{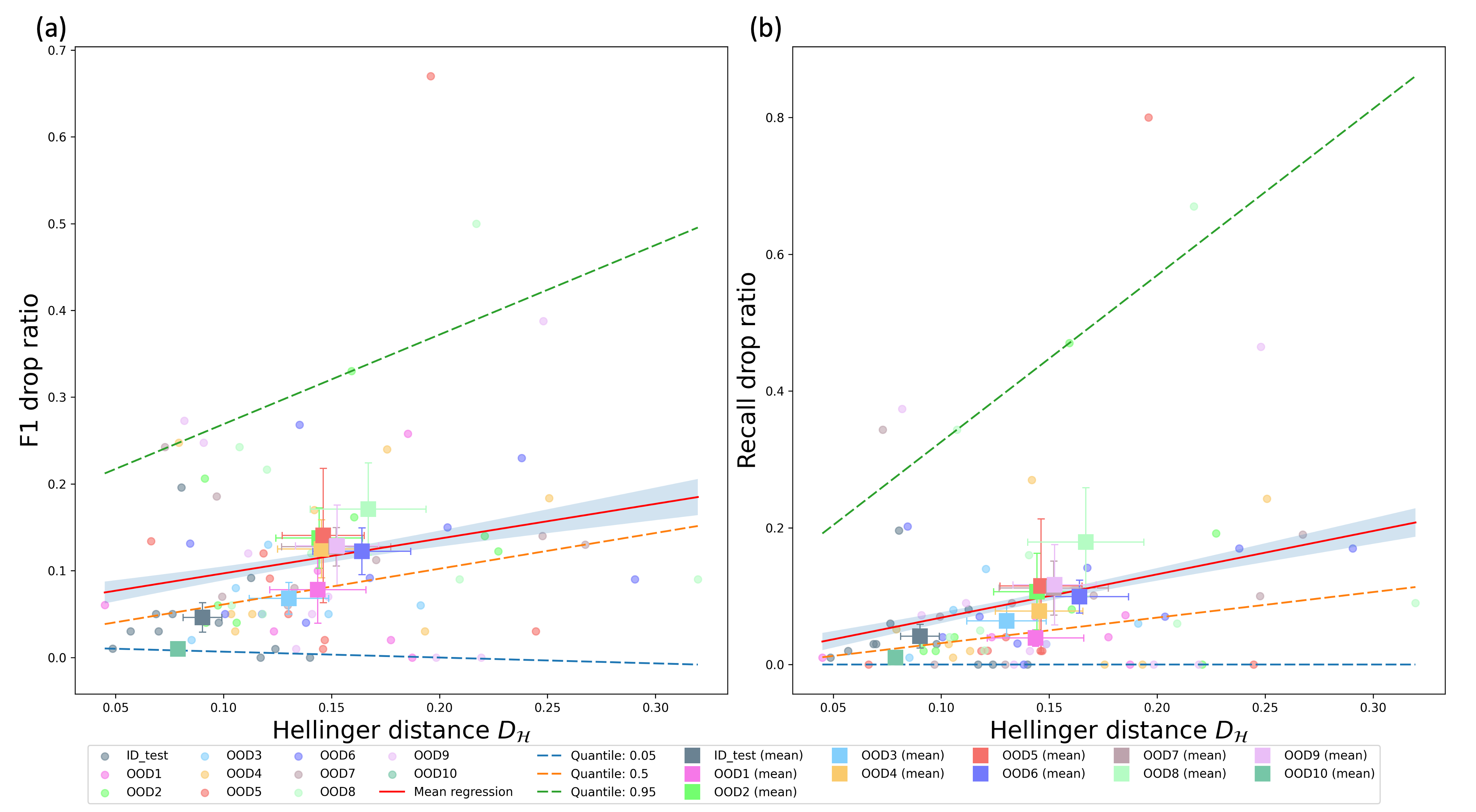}
    \caption{\textbf{(a)}: F1 drop [Eq.~\eqref{eq:drop}] as a function of the Hellinger distance from the training set [Eq.~\eqref{eq:df}]. The error bars are calculated by taking the fluctuations among the different classes in each test cell. \textbf{(b)}: Same, for the recall. These figures are similar to Fig.~\ref{fig:cor_mean}, but using the \bestmodel model.
    }
    \label{fig:cor_mean_best}
\end{figure}

In Fig.~\ref{fig:cor_mean_best}, we show the equivalent of Fig.~\ref{fig:cor_mean}, but for the \bestmodel model. The correlation between performance drop and $\HD$ persists also with a different model.

\begin{figure}[t]
    \centering
    \includegraphics[width=\columnwidth]{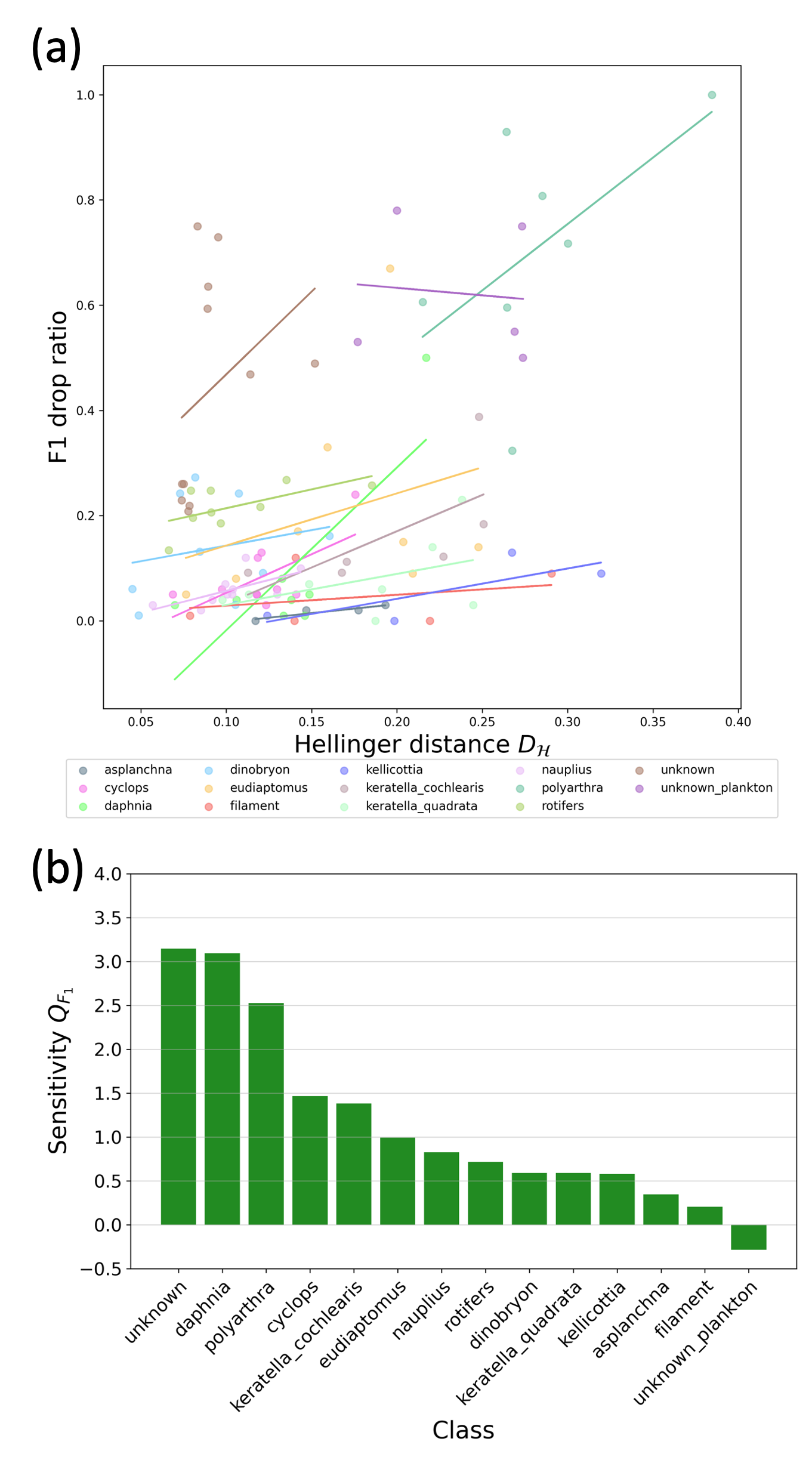}
    \caption{Similar figures to Fig.~\ref{fig:dist-classes}, but using the \bestmodel model. \textbf{(a):} per-class F1 drop as a function of the Hellinger distance. 
    \textbf{(b):} sensitivity $Q_{F_1}$ related to each class of those for which enough data is available. 
    }
    \label{fig:dist-classes_best}
\end{figure}

In Fig.~\ref{fig:dist-classes_best}, we show the per-class sensitivities  for the \bestmodel model. This figure is the equivalent of Fig.~\ref{fig:dist-classes}. Note that, as a result of our pipeline addressing OOD robustness, the values of the sensitivities are lower, and that \texttt{dinobryon} is not the most sensitive plankton class anymore.

\section{Comparison of different architectures on the OOD cells}\label{app:arch}
\begin{figure*}[t]
    \centering
    \includegraphics[width=\textwidth]{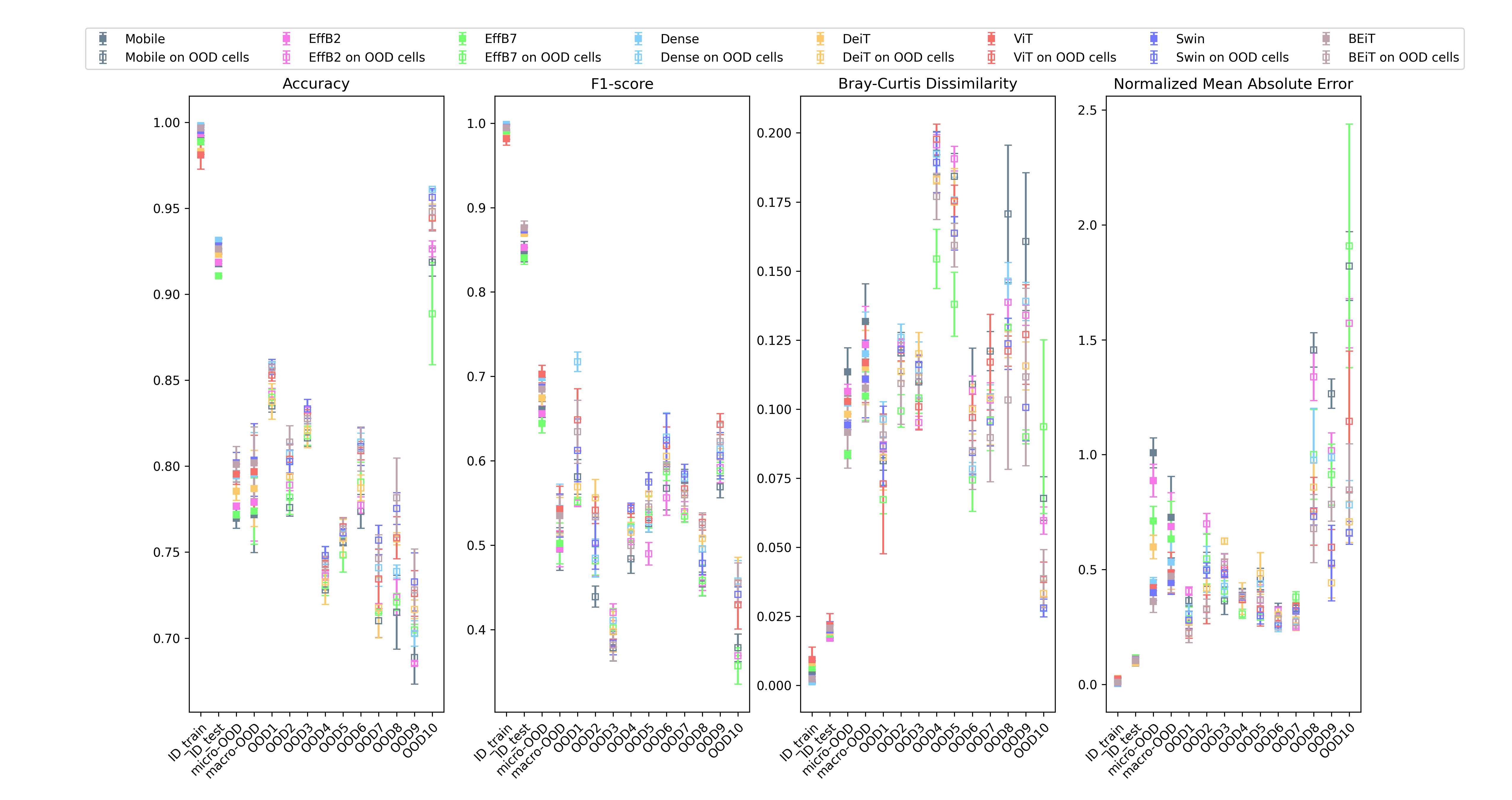}
    \caption{Comparison of different architectures in each of the ID and OOD sets.}
    \label{fig:single_basic-byOOD}
\end{figure*}

In Fig.~\ref{fig:single_basic-byOOD} we compare the performance of all the simulated architectures in each of the ID and OOD sets. There is no single architecture that performs best in each of the OOD cells.

\section{Different ways of ensembling}\label{app:ensemble}
Ensembling can be carried out both by comparing different architectures, and by comparing different instances of the same architecture~\cite{kyathanahally:21}. 
We train 3 models for each architecture, with basic augmentations, and compare both methodologies. 

For the single-architecture ensembling, we join the three trained instances. For the multi-architecture ensembling, we do not mix CNNs with Vision Transformers, because they have very different confidence vectors~\cite{kyathanahally:22}. Hence, we obtain a single CNN ensemble and a single Transformer ensemble (we call them E-CNN and E-Trans). In both cases, we take a single model for each architecture, the one with the best validation F1 score.
Note that multi- are more expensive than single-architecture ensembles, since they use 4 models, and require the training of 12.

\begin{figure*}[t]
    \centering
    \includegraphics[width=\textwidth]{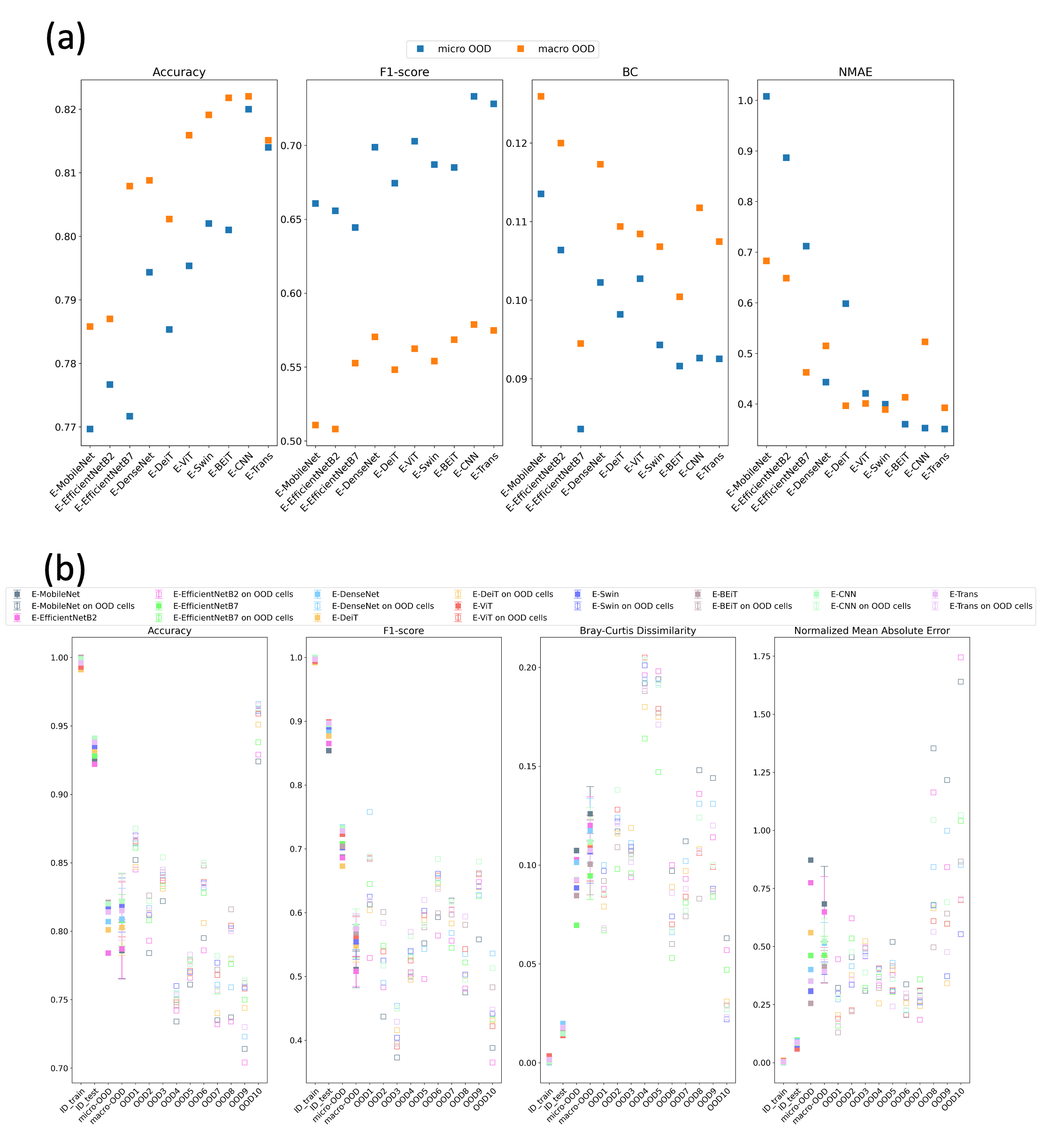}
    \caption{Performances of the ensemble models, trained with basic data augmentation. \textbf{(a)}: Micro- and Macro-OOD performances, for each ensemble. \textbf{(b)}: Performances on each OOD cell. Error bars on the macro-OOD are the standard errors among the single OOD cells.}
    \label{fig:ensemble-compare}
\end{figure*}

In Fig.~\ref{fig:ensemble-compare}a, we compare the OOD performance of all the trained ensembles. We have several remarks:
\begin{itemize}
    \item Multi-architecture CNN ensembles outperform those with transformers. 
    \item While, depending on the chosen metrics, the best-performing ensemble changes, we see that the multi-architecture ensembles are not outperforming the single-architecture ones. However, while E-Trans are not better than their single-model counterparts, E-CNN offers a big improvement with respect to single-architecture CNN ensembles. This is consistent with the ID results reported in Ref.~\cite{kyathanahally:21}.
    \item The best single-architecture CNN is EfficientNet-B7. This is consistent with the (in-dataset) results of Ref.~\cite{kyathanahally:21}.
    \item Single-model transformer ensembles, in particular BEiTs, have very good performances. The good performance of BEiTs was already reported in Ref.~\cite{maracani:23}, while that of single-model Transformer ensembles was noted in Ref.~\cite{kyathanahally:22}.
    \item The best ensembles are EfficientNet-B7, and BEiT. EfficientNets have the best macro-averaged metrics, while BEiTs have the best micro-averaged metrics. This essentially indicates that the former perform better with minority classes, and the latter with majority classes. 
\end{itemize}

In Fig.~\ref{fig:ensemble-compare}b, we show the performances of all models on each single OOD cell.

\section{More test time augmentations}\label{app:tta}
In Tab.~\ref{tab:TTA_flip}, we show a comparison between flipping and rotating. 
Flipping does not increase the performance with respect to rotating. The time complexity of the TTA operation in the last row is doubled, but the model performance is not improved.
This is expected, since the only difference stays in the chirality of the images, but plankton images are generally not chiral. 
\begin{table*}[t]
    \centering
    \caption{Comparison of TTA result of with/without additional flipping. The }
    \begin{tabular}{c|cccc}
    \toprule
         & \textbf{Accuracy} $\uparrow$ & \textbf{F1-score} $\uparrow$ & \textbf{BC} $\downarrow$ & \textbf{NMAE} $\downarrow$ \\
        & \multicolumn{4}{c}{(Micro-OOD/Macro-OOD)} \\
    \midrule
         \textbf{TTA with 4 rotation angles} & 0.829/0.832 & 0.761/0.610 & 0.086/0.100 & 0.261/0.398  \\
         \textbf{TTA with 4 flipped states} & 0.827/0.829 & 0.758/0.595 & 0.086/0.102 & 0.274/0.382  \\
         \textbf{TTA with 4 rotation angles and corresponding vertical flipping} & 0.829/0.832 & 0.761/0.616 & 0.086/0.100 & 0.262/0.386  \\
    \bottomrule
    \end{tabular}
    \label{tab:TTA_flip}
\end{table*}

\bibliography{marco, cheng}

\clearpage

\section*{Acknowledgments}
This project was funded by the Eawag DF project Big-Data Workflow (\#5221.00492.999.01), the Swiss Federal Office for the Environment (contract Nr Q392-1149), and the Swiss National Science Foundation (project 182124).

\end{document}